%% file: neurips_2026.tex
\documentclass{article}

\usepackage[preprint]{neurips_2026}

\usepackage[utf8]{inputenc} 
\usepackage[T1]{fontenc}    
\usepackage{hyperref}       
\usepackage{url}            
\usepackage{booktabs}       
\usepackage{amsfonts}       
\usepackage{nicefrac}       
\usepackage{microtype}      
\usepackage{xcolor}         
\usepackage{amsmath}
\usepackage{paralist}
\usepackage{amssymb}
\usepackage[caption=false]{subfig}  
\usepackage{multirow}
\usepackage{listings}
\usepackage{float}
\newtheorem{example}{Example}

\newtheorem{definition}{Definition}
\input{commands}
\input{tikz_config}
\usetikzlibrary{patterns}
\usepackage{enumitem}
\usepackage{booktabs}
\usepackage{algorithm}
\usepackage{algorithmicx}
\usepackage[noend]{algpseudocode}

\algrenewcommand\algorithmicindent{0.8em}
\usepackage{listings} 
\usepackage{arydshln}
\usepackage{wrapfig}
\usepackage{caption}
\usepackage{multicol}

\usepackage{appendix}

\title{ORCAID: Oblique Rule-Based Continuous-Action \\ Interpretation for Deep RL Policies}

\author{%
  Ignacio D. Lopez-Miguel\\
  TU Wien\\
  Vienna, Austria\\
  \texttt{ignacio.lopez@tuwien.ac.at} \\
  \And
  Ezio Bartocci \\
  TU Wien\\
  Vienna, Austria\\
  \texttt{ezio.bartocci@tuwien.ac.at} \\
  \And
  Thomas Eiter\\
  TU Wien\\
  Vienna, Austria\\
  \texttt{thomas.eiter@tuwien.ac.at} \\
  \And
  Martin Tappler\\
  TU Wien\\
  Vienna, Austria\\
  \texttt{martin.tappler@tuwien.ac.at}
}

\begin{document}

\maketitle

\begin{abstract}

Explainability remains a key issue in reinforcement learning (RL). Distilling an interpretable policy from an agent trained in a complex environment is particularly challenging when the action space is continuous. We introduce \toolname, a novel method for extracting interpretable rule-based policies from RL agents operating in mixed continuous-discrete environments with continuous action spaces. Our main contribution is an efficient oblique decision tree training algorithm that partitions the state space by hyperplanes and fits local linear models. The key idea lies in a three-stage split search: efficient random initialization, local refinement, and backward elimination. Finally, adjacent leaves are merged to yield a concise set of interpretable rules describing a given deep RL policy. We evaluate \toolname across multiple RL environments, demonstrating that the extracted rule-based policies maintain strong performance with a low number of parameters and can even be used to improve the performance of the original deep RL policy.
\end{abstract}

\section{Introduction}
\input{introduction}

\section{Related Work}
\label{sec:rel_work}
\input{related_work}

\section{Preliminaries}
\label{sec:prelim}
\input{prelim}

\section{\toolname Rule Mining}
\label{sec:method}
\input{method}

\input{experiments}

\section{Conclusion}
\label{sec:concl}
\input{conclusion}



\bibliographystyle{plain}
\bibliography{main}

\clearpage

\appendix

\appendixpage

\renewcommand{\arraystretch}{1.2}
\begin{tabular}{p{3cm} p{8cm} r}
\hline
\textbf{Appendix} & \textbf{Title} & \textbf{Page} \\
\hline
\ref{sec:appDagger} & \nameref{sec:appDagger} & \pageref{sec:appDagger} \\
\ref{sec:app_transformation_improvement} & \nameref{sec:app_transformation_improvement} & \pageref{sec:app_transformation_improvement} \\
\ref{sec:appTool} & \nameref{sec:appTool} & \pageref{sec:appTool} \\
\ref{sec:appLLM} & \nameref{sec:appLLM} & \pageref{sec:appLLM} \\
\ref{sec:Apphyperparams} & \nameref{sec:Apphyperparams} & \pageref{sec:Apphyperparams} \\
\ref{sec:app_main_exp} & \nameref{sec:app_main_exp} & \pageref{sec:app_main_exp} \\
\ref{sec:appAblation} & \nameref{sec:appAblation} & \pageref{sec:appAblation} \\
\hline
\end{tabular}

\input{appendixDagger}

\section{Transformations for Policy Improvement}
\label{sec:app_transformation_improvement}
\input{appendix_legible}

\input{appendixTool}

\section{LLM-as-a-Judge}
\label{sec:appLLM}
\input{appendixLLM}

\input{appendix}

\section{Ablation study}
\label{sec:appAblation}
\input{appendixAblation}

\end{document}

%% file: commands.tex
\newcounter{myenumctr}

\newenvironment{myitemize}
{\begin{list}{--}{\setlength{\topsep}{0pt}
\setlength{\leftmargin}{0pt}
\setlength{\itemsep}{0pt}
\setlength{\itemindent}{13.5pt}}}
{\end{list}}

\usepackage{todonotes}
\usepackage{xspace}
\newcommand{\toolname}{\textsc{Orcaid}\xspace}
\newcommand{\toolnameAbbr}{\textsc{Orc.}\xspace}
\newcommand{\legible}{\textsc{Legible}\xspace}

\newcommand{\mountainCar}{\textbf{MC}}
\newcommand{\lunarLander}{\textbf{LL}}
\newcommand{\pendulum}{\textbf{P}}
\newcommand{\invPendulum}{\textbf{IP}}
\newcommand{\invDoublePendulum}{\textbf{IDP}}
\newcommand{\reacher}{\textbf{R}}
\newcommand{\swimmer}{\textbf{S}}
\newcommand{\hopper}{\textbf{H}}
\newcommand{\halfCheetah}{\textbf{HC}}

\newcommand{\size}[1]{\|#1\|}

\newif\ifappendixhyperlinks
\appendixhyperlinkstrue   

\newcommand{\appref}[1]{%
  \ifappendixhyperlinks
    \ref{#1}
  \else
    \ref*{#1}
  \fi
}

%% file: introduction.tex
Reinforcement learning (RL)~\cite{Sutton:1998} builds decision-making policies by trial and error to optimize rewards associated with a control task. It substantially advanced in the last decade by the integration of deep neural networks into training; such \emph{deep RL approaches}\/  have managed to learn to play computer games at a human level~\cite{DBLP:journals/nature/MnihKSRVBGRFOPB15} and reached superhuman performance in complex board games, like Go~\cite{DBLP:journals/nature/SilverHMGSDSAPL16}. 
Deep RL has also ventured into areas where reward-based task specifications are simpler than alternative formulations. Recent applications include fine-tuning LLMs~\cite{DBLP:journals/corr/abs-2307-09288}, learning policies for control problems in the natural sciences~\cite{DBLP:journals/nature/DegraveFBNTCEHA22}, and use in several autonomous systems~\cite{Govinda2025}.

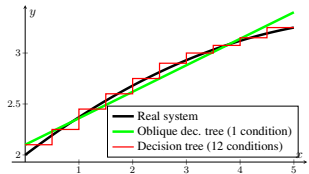
\begin{wrapfigure}{R}{0.3\textwidth}
    \vspace{-\intextsep}
    \scalebox{0.47}{\input{images/dt_vs_odt}}
    \caption{DT vs. Oblique DT.}
    \vspace{-8pt}
    \label{fig:dt_vs_odt}
\end{wrapfigure}

Despite this success, the application of deep RL is hampered by the opacity of learned control policies, which are difficult to verify and understand. This has led regulators to require greater transparency in AI systems used in critical domains, e.g.\ the \cite{exec_order_14110} and the \cite{AI_act}. For example, the EU AI Act specifies in Article~13.3 ``The instructions for use shall contain [...] technical capabilities and characteristics of the high-risk AI system to provide information relevant to explain its output.''  Explainable RL (XRL) methods~\cite{MilaniTVF24} address this by providing human-interpretable explanations of RL policies. 
Explanations are often symbolic models, such as \emph{decision trees}~\cite{BastaniPS18}, which can be verified
when sufficiently small, paving the way for deploying RL 
in safety-critical domains. However, much XRL work targets tasks with discrete control actions, as in early deep RL applications with discrete (yet very large) state spaces \cite{DBLP:journals/nature/MnihKSRVBGRFOPB15,DBLP:journals/nature/SilverHMGSDSAPL16}.

In this paper, we address the challenge of learning \textbf{small, interpretable surrogate models} of RL policies with continuous action spaces and continuous-discrete state spaces. We propose \toolname (Oblique Rule-Based Continuous-Action Interpretation for Deep RL Policies) to construct rule-based models from RL policies. \toolname first partitions the state space by learning an oblique decision tree~\cite{murthy:1994} (extended for discrete/categorical features). 
Unlike axis-aligned trees, oblique trees use 
hyperplanes that are linear combinations of features, allowing them to capture non-orthogonal boundaries compactly. As illustrated in 
Figure~\ref{fig:dt_vs_odt}, an oblique decision tree can approximate a curvilinear boundary with much fewer splits than an axis-aligned decision tree. 

To model continuous actions, we fit linear regressions for each action dimension inside every state-space region (leaf). We then simplify the learned tree by merging similar regions of the state space and using backward elimination to drop unnecessary features. This process generates \emph{oblique decision rules}~\cite{SETIONO19971}, known to be explainable~\cite{Krishnamurthy2021}, extended to produce continuous outputs via a linear model over features in each rule's head. 
Experiments on nine control tasks show that \toolname models outperform SOTA baselines by more accurately representing the original policy (fidelity), maintaining task-solving capability (performance), and achieving a smaller model size (complexity).

\textbf{Our contributions} can be summarized as follows:
\vspace{1pt}
\begin{myitemize}
    \item We introduce \toolname, an approach for learning decision rules from deep RL policies in continuous action spaces over mixed continuous-discrete state spaces.
    \item We provide an implementation of \toolname: \url{https://gitlab.tuwien.ac.at/ignacio.lopez/ORCAID}.
    \item We evaluate \toolname, showing superior performance compared to existing approaches, actionability through policy improvement, and better interpretability measured by model size.
\end{myitemize}

%% file: images/dt_vs_odt.tex
\begin{tikzpicture}
\begin{axis}[
    xlabel={$x$},
    ylabel={$y$},
    axis lines=middle,
    enlargelimits=0.05,
    legend pos=south east,
    legend style={font=\footnotesize, cells={anchor=west}},
    width=9.9cm,
    height=6cm,
    domain=0:5,
    samples=100,
    clip=false
]

\addplot[line width=2pt, black, smooth, black] {2 + 0.4*x - 0.03*x^2};
\addlegendentry{\normalsize Real system}

\addplot[line width=2pt, green, mark=none] coordinates {
        (0,2.1) (5,3.4)};
\addlegendentry{\normalsize Oblique dec. tree (1 condition)}

\addplot[line width=1pt, red, const plot] coordinates {
    (0,2.1) (0.5,2.25) (1,2.45) (1.5,2.6)
    (2,2.75) (2.5,2.9) (3,3) (3.5,3.075)
    (4,3.15) (4.5,3.25) (5,3.25)
};
\addlegendentry{\normalsize Decision tree (12 conditions)}

\end{axis}
\end{tikzpicture}

%% file: related_work.tex
Our work focuses on explainable reinforcement learning (XRL) through decision trees and rules. Milani et al.~\cite{MilaniTVF24} categorize XRL methods into various ways. We specifically target post-hoc feature importance explanations, aiming to interpret actions based on the current state. 

VIPER~\cite{BastaniPS18} is an example in this category, generating \emph{Decision Trees} (DTs) through imitation learning. However, VIPER and its multi-agent extension MAVIPER~\cite{MilaniZTSKPF22} use the CART algorithm~\cite{BreimanFOS84} that generates DTs with axis-aligned predicates, yielding large models, and are limited to Q-value-based policies. Recently, Dhebar et al.~\cite{DhebarDNZF24} have investigated the use of DTs with nonlinear predicates using evolutionary search, but they consider only discrete actions.
The same limitation also appears in \cite{DAI2022107932} on policy extraction using oblique DTs, \cite{SilvaGKJS20} on differentiable DTs for XRL, \cite{Coppens2019} on soft DTs, and \cite{Nageshrao2019} on IF-THEN fuzzy rules for XRL.

Similarly to us, \cite{HeinUR18} considered continuous states and actions, but using genetic
programming to derive interpretable policies in the form of loop-free programs, and limited to simpler RL problems. 
\cite{10.5555/3709347.3743863} also proposed learning rules in continuous states. However, they only considered discrete actions and axis-aligned predicates.  
In contrast, our \toolname approach supports mixed continuous-discrete states and continuous actions.

Additionally, some XRL methods provide insight into the temporal behavior of RL policies in relation to state transitions, either in the environment~\cite{McCalmonLA022,TopinV19,SreedharanSK20} or within policies using memory~\cite{DaneshKFK21,KoulFG19}. We only focus on non-temporal explanations in this paper. Another area of research considers learning inherently interpretable policies~\cite{DBLP:conf/rss/PalejaNSRCG22}, whereas we aim to explain opaque policies.

%% file: prelim.tex
\paragraph{Reinforcement learning} is a machine learning approach where an agent learns to maximize cumulative reward through interaction~\cite{Sutton:1998}. Deep RL with neural networks yields powerful but opaque policies, challenging interpretability (e.g., Go~\cite{DBLP:journals/nature/SilverHMGSDSAPL16}). We formalize continuous control as an MDP $\langle \mathcal{S}, \mathcal{A}, p, \rho_0, r, \gamma\rangle$
\cite{DBLP:journals/corr/LillicrapHPHETS15} with state space $\mathcal{S}\subseteq \mathbb{R}^n$, action space $\mathcal{A}\subseteq\mathbb{R}^m$, transition density $p(\mathbf{s}'|\mathbf{s},\mathbf{a})$, initial state density $\rho_0(\mathbf{s})$, reward $r$, and discount factor $\gamma$. The goal is to learn a policy $\boldsymbol{\pi}: \mathcal{S}\to\mathcal{A}$ maximizing the expected discounted return $\mathbb{E}\bigl[\sum_{t=0}^\infty \gamma^t\,r(\mathbf{s}_t,\mathbf{a}_t)\bigr]$. We use deep deterministic policy gradient (DDPG)~\cite{DBLP:conf/icml/SilverLHDWR14} to learn policies in our experiments.

\begin{example}[Mountain Car]
    The Mountain Car is a popular control problem in RL~\cite{Moore1990, gym}. A car starts at the bottom of a valley. By observing its continuous position and velocity, it applies continuous left/right acceleration, learning to accumulate enough energy to climb the hill.
\end{example}

\paragraph{Explainability.} 
We focus on explainable RL (XRL), which aims to create explanations for the decisions made by RL policies. Specifically, we focus on post-hoc explanations, especially through \emph{feature importance}~\cite{MilaniTVF24} methods. 
\emph{Post-hoc} methods generate interpretable surrogate models that mimic non-interpretable policies, emphasizing the link between individual states and actions as opposed to long-term behavior and training influences.

We assume a \textbf{given} policy $\boldsymbol{\pi}\,{:}\, \mathcal{S} \,{\rightarrow }\, \mathcal{A}$ encoded by a deep neural network. The \textbf{goal} is to learn a small surrogate $\boldsymbol{\hat{\pi}}\,{:}\,{\mathcal{S}}\,{\rightarrow}\, \mathcal{A}$ producing actions close to those of $\boldsymbol{\pi}$. 
As common \cite{DAI2022107932,DhebarDNZF24,DBLP:conf/rss/PalejaNSRCG22}, we measure \textit{model complexity} by parameter count to assess interpretability. Closeness between actions (\emph{fidelity}~\cite{MilaniTVF24}) is measured by an error function $\mathcal{L}(\boldsymbol{\pi}(\mathbf{s}), \boldsymbol{\hat{\pi}}(\mathbf{s}))$, like mean-squared error (MSE). We also assess \emph{performance}~\cite{MilaniTVF24}, which refers to how well the generated surrogate model solves the task.

As we consider finite-horizon tasks ($H\,{\in}\,\mathbb{N}$ steps), we measure performance by the undiscounted return $G_\pi = \sum_{t=0}^H r_t$; hence the ratio $G_{\boldsymbol{\hat{\pi}}}/G_{\boldsymbol{\pi}}$ should be close to $1$. To demonstrate \emph{actionability}, we adapt the \legible approach~\cite{DBLP:conf/ijcai/TapplerLTB25} from discrete to continuous environments, effectively showing that \toolname rules provide actionable insights to improve RL policies. Furthermore, we use LLM-as-a-Judge~\cite{DBLP:journals/corr/abs-2411-15594} as an extra way to assess the explainability of $\boldsymbol{\hat{\pi}}$. Finally, complementing this work, an interactive tool is developed to help explain \toolname models (App.~\appref{sec:appTool}).

\paragraph{Rule mining} discovers IF-THEN rules that identify patterns (descriptive) or make predictions (predictive)~\cite{FurnkranzK15}. 
Various approaches to rule mining exist, like covering algorithms~\cite{FurnkranzK15} and tree-based methods. RuleFit~\cite{friedman2008ruleensembles} is an example of the latter, which uses decision trees to generate rule predicates and then incorporates them as binary features into a single, global linear model with the original features. We use trees (albeit oblique) to discover predicates that define regions within the state space and fit separate linear models per region using only the original features.

\begin{figure*}[tb]
    \centering
    \input{images/overview_method}
    \vspace{-16pt}
    \caption{Overview of our proposed method.}
    \label{fig:overview}
    \vspace{-12pt}
\end{figure*}
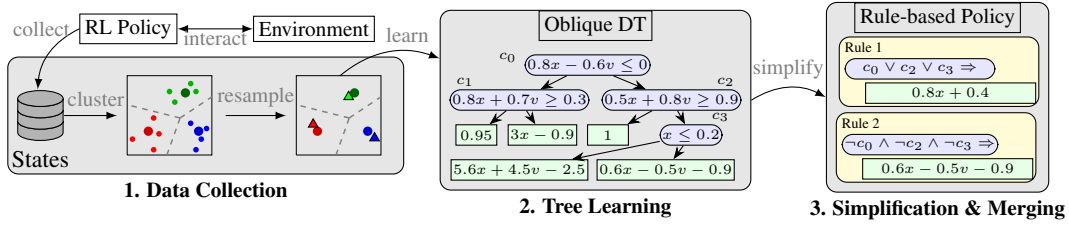

%% file: images/overview_method.tex
\begin{tikzpicture}[
    node distance=2cm and 4cm,
    simple/.style={rectangle, draw, fill=blue!20, text centered, minimum height=2em},
    difficult/.style={rectangle, draw, fill=red!20, text centered, minimum height=2em},
    arrow/.style={-Stealth, thick},
    env/.style={ellipse, draw, fill=green!20, text centered, minimum height=2em    
    },
    font=\footnotesize,
]

\node[draw, inner sep=2pt](agent_1){RL Policy};
\node[right = 1cm of agent_1,draw, inner sep=2pt](env){Environment};
\draw[latex-latex](agent_1) --node[below=-0.1cm](interaction){\textcolor{gray}{interact}} (env);

\node[database, database radius=0.3cm, database segment height=0.15cm, database top segment={fill=black!30}, 
database middle segment={fill=black!30}, 
database bottom segment={fill=black!30},
below left = 0.6cm and 0.2cm of agent_1,
label=below:\small \shortstack{States}](traj){};

\node[right = 0.7cm of traj] (cluster){\scalebox{0.7}{\input{images/kmeans}}};
\node[right = 0.8cm of cluster](resample){\scalebox{0.7}{\input{images/resample}}};

\node[right = 0.6cm of resample](dt){\input{images/dt}};
\node[below left = 0.1275cm and 0.01cm of resample.north](resampleNorth){};

\node[above right = -2.0cm and 1cm of dt](rules){\input{images/rules_image}};

\draw[-latex](agent_1.west) to[bend right] node[pos=0.0, left]{\textcolor{gray}{collect}} (traj);
\draw[-latex](traj) to node[pos=0.6, above]{\textcolor{gray}{cluster}} (cluster);
\draw[-latex](cluster) to node[above]{\textcolor{gray}{resample}} (resample);

\node[above = 0.0cm of dt,draw,inner sep=2pt](dt_text){\shortstack{Oblique DT}};

\begin{scope}[on background layer]
\node[fit=(dt)(dt_text), rounded corners, fill=black!10,draw, inner sep=0](dt_fit){};
\end{scope}

\draw[-latex](resampleNorth) to[bend left] node[pos=0.7, above]{\textcolor{gray}{\shortstack{learn}}} (dt_fit.165);

\node[above = -0.08cm of rules,draw, inner sep=2pt](rule_text){\shortstack{Rule-based Policy}};

\begin{scope}[on background layer]
\node[fit=(rules)(rule_text), rounded corners, fill=black!10,draw, inner sep=0pt](rule_fit){};
\end{scope}

\draw[-latex](dt_fit.east) to[bend left] node[above]{\textcolor{gray}{\shortstack{simplify}}} (rule_fit.west);

\begin{scope}[on background layer]
\node[fit=(traj)(cluster) (resample), rounded corners, fill=black!10,draw](collect_fit){};
\end{scope}

\node[below = 0.1cm of collect_fit, inner sep=0pt]{\textbf{1. Data Collection}};

\node[below = 0.1cm of dt_fit, inner sep=0pt]{\textbf{2. Tree Learning}};

\node[below = 0.1cm of rule_fit, inner sep=0pt]{\textbf{3. Simplification \& Merging}};


\end{tikzpicture}

%% file: images/kmeans.tex
\begin{tikzpicture}[scale=1, every node/.style={font=\tiny}]


\draw[-](0.4,0.4) -- (0.4,1.9) -- (2.0,1.9)--(2.0,0.4)--(0.4,0.4);

\foreach \x/\y in {0.5/0.7, 0.9/0.5, 1.0/0.8, 0.9/1.1} {
  \fill[red] (\x,\y) circle (1.5pt);
}
\fill[red!80!black] (0.8,0.85) circle (2.5pt); 

\foreach \x/\y in {1.8/0.7, 1.7/0.5, 1.9/0.9, 1.6/1.0} {
  \fill[blue] (\x,\y) circle (1.5pt);
}
\fill[blue!80!black] (1.75,0.85) circle (2.5pt); 

\foreach \x/\y in {1.6/1.75, 1.2/1.6, 1.7/1.6, 1.3/1.4} {
  \fill[green!70!black] (\x,\y) circle (1.5pt);
}
\fill[green!40!black] (1.5,1.575) circle (2.5pt); 

\draw[dashed, thick, gray] (1.4,1.1) -- (0.4,1.35); 
\draw[dashed, thick, gray] (1.15,0.4) -- (1.4,1.1); 
\draw[dashed, thick, gray] (1.4,1.1) -- (2.0,1.5); 

\end{tikzpicture}

%% file: images/resample.tex
\begin{tikzpicture}[scale=1, every node/.style={font=\tiny}]


\draw[-](0.4,0.4) -- (0.4,1.9) -- (2.0,1.9)--(2.0,0.4)--(0.4,0.4);

\fill[red!80!black] (0.8,0.85) circle (2.5pt); 

\node[fill=red, draw, regular polygon, regular polygon sides=3, minimum size=6pt, inner sep=0pt] at (0.65,0.98) {};
\fill[blue!80!black] (1.75,0.85) circle (2.5pt); 

\node[fill=blue, draw, regular polygon, regular polygon sides=3, minimum size=6pt, inner sep=0pt] at (1.81,0.72) {};
\fill[green!40!black] (1.5,1.575) circle (2.5pt); 
\node[fill=green, draw, regular polygon, regular polygon sides=3, minimum size=6pt, inner sep=0pt] at (1.32,1.5) {};

\draw[dashed, thick, gray] (1.4,1.1) -- (0.4,1.35); 
\draw[dashed, thick, gray] (1.15,0.4) -- (1.4,1.1); 
\draw[dashed, thick, gray] (1.4,1.1) -- (2.0,1.5); 

\end{tikzpicture}

%% file: images/dt.tex
\begin{tikzpicture}[
  font=\tiny, node distance=0.2cm,
  every node/.style={align=center,
  inner sep= 1pt},
  decision/.style={rectangle, draw, rounded corners, minimum width=0.5cm, minimum height=0.3cm, fill=blue!10},
  leaf/.style={rectangle, draw, minimum width=0.6cm, minimum height=0.3cm, fill=green!10},
  ->, >=Stealth,scale=0.9, transform shape
]

\node[decision] (n1) {$0.8x -0.6v \leq 0$}; 
  \node[decision,below left =  0.2cm and -0.95cm of n1] (n2) {$0.8x +0.7v \geq 0.3$}; 
  \node[decision,below right =  0.2cm and -0.72cm of n1] (n3) {$0.5x +0.8v \geq 0.9$}; 
  \node[leaf,below left = 0.2cm and -0.7cm of n2] (l1) {$0.95$}; 
  \node[leaf,below right =  0.2cm and -1.2cm  of n2] (l2) {$3x - 0.9$}; 
  \node[leaf,below left =  0.2cm and -0.4cm of n3] (l3) {$1$};  
  \node[decision,below right =  0.2cm and -1.2cm of n3] (n4) {$x \leq 0.2$};
  \node[leaf,below left =  0.2cm and 1.05cm of n4] (l4) {$ 5.6x + 4.5v -2.5$};
  \node[leaf,below right =  0.2cm and -1.85cm of n4] (l5) {$0.6x-0.5v-0.9$};

\node[above left = -0.15cm and 0.08cm of n1] {$c_0$};
\node[above left = 0cm and -0.4cm of n2] {$c_1$};
\node[above right = 0cm and -0.4cm of n3] {$c_2$};
\node[above right = 0cm and -0.2cm of n4] {$c_3$};

\draw (n1) -- (n2);
\draw (n1) -- (n3);
\draw (n2) -- (l1);
\draw (n2) -- (l2);
\draw (n3) -- (l3);
\draw (n3) -- (n4);
\draw (n4) -- (l4);
\draw (n4) -- (l5);

\end{tikzpicture}




%% file: images/rules_image.tex
\begin{tikzpicture}[
  font=\tiny, node distance=0.2cm,
  every node/.style={align=center,
  inner sep= 1pt},
  decision/.style={rectangle, draw, rounded corners, minimum width=2.2cm, minimum height=0.3cm, fill=blue!10},
  leaf/.style={rectangle, draw, minimum width=2.5cm, minimum height=0.3cm, fill=green!10},
  ->, >=Stealth,scale=0.9, transform shape
]

\node[decision](b1){$c_0 \lor c_2 \lor c_3  \Rightarrow$};
\node[below right = 0.2cm and 0.3cm of b1.west,leaf](h1){$0.8x + 0.4$};

\node[above left =  0.2cm and -0.6cm of b1.west](r1){Rule 1};
\node[decision, below = 0.8cm of b1](b2){$\lnot c_0 \land \lnot c_2 \land \lnot c_3 \Rightarrow$};        

\node[below right = 0.2cm and 0.3cm of b2.west,leaf](h2){$0.6x - 0.5v -0.9$};

\node[above left = 0.2cm and -0.6cm of b2.west](r2){Rule 2};

\begin{scope}[on background layer]
\node[fit=(b1)(h1)(r1), rounded corners, fill=yellow!20,draw]{};
\node[fit=(b2)(h2)(r2), rounded corners, fill=yellow!20,draw]{};
\end{scope}

\end{tikzpicture}

%% file: method.tex
\toolname learns a rule-based surrogate model that approximates the decision-making behavior of an RL policy over continuous state and action spaces. 
It operates in three main phases, shown in Fig.~\ref{fig:overview}:
\begin{compactenum}[1.]
\item \textbf{Data Collection.} 
We sample policy trajectories from interactions between the policy and the environment, cluster the visited states, and then resample by perturbing random centroids with noise. This data-augmentation strategy reduces the amount of data and mitigates overfitting.
\item \textbf{Tree Learning.}
We construct oblique decision trees and fit a separate linear model per action dimension in each leaf, enabling a more accurate policy approximation. We use backward feature elimination to reduce the complexity of the trees and the linear functions.
\item \textbf{Simplification \& Merging.} We finally produce simplified representations of the policy in the form of rules. This involves merging regions defined by decision tree nodes, simplifying the conditions that define these merged regions, and updating the associated linear models.
\end{compactenum}
We further embed these steps into a \textbf{DAgger (Dataset Aggregation)}~\cite{ross2011} loop to improve accuracy. Ablation results (App.~\ref{sec:appAblation}) also justify our approach by analyzing the effect of isolating each component’s effect and varying hyperparameters.

\paragraph{Method Outline.}
Given an RL policy $\boldsymbol{\pi} : \mathcal{S} \rightarrow \mathcal{A}$ over continuous states $\mathcal{S} \subseteq \mathbb{R}^n$ and actions $\mathcal{A} \subseteq \mathbb{R}^m$, \toolname computes a policy $\boldsymbol{\hat{\pi}}: \mathcal{S} \rightarrow \mathcal{A}$ that approximates $\boldsymbol{\pi}$ consisting of rules $r_i$ of the form $c_i \Rightarrow \boldsymbol{\hat{\pi}}_i$, where $c_i$ is a \emph{triggering condition} from the set $C(\mathcal{S})$ of Boolean formulas over linear inequalities over states, and $\boldsymbol{\hat{\pi}}_i$ is an \emph{action proxy} from the set $F(\mathcal{S},\mathcal{A})$  of affine functions from $\mathcal{S}$ to $\mathcal{A}$. 
The conditions $c_i$ collectively cover $\mathcal{S}$, and
\toolname aims to minimize both the approximation error $\mathcal{L}(\boldsymbol{\pi},\boldsymbol{\hat{\pi}})$
and the model size of $\boldsymbol{\hat{\pi}}$, measured by the number of parameters in the rules $r_i$.

\subsection{Data collection}
\label{sec:data_collection}
The initial data collection involves the following three steps.

\begin{compactenum}[1.]
    \item \textbf{Agent execution}. Running the original RL policy in the environment collects trajectories (state-action sequences). Each state $\mathbf{x}=(\mathbf{x}^c,\mathbf{x}^d)\in\mathbb{R}^n$ concatenates continuous features $\mathbf{x}^c = [x_1^c, x_2^c, \dots, x_{n_c}^c]\in[0,1]^{n_c}$
    over the unit interval and binary features $\mathbf{x}^d = [x_1^d, x_2^d, \dots, x_{n_d}^d]\in\{0,1\}^{n_d}$ (ensured by scaling or one-hot encoding).
    \item \textbf{Clustering}. Trajectories visit some regions (e.g., near the initial state) far more often than others, which can bias regressions toward them. To obtain a more uniform state‑space representation, we group observed states into $k$ clusters using k‑means, where $k$ is chosen by the elbow method~\cite{Thorndike_1953}, collapsing dense regions into single centroids to reduce their oversampling effect.
    \item \textbf{Sampling}. To mitigate potential model drift—where the surrogate model deviates from the original trajectories due to accumulated small errors—we sample states from the centroids with Gaussian noise. That is, each sampled point $\mathbf{x}$ consists of a centroid $\boldsymbol{\mu}_i$ chosen uniformly, to which we add noise only on the continuous features:
    $\mathbf{x}\!=\!\boldsymbol{\mu}\!+\!(\epsilon_1,\dots,\epsilon_{n_c},0,\dots,0)$, with $\epsilon_i\!\sim\!\mathcal{N}(0,\sigma)$. 
    For each sample $\mathbf{x}$, we query $\boldsymbol{\pi}$ to get its action $\mathbf{y}\!=\!\boldsymbol{\pi}(\mathbf{x})\!\in\!\mathbb{R}^{m}$ ($m\!=\!1$ for 1-D action environments).
\end{compactenum}

Clustering accelerates computation, and sampling helps prevent overfitting by covering more of the state space. This makes the surrogate model more robust to errors that accumulate during execution. In the experiments, we observed that small noise (viz. $\sigma=0.15$, see~App.~\ref{sec:Apphyperparams}) was enough to obtain good results without deviating too much from the original trajectories.

Random state space sampling to collect data could be an alternative, but it scales poorly with dimension. Instead, we focus on regions frequently visited by the RL agent and their neighborhoods, i.e., the relevant states for task performance.

\subsection{Region Definition}

\toolname partitions the state space via oblique decision trees (i.e., using hyperplanes as branching conditions), which are then merged to form disjoint regions.
To formalize this, we first introduce a \emph{condition}, which is a hyperplane inequality that serves as a branching condition.

\begin{definition}[Linear split condition]
\label{def:split}
Let the feature vector be 
\(\mathbf{x} = (\mathbf{x}^c, \mathbf{x}^d) \in [0,1]^{n_c} \times \{0,1\}^{n_d}\), 
i.e., \(n_c\) continuous features (scaled to \([0,1]\)) and \(n_d\) binary features.
A \emph{linear split condition} with weight vector \(\mathbf{w} \in \mathbb{R}^{n_c+n_d}\) and threshold \(b \in \mathbb{R}\) is the function
$
c_{\mathbf{w},b}(\mathbf{x}) = \mathbb{I}\bigl( \mathbf{w} \cdot \mathbf{x} \leq b \bigr)
$,
where \(\mathbb{I}\) is the indicator function (1 if true, 0 otherwise).
\end{definition}

\noindent
By imposing constraints on the weights, the same linear form can represent three kinds of splits:

\begin{compactenum}[1.]
    \item \textbf{Split on continuous features only.}
    If \(\mathbf{w}^d = \mathbf{0}\), then \(c_{\mathbf{w},b}(\mathbf{x}) = \mathbb{I}(\mathbf{w}^c \cdot \mathbf{x}^c \leq b)\).

    \item \textbf{Split testing ``binary feature \(i\) equals 0''.}
    If \(\mathbf{w}^c = \mathbf{0}\), \(w_i^d = 1\), \(b = 0\), and $w_j^d=0$ for all $j\neq i$, then \(\mathbf{w} \cdot \mathbf{x} = x_i^d\), and the inequality \(x_i^d \leq 0\) forces \(x_i^d = 0\).

    \item \textbf{Split testing ``binary feature \(i\) equals 1''.}
    If \(\mathbf{w}^c\! =\! \mathbf{0}\), \(w_i^d\! =\! -1\), \(b\! =\! -1\), and $w_j^d\!=\!0$ for all $j\!\neq\! i$, then \(\mathbf{w} \cdot \mathbf{x} = -x_i^d\), and the inequality \(-x_i^d \leq -1\) simplifies to \(x_i^d \geq 1\), i.e., \(x_i^d = 1\).
\end{compactenum}

\noindent
If the binary features originate from one‑hot encoding a categorical variable, case 2 checks that the category is \emph{not} the one represented by feature \(i\), and case 3 checks that it \emph{is} that category. For a native binary variable (already 0/1), the same logic applies directly.

We now introduce regions. Each leaf of an oblique decision tree defines a \emph{simple region} induced by the conditions at its ancestor nodes. Formally, simple regions are defined as follows.
\begin{definition}[Simple region]
    Given a set $C$ of conditions, the simple region $R_C$ defined by $C$ is
\smallskip
    
\centerline{$R_C = \{\mathbf{x}\in [0,1]^{n_c}\times\{0,1\}^{n_d} \mid \forall c\in C: c(\mathbf{x}) =1\, \}.$}
\end{definition}
\noindent That is, $\mathbf{x}\,{\in}\,R_C$ iff
$\mathbf{w}\,{\cdot}\,\mathbf{x}\,{\leq}\,b$ holds for each condition $c_{\mathbf{w},b}$ in $C$. For merging leaves, we need more expressive regions. 
\begin{definition}[Region]
A region is the union $R = \bigcup_{i=1}^k R_{C_i}, (k\geq 1),$ of simple regions $R_{C_i}$.
\end{definition}
Thus $\mathbf{x} \in R$ iff for some $R_{C_i}$ forming $R$, $\mathbf{w}\cdot \mathbf{x}\leq b$ holds for each condition $c_{\mathbf{w},b}\in C_i$.
\toolname efficiently assigns points to regions via matrix multiplication, thanks to the structure of the conditions.

\begin{algorithm}[tb]
\caption{Oblique Tree Training with Hyperplane Optimization}
\label{algo:oblique_tree_full}
\small
\vspace{-14pt}
\begin{multicols}{2}
\begin{algorithmic}[1]
\Require $X\in\mathbb{R}^{N\times n}$, $Y\in\mathbb{R}^{N\times m}$, \texttt{max\_depth},  
\Statex \hspace{-\leftmargin}\texttt{min\_spl}, \texttt{max\_mse}, \texttt{pca\_dim}
\Ensure Root node of the oblique decision tree

\Procedure{BuildNode}{$X, Y, depth$}
    \If{$depth \ge \texttt{max\_depth}$ \textbf{or} $|X| < \texttt{min\_spl}$}
        \State \Return \Call{LeafNode}{$X, Y$}
    \EndIf
    \State $\boldsymbol{\beta} \gets (\mathbf{X}^\top \mathbf{X})^{-1}\mathbf{X}^\top Y$
    \State $\mathbf{MSE} \gets \frac{1}{|X|}\big[\operatorname{diag}(Y^\top Y)\!-\operatorname{diag}(\boldsymbol{\beta}^\top \mathbf{X}^\top Y)\big]$
    \State $\text{MSE} \gets mean(\mathbf{MSE})$
    \If{$\text{MSE} < \texttt{max\_mse}$}
        \State \Return \Call{LeafNode}{$X, Y$}
    \EndIf
    \State $(w, b) \gets \Call{FindSplit}{X, Y}$
    \State $L \gets \{i\mid X_i w \le b\}$, $R \gets \{i\mid X_i w > b\}$
    \State $\textit{left} \gets \Call{BuildNode}{X[L], Y[L], depth+1}$
    \State $\textit{right}\!\gets\!\!\Call{BuildNode}{X[R], Y[R], depth+\!1}$
    \State \Return $\langle w,b,\textit{left},\textit{right}\rangle$
\EndProcedure

\Procedure{FindSplit}{$X, Y$}
    \State Apply PCA (\texttt{pca\_dim} comp.): $X_{\text{PCA}}, P, \mu$
    \State Min‑max scale $X_{\text{PCA}}$ to $[0,1]$ ($S$ = scaler)
    \State $(w_{\text{PCA}}, \textit{loss}) \gets \Call{RandHS}{X_{\text{PCA}},\texttt{pca\_dim}, Y}$
    \If{$\textit{loss} > \texttt{max\_mse}$}
        \State Run differential evolution (obj = loss, constraint: min. samples per side)
        \State $w_{\text{PCA}} \gets$ result of evolution
    \EndIf
    \State $w_{\text{orig}} \gets P^\top S^{-1} w_{\text{PCA}}$ \Comment{map back}
    \State \Return $(w_{\text{orig}}, b_{\text{orig}})$
\EndProcedure

\Procedure{RandHS}{$X_{\text{PCA}}, \texttt{pca\_dim}, Y$}
    \State $\textit{best\_loss} \gets \infty$, $\textit{best\_w} \gets \mathbf{0}$
    \For{each batch (run in parallel)}
        \State Random  \texttt{pca\_dim}  rows of $X_{\text{PCA}} \to A$
        \State Solve $A w = -\mathbf{1}$ for $w$
        \State $p \gets X_{\text{PCA}}\,w$
        \State $L \gets \{p \le -1\}$, $R \gets \{p > -1\}$
        \If{$|L| \ge \texttt{min\_spl}$ \textbf{and} $|R| \ge \texttt{min\_spl}$}
            \State $wMSE \gets \frac{|L|}{|X|}\text{MSE}_L + \frac{|R|}{|X|}\text{MSE}_R$ 
            \If{$wMSE < \textit{best\_loss}$}
                \State $\textit{best\_loss},\  \textit{best\_w} \gets wMSE,\ w$
            \EndIf
        \EndIf
    \EndFor
    \State \Return $(\textit{best\_w}, \textit{best\_loss})$
\EndProcedure

\Procedure{LeafNode}{$X, Y$}
    \For{$j=1,\dots,m$}
        \State $\beta_j \gets$ OLS coeff. for $Y_{:,j}$
        \Repeat
            \State Remove feature $k$ if: $|\beta_{j,k}| < \epsilon$ \textbf{or}
            \State \hspace{\algorithmicindent}its removal increases MSE by $<\delta$
        \Until{no feature can be removed}
        \State Store pruned model for dim. $j$
    \EndFor
    \State \Return leaf node with pruned linear models
\EndProcedure
\end{algorithmic}
\end{multicols}
\vspace{-12pt}
\end{algorithm}

\subsection{Tree building with hyperplane optimization}
\label{sec:tree_building}

Given a set of states $X \in \mathbb{R}^{N \times n}$ and corresponding multi‑dimensional actions $Y \in \mathbb{R}^{N \times m}$, the oblique tree recursively partitions the input space with hyperplanes, which is built using Algorithm~\ref{algo:oblique_tree_full}. In each leaf, a pruned linear model is stored for every action dimension.  
The quality of a split is measured by the weighted mean squared error after fitting separate ordinary least‑squares (OLS) regressions in each child region.
Each recursive step includes: (1) \textbf{dimensionality reduction and scaling} to handle high-dimensional state space, (2) \textbf{random hyperplane initialization} to find a good partitioning fast (warm start), and (3) \textbf{local split optimization} to partition states optimally. 

\begin{definition}[Weighted split MSE]
Let $(\mathbf{w}, b)$ define a linear split condition as in Def.~\ref{def:split}, i.e., 
the hyperplane $\mathbf{w}\cdot \mathbf{x} = b$ splits the state space into $
L = \{i \mid \mathbf{w} \cdot \mathbf{x}_i \le b\}$, and $  
R = \{i \mid \mathbf{w} \cdot \mathbf{x}_i > b\}$. For each action dimension $k \in \{1,\dots,m\}$, let $\hat{\boldsymbol{\beta}}^{L,k}$ and $ \hat{\boldsymbol{\beta}}^{R,k}$ the OLS coefficients fitted on $(\mathbf{X}_L, Y_{L,k})$ and $(\mathbf{X}_R, Y_{R,k})$, respectively, where $\mathbf{X}$ is $X$ with an added column of $1$s. The weighted split MSE is
\[ \textstyle 
\ell(\mathbf{w},b) \;=\;
\frac{|L|}{N}\,\frac{1}{m}\sum_{k=1}^{m} \mathrm{MSE}(\mathbf{X}_L,Y_{L,k},\hat{\boldsymbol{\beta}}^{L,k})
\;+\;
\frac{|R|}{N}\,\frac{1}{m}\sum_{k=1}^{m} \mathrm{MSE}(\mathbf{X}_R,Y_{R,k},\hat{\boldsymbol{\beta}}^{R,k}),
\]
where $\mathrm{MSE}$ is the mean squared residual. An optimal oblique split is one that minimizes this loss.
\end{definition}

Algorithm~\ref{algo:oblique_tree_full} recursively
partitions the data until a stopping criterion is met -- reaching \texttt{max\_depth}, having fewer than \texttt{min\_spl} samples (line~2), or achieving a low mean MSE by an OLS model (line~5--6) -- and a leaf node is created.
 The algorithm creates branch conditions using \textsc{FindSplit} (line~9), which searches for a hyperplane $(\mathbf{w},b)$ minimizing the weighted split MSE. The data is partitioned via $\mathbf{w}\cdot\mathbf{x}_i \le b$, and \textsc{BuildNode} recurses on each side (lines~10–13). The search proceeds in three stages in a reduced PCA space to improve numerical stability and speed.

\begin{compactenum}[1.]
    \item \textbf{Dimensionality reduction and scaling.}
    PCA is applied to the set of states $X$, retaining \texttt{pca\_dim} components (line~15). The projected data $X_{\mathrm{PCA}}$ is then min‑max scaled to $[0,1]$ (line~16). The scaler $S$ and the PCA parameters ($P$, mean $\mu$) are kept for later back‑mapping.
    
    \item \textbf{Random hyperplane initialization (lines~23-34).}
    \texttt{pca\_dim} states of $X_{\mathrm{PCA}}$ are sampled to form a square matrix $A$ and the system $A\mathbf{w} = -\mathbf{1}$ is solved (line~27) to obtain a hyperplane that contains those \texttt{pca\_dim} states. This step can be done efficiently and in parallel for multiple hyperplanes. 
    $A\mathbf{w} = -\mathbf{1}$ prevents the trivial $\mathbf{w}\!=\!\mathbf{0}$ solution, forces a direction through the selected points, and absorbs the bias into the fixed offset $-1$. 
    Points with $\mathbf{w}\cdot\mathbf{x}_i \le -1$ go left, others right (line~29). Only splits with at least \texttt{min\_spl} samples per side are considered (line~30); their weighted MSE is computed (line~31), and the best $\mathbf{w}_{\mathrm{PCA}}$ and its loss are returned (lines~32--34). 

    \item \textbf{Local split optimization via differential evolution.}
    If the loss of the best candidate exceeds \texttt{max\_mse} (line~18), a differential evolution locally optimizes $\mathbf{w}_{\mathrm{PCA}}$, minimizing  the weighted split MSE under a constraint that both children contain at least \texttt{min\_spl} instances (lines~19--20).  The optimized vector replaces $\mathbf{w}_{\mathrm{PCA}}$. Finally, $\mathbf{w}_{\mathrm{PCA}}$ is transformed to the original space via
    $P^{\top} S^{-1}$ (line~21).
\end{compactenum}

\noindent
\textbf{Leaf construction (\textsc{LeafNode}).}  
For every action dimension $j$, an OLS model is fitted (line~37).  
Backward elimination then removes features whose absolute coefficient is smaller
than $\epsilon$ or whose removal increases the MSE by less than $\delta$
(lines~39--40). The pruned model is stored, and the leaf node returns the set of all $m$ linear models (line~43).

\begin{example}[Mountain Car (oblique tree)]
    Fig.~\ref{fig:overview}, block 2, shows an oblique tree of depth 3 for Mountain Car. Note that not all leaves have depth 3 since leaves stop growing if they do not have enough samples or if their MSE is below \texttt{max\_mse}.
\end{example}

\subsection{Simplification}

After constructing the oblique decision tree, we merge adjacent regions to create a simplified rule-based model. Each rule body defines a state-space region in which it triggers, and the corresponding head defines the actions via linear models.
This gives us three major benefits: (1)\ \textbf{reduced complexity}: by decreasing the number of regions, we reduce the number of linear regressions, leading to fewer parameters; (2) \textbf{improved interpretability}: a smaller set of regions makes the surrogate model easier to analyze; and (3) \textbf{diminished overfitting}: merging similar regions discourages overfitting and improves generalization.
For region merging, we use the notion of active condition.

\begin{definition}[Active condition]
For a set  $C$ of conditions, a condition $c_{\mathbf{w},b}\in C$ is called active in the simple region $R_C$ if some $\mathbf{x} \in [0,1]^{n_c}\times\{0,1\}^{n_d}$ exists such that
$(1)\ c(\mathbf{x})\!=\!1$ for every $c \in C \setminus \{c_{\mathbf{w},b}\}$, and $(2)\ \mathbf{w}\!\cdot\! \mathbf{x} = b$, i.e., the point is on the boundary of $R_{c_{\mathbf{w},b}}$.
\end{definition}
This definition also applies to binary features.
Any leaf $R_C$ has at least one active condition in its $C$ (the last split), and can have all of them active. We next define when two simple regions are adjacent.

\begin{definition}[Adjacent simple regions]
Given sets $C_1$ and $C_2$ of conditions, their simple regions $R_{C_1}$ and $R_{C_2}$ are adjacent if a vector weight $\mathbf{w}$ and a threshold $b$ exists such that (i) $c_{\mathbf{w},b}$ is active in $R_{C_1}$ and (ii) $c_{-\mathbf{w},-b}$ is active in $R_{C_2}$.
\end{definition}
Sibling leaves $R_{C_1}$ and $R_{C_2}$ (i.e., having the same parent $R_C$) are always adjacent, as they differ only by the split condition at $R_C$. Non-sibling leaves can also be adjacent; this happens when the split condition is active in both regions.

After identifying adjacent regions, we evaluate potential merges in an iterative procedure. For each pair $R_i$ and $R_j$ of adjacent regions, we fit single linear regressions for each target variable on $R=R_i \cup R_j$ and compute its mean MSE, as well as the MSE of the new and the original regressions on the subparts. 
Adjacent regions $R_i$ and $R_j$ can be merged if  (1) $\text{MSE} < \tau$ (global error bound), (2) $\text{MSE}_{i} < \tau$ and $\text{MSE}_{j} < \tau$ (local error bound), or (3) $\text{MSE}_{k}\,{<}\,(1\,{+}\,\epsilon) \,{\cdot}\,\text{MSE}_k^{\text{orig}}$, $k\,{=}\,i,j$ (rel.\ improvement), where $\epsilon=0.05$ and $\tau=2\cdot$\texttt{max\_mse}.
The resulting merged region is the union of $R_i$ and $R_j$. This process repeats until no further merging is possible.

After region merging, each resulting region $R$ is defined by one or more sets $C_i$ of conditions that define a simple region for a leaf node. 
In logical terms, the definition of $R$ amounts to a disjunction of conjunctions of conditions, i.e., a disjunctive normal form (DNF) over the set of conditions. Hence, we can simplify the merged region definitions using the QM method~\cite{Quine1952,McCluskey1956}. 
Furthermore, for the final linear regressions, we perform backward elimination to remove irrelevant features.

The following definition is used for a fair complexity comparison of different surrogates. We measure complexity with two components: what it takes to (i) \textit{get} to a decision (\textbf{path conditions}) and to (ii) \textit{make} the decision (\textbf{action logic cost}).

\begin{definition}[\toolname  model size]\label{def:number_params}
The size of an \toolname model $M$ is $\size{M}\!=\!M_{conds}\! +\! M_{Bools}\! +\! M_{regrs}$, where $M_{conds}$ is the number of non-zero coefficients in the conditions, $M_{Bools}$ is the count of conditions used to define the regions, and $M_{regrs}$ is the number of non-zero coefficients in the regressions. $M_{conds}+M_{Bools}$ accounts for the path conditions, and $M_{regrs}$ for the action logic.
\end{definition}

\begin{example}[Mountain Car (simplification)]
    The simplification of the oblique tree (Fig.~\ref{fig:overview}, block 3) merges its leaves (simple regions) into two regions (rules): $\mathit{Rule} \ 1$ ($c_0\lor c_2\lor c_3 \Rightarrow 0.8x + 0.4)$, and $\mathit{Rule}\ 2$, $(\lnot c_0 \land \lnot c_2 \land \lnot c_3 \Rightarrow 0.6x - 0.5v - 0.9)$. With only these two rules, we achieve the same performance as with the original RL policy. Conditions have been simplified, e.g., the two leaves $(c_0\land c_1)$ and $(c_0\land\lnot c_1)$ in the oblique tree have been simplified to $c_0$. By Def.~\ref{def:number_params}, $\size{M}\! =\! \overbrace{\underbrace{2}_{c_0}\! +\! \underbrace{2}_{c_2}\! +\! \underbrace{1}_{c_3} }^{M_{conds}}\! +\! \overbrace{\underbrace{2}_{\text{Rule}\ 1}\! +\! \underbrace{3}_{\text{Rule}\ 2}}^{M_{regrs}} + \overbrace{\underbrace{3}_{\text{Rule}\ 1} + \underbrace{3}_{\text{Rule}\ 2}}^{M_{Bools}}\! =\! 16$. 
    \vspace{-9pt}
\end{example}
\subsection{Iterative Improvement via DAgger}
To remedy the accumulation of errors between the original deep RL policy and rules, we iteratively improve rules through \textbf{DAgger (Dataset Aggregation)}~\cite{ross2011} by the following steps (cf.\ Appendix~\appref{sec:appDagger}):
\begin{compactenum}[1.]
    \item \textbf{Collection of expert-labeled trajectories} by executing the current \toolname model in the environment and labeling the actions in the trajectories with the original deep RL policy.
    \item \textbf{Data aggregation} by keeping the states in which the \toolname and the original actions deviate significantly.
    As in Section~\ref{sec:data_collection}, we cluster and resample observations. 
    \item \textbf{Retraining} of the model on all aggregated data.
\end{compactenum}
This repeats until performance stops improving or \toolname's actions and those of the RL agree.

%% file: experiments.tex
\section{Experiments}
\label{sec:exp}

We empirically evaluate \toolname on nine control problems, comparing it to (1) standard decision trees (DT) learned using CART~\cite{BreimanFOS84} implemented in scikit-learn~\cite{scikit-learn}, (2) rules learned with Cubist~\cite{pycubist,Quinlan1992LearningWC}, and (3) RuleFit~\cite{friedman2008ruleensembles}.
We also combine each baseline with DAgger. For DTs, this setup is effectively a Q-value-free variant of VIPER~\cite{BastaniPS18}, including iterative aggregation and learning, but replacing Q-weighted supervision with policy labels to support general deterministic RL policies.

Our \textbf{research questions}\/ are the following:

\begin{compactitem}
    \item[{\bf RQ1:}] Can \toolname learn surrogates of small size, while retaining \emph{performance}?
    \item [{\bf RQ2:}] Can \toolname represent the RL policies accurately, i.e., achieve low MSE (\emph{fidelity})?
    \item [{\bf RQ3:}] Can \toolname rules improve an RL policy?
    \item [{\bf RQ4:}] How interpretable are \toolname models with respect to other surrogates?
\end{compactitem}

For \textbf{RQ1} and \textbf{RQ2}, we compare the average cumulative reward and MSE of each surrogate and contrast it with its model size (Def.~\ref{def:number_params}). For \textbf{RQ3}, we adapt \legible~\cite{DBLP:conf/ijcai/TapplerLTB25}.
\textbf{RQ4} is related to \textbf{RQ1}, as model size is often used for interpretability~\cite{molnar2025}. Human evaluation can provide further insights into comprehensibility. However, it is very costly, audience-specific, and potentially biased. We thus follow the LLM-as-a-Judge framework~\cite{DBLP:journals/corr/abs-2411-15594} to complement model size with a comprehensive LLM-based evaluation of interpretability.

\paragraph{Experimental Setup \& Hyperparameter Selection.} 
We conducted all experiments on a computing cluster with four NVIDIA A100-SXM4-40GB GPUs and two AMD EPYC CPUs. Every RL policy was trained with 3 different seeds, using the DDPG implementation of SB3~\cite{stable-baselines3} (cf.\ App.~\appref{sec:Apphyperparams} Table~\appref{tab:rl_hyperparams}). Every surrogate was also trained with 3 different seeds. Thus, every result for \textbf{RQ1 -- RQ3} is the average of 9 runs. For \textbf{RQ1}, performance was evaluated by running 1,000 episodes per model.

For a fair comparison, we use identical sampling hyperparameters across all surrogates and tuned each surrogate's main hyperparameter: tree depths for \toolname and DT, number of rules for Cubist and RuleFit, and additionally the \texttt{max\_mse} for \toolname. App.~\ref{sec:app_main_exp} lists all results for all surrogate model sizes, and App.~\appref{sec:Apphyperparams}, Table~\ref{tab:orcaid_hyperparams}, all the sampling hyperparameters, and \toolname \texttt{max\_mse}.

We \textbf{evaluate} \toolname and the baselines in nine 
control problems~\cite{gym}: from simple like \emph{Mountain Car} (\textbf{\mountainCar}), to complex like \emph{Hopper} (\textbf{\hopper}) and \emph{Half Cheetah} (\textbf{\halfCheetah}). \emph{Lunar Lander} (\textbf{\lunarLander}) was a case of mixed continuous-discrete state space. The other problems are \emph{Pendulum} (\textbf{\pendulum}), \emph{Inverted Pendulum} (\textbf{\invPendulum}), \emph{Inverted Double Pendulum} (\textbf{\invDoublePendulum}), \emph{Reacher} (\textbf{\reacher}), and \emph{Swimmer} (\textbf{\swimmer}).

\begin{wrapfigure}{r}{0.59\linewidth}
    \vspace{-32pt}
    \scalebox{0.58}{\input{images/plot_ratio_model_size}}
    \caption{Model size \& surrogate/RL reward ratio per environment.}
    \label{fig:plot_ratio_model_size}
    \smallskip
    \scalebox{0.58}{\input{images/plot_MSEs}}
    \caption{Test MSE scaled (best:0) according to the output range to allow comparisons between environments.}
    \label{fig:plot_mses}
    \vspace{-26pt}
\end{wrapfigure}

\subsection{Empirical Results}
For each environment, we learn \toolname and DT models with increasing maximum depth and increasing maximum rules for RuleFit and Cubist, selecting the best-performing model for each surrogate until test MSE stops improving (cf.~App.~\appref{sec:app_main_exp} for all results).
As examples of approximated training times, for \mountainCar, DT takes $5$s, Cubist $40$s, RuleFit $10$s, and \toolname $20$s. For \halfCheetah, DT takes $15$s, Cubist $22$min, RuleFit $2$min, and \toolname $30$min. These values also consider the time spent on model evaluation for DAgger.

Model size for baselines aligns with Def.~\ref{def:number_params} by summing (1) path conditions and (2) action logic cost. For DTs, (1) \#features$\times$\#leaves, (2) depth per leaf. For Cubist: analogous to \toolname, but with single-feature conditions and one model per action dimension. For RuleFit, with one model per action dimension, (1) combined length of all rules, (2) \#coefficients.

\paragraph{Performance.} Fig.~\ref{fig:plot_ratio_model_size} addresses {\bf RQ1} by contrasting performance with model size. Performance is measured by the average ratio between the cumulative reward of each surrogate and the RL policy.
\toolname performs better than or equal to the other surrogates, with RuleFit and DT being far less performant, sometimes failing to reach $75\%$. For \invDoublePendulum, this can be due to the high impact of small deviations in actions on the stick's stability, which also affects Cubist, leading to a high variance in its reward ratio. 
{\toolname}’s model size is much smaller than DT’s (except in \halfCheetah, where performance differs substantially) and smaller than RuleFit’s when performance is similar.
While Cubist shows competitive ratios across most environments, even better for \halfCheetah\ (non-significant, by the confidence intervals), its size is consistently larger than \toolname's (note the log scale). Overall, for {\bf RQ1}, \toolname learns smaller surrogate models with strong performance.

\begin{table}
        \small
        \centering
        \setlength{\tabcolsep}{1.7pt}
        \renewcommand{\arraystretch}{.4}
\parbox{.45\linewidth}{
\centering
\begin{tabular}{l|ll:ll:ll:ll}
                     & \multicolumn{2}{c:}{\toolname} & \multicolumn{2}{c:}{DT} & \multicolumn{2}{c:}{Cubist} & \multicolumn{2}{c}{RuleFit} \\
                      Env. & Rate  & Best  & Rate  & Best  & Rate  & Best  & Rate  & Best \\
                     \hline
        \lunarLander & \textbf{53} & \textbf{5.3} & 50 & 3.9 &  n/a & n/a & n/a & n/a \\
        \invPendulum & \textbf{17} & 20 & 0 & 0 & 11 & \textbf{21} & 11 & 17 \\
        \invDoublePendulum & \textbf{38} & \textbf{5.0} & 0 & 0 & 0 & 0 & 0 & 0 \\
        \reacher & 0 & 0 & 0 & 0 &  n/a & n/a & n/a & n/a\\
        \swimmer & \textbf{61} & \textbf{4.2} & 56 & 3.3 &  n/a & n/a & n/a & n/a\\
        \hopper & \textbf{92} & \textbf{5.5} & 89 & 3.2 &  n/a & n/a & n/a & n/a\\
        \halfCheetah & 14 & \textbf{2.9} & \textbf{19} & 2.6 &  n/a & n/a & n/a & n/a
        \end{tabular}
\vspace{2pt}
\caption{\% of times an improvement was found (Rate) and best improvement found w.r.t.\ the RL policy in \% out of 9 runs (Best).}
\label{tab:results_legible}
}
\hfill
\parbox{.54\linewidth}{
\centering
\begin{tabular}{l|ccccccccc}
                       & \mountainCar  & \pendulum & \lunarLander & \invPendulum & \invDoublePendulum & \reacher & \swimmer & \hopper & \halfCheetah \\
                       \hline
\toolname & \textbf{4.3} & \textbf{4.0} & \textbf{3.6} & \textbf{4.5} & \textbf{3.9} & \textbf{3.1} & \textbf{4.3} & 2.6 & 2.6\\
DT & 2.1 & 1.7 & 1.6 & 1.9 & 1.6 & 2.8 & 1.8 & \textbf{2.8} & \textbf{3.0}\\
\toolnameAbbr wins (\%) & 100       & 100       & 100      & 100       & 100      & 100      & 100       & 50      & 50     \\
\hline
\hline
\toolname & \textbf{4.4} & \textbf{4.6} & \textbf{3.7} & \textbf{4.2} & \textbf{4.5} & \textbf{3.7} & \textbf{4.5} & \textbf{3.0} & \textbf{3.5}\\
Cubist & 2.5 & 1.7 & 1.7 & 2.9 & 1.4 & 1.9 & 1.4 & \textbf{3.0} & 2.7\\
\toolnameAbbr wins (\%) &  100 & 100 & 100 & 100 & 100 & 100 & 100 & 50 & 100
    \end{tabular}
\vspace{5pt}
\caption{Score average across 6 LLMs and 15 questions (worst:1, best:5), and \toolname win rate from LLM verdicts, per environment.}
\label{tab:results_llms}
}
\vspace{-24pt}
\end{table}

\paragraph{Fidelity.}
To assess fidelity, we computed the MSE on test data (Fig.~\ref{fig:plot_mses}). Similar to \textit{performance}, Cubist learns a surrogate that is closest to \toolname. The other surrogates usually have higher MSEs, deviating more from the RL policy. We thus can answer {\bf RQ2} affirmatively, i.e., explanations derived by \toolname closely represent the RL policy with smaller model sizes.
In contrast with Fig.~\ref{fig:plot_ratio_model_size}, lower MSE does not necessarily mean higher performance due to potential model drift, where small deviations compound over time and lead to different trajectories.

\input{legible}

\paragraph{Interpretability.} As shown in Fig.~\ref{fig:plot_ratio_model_size}, \toolname models are smaller, strongly suggesting better interpretability. As a further interpretability analysis, and leaving a human survey for future work, we adopted the LLM-as-a-Judge framework. Based on the principles and checklist of \cite{contrerasolivas2025}, we selected 7 explainability questions relevant to feature-importance methods beyond performance, added 8 domain-specific questions for rule- and DT-based control models, and a final verdict question. We omitted performance values so LLMs focused solely on explainability. Six LLM judges rated each surrogate on a 1–5 Likert scale using a rubric, with thinking mode enabled when available. We compared \toolname with DTs, which are typically considered interpretable, and Cubist, its closest competitor from earlier results. Table~\ref{tab:results_llms} reports average scores and verdict rates across the 6 judges and 15 questions (details in App.~\appref{sec:appLLM}). \toolname outperformed DTs given comparable performance, and outperformed Cubist in all environments except \hopper, where they tied due to its small model (20 rules per action dimension), but Cubist showed worse performance. We thus have evidence for \textbf{RQ4} that \toolname rules are usually more interpretable than other methods with comparable performance.

%% file: images/plot_ratio_model_size.tex
\begin{tikzpicture}[
    font=\sffamily\footnotesize,
    bar/.style={minimum height=0.308cm, rounded corners=1pt},
    label/.style={anchor=center, inner sep=1pt},
    scale bar/.style={anchor=center, inner sep=1pt, font=\LARGE},
    title/.style={font=\sffamily\bfseries, anchor=center},
    ci/.style={line width=1.2pt, color=gray, cap=round}
]

\definecolor{color_orcaid}{RGB}{31, 119, 180}  
\definecolor{color_dt}{RGB}{255, 127, 14}      
\definecolor{color_cubist}{RGB}{44, 160, 44}   
\definecolor{color_rulefit}{RGB}{214, 39, 40}  

\def\models{{"Orcaid","Orcaid","Orcaid","Orcaid","Orcaid","Orcaid","Orcaid","Orcaid","Orcaid","DT","DT","DT","DT","DT","DT","DT","DT","DT","Cubist","Cubist","Cubist","Cubist","Cubist","Cubist","Cubist","Cubist","Cubist","RuleFit","RuleFit","RuleFit","RuleFit","RuleFit","RuleFit","RuleFit","RuleFit","RuleFit"}}
\def\envs{{"MC","P","LL","IP","IDP","R","S","H","HC","MC","P","LL","IP","IDP","R","S","H","HC","MC","P","LL","IP","IDP","R","S","H","HC","MC","P","LL","IP","IDP","R","S","H","HC"}}
\def\ratios{{0.98,0.923,0.904,0.842,1.027,0.797,0.9,0.899,0.841,0.993,0.931,0.389,0.85,0.054,0.562,0.95,0.175,0.473,0.97,0.951,0.883,0.819,0.687,0.808,1.011,0.859,0.847,0.823,0.595,0.793,0.779,0.07,0.416,0.873,0.848,0.459}}
\def\ratiosperc{{98,92,90,84,103,80,90,90,84,99,93,39,85,5,56,95,17,47,97,95,88,82,69,81,101,86,85,82,60,79,78,7,42,87,85,46}}
\def\ciratiolower{{0.966,0.905,0.8,0.606,1.006,0.776,0.852,0.864,0.795,0.985,0.915,0.235,0.638,0.047,0.535,0.922,0.116,0.384,0.931,0.933,0.798,0.546,0.567,0.783,0.99,0.776,0.818,0.709,0.578,0.634,0.635,0.043,0.389,0.828,0.779,0.412}}
\def\ciratioupper{{0.995,0.942,1.007,1.078,1.049,0.817,0.947,0.954,0.886,1.001,0.947,0.542,1.062,0.061,0.589,0.977,0.233,0.561,1.01,0.969,0.967,1.092,0.827,0.833,1.031,0.912,0.876,0.936,0.612,0.952,0.922,0.097,0.444,0.917,0.917,0.506}}

\def\modelsizes{{1.38,1.77,2.5,1.29,2.16,2.66,1.79,2.86,3.51,2.27,3.21,3.61,2.93,3.62,3.52,3.38,3.63,3.78,1.69,2.4,3.06,1.43,3.2,3.68,3.44,2.93,3.88,1.78,1.64,2.14,1.78,1.97,2.33,2.04,2.46,2.86}}
\def\modelsizesabs{{24,59,316,19,146,457,62,720,3215,185,1625,4099,848,4147,3328,2384,4238,6032,49,250,1161,27,1595,4820,2761,850,7500,61,43,137,61,94,212,109,287,721}}
\def\cisizelower{{1.35,1.74,2.48,1.21,2.14,2.65,1.75,2.84,3.46,2.25,3.19,3.59,2.87,3.6,3.51,3.34,3.61,3.75,1.68,2.39,3.06,1.42,3.2,3.67,3.41,2.92,3.86,1.68,1.54,2.06,1.65,1.83,2.29,1.99,2.43,2.83}}
\def\cisizeupper{{1.42,1.8,2.51,1.35,2.19,2.67,1.83,2.87,3.55,2.29,3.23,3.63,2.98,3.64,3.53,3.41,3.64,3.81,1.7,2.4,3.07,1.45,3.21,3.69,3.47,2.94,3.89,1.87,1.72,2.2,1.89,2.08,2.36,2.08,2.48,2.89}}

\def\environments{{"MC","P","LL","IP","IDP","R","S","H","HC"}}

\def\numentries{36}
\def\bargap{0.1}
\def\envgap{0.065}
\def\barheight{0.22}
\def\leftwidth{5}    
\def\rightwidth{5}   
\def\labelwidth{2.5}
\def\centralgap{0.3}
\def\ciheight{0.08}

\foreach \envindex in {0,...,8} {
    \pgfmathsetmacro{\envname}{\environments[\envindex]}
    \def\entrycount{0}
    
    \pgfmathsetmacro{\ypos}{-(\envindex)*(\barheight+\bargap+\envgap)*4 + 0.35}
    
    \node[label, anchor=east, align=right, font=\LARGE] at (-6.7,\ypos-0.4) {\envname};

    \ifnum\envindex>0{
        \draw[gray!50, line width=0.5pt, dash pattern=on 2pt off 2pt] 
            (-5.5,\ypos+0.3) -- (5.5,\ypos+0.3);
    }
    \fi
    
    \foreach \i in {0,...,35} {
        \pgfmathsetmacro{\currentenv}{\envs[\i]}
        \pgfmathsetmacro{\currentmodel}{\models[\i]}
        \pgfmathsetmacro{\currentratio}{\ratios[\i]}
        \pgfmathsetmacro{\currentratioperc}{\ratiosperc[\i]}
        \pgfmathsetmacro{\currentsize}{\modelsizes[\i]}
        \pgfmathsetmacro{\currentsizeabs}{\modelsizesabs[\i]}

        \pgfmathsetmacro{\currentciratiolower}{\ciratiolower[\i]}
        \pgfmathsetmacro{\currentciratioupper}{\ciratioupper[\i]}
        \pgfmathsetmacro{\currentcisizelower}{\cisizelower[\i]}
        \pgfmathsetmacro{\currentcisizeupper}{\cisizeupper[\i]}

        \edef\tempmodel{\currentmodel}
        \ifx\currentenv\envname
            \pgfmathsetmacro{\groupy}{\ypos - \entrycount*(\barheight+\bargap)}
            
            \def\modelcolor{color_rulefit}
            \def\modeltransparency{30}
            
           \def\testorcaid{Orcaid}
            \ifx\tempmodel\testorcaid
                \def\modelcolor{color_orcaid}
                \def\modeltransparency{100}
            \else
                \def\testdt{DT}
                \ifx\tempmodel\testdt
                    \def\modelcolor{color_dt}
                \else
                    \def\testcubist{Cubist}
                    \ifx\tempmodel\testcubist
                        \def\modelcolor{color_cubist}
                    \fi
                \fi
            \fi

            \node[label, anchor=east, align=right, font=\large] at (-5.35,\groupy) {\currentsizeabs};
            
            \node[label, anchor=east, align=right, font=\large] at (6.2,\groupy) {\currentratioperc};

            \pgfmathsetmacro{\scaledratio}{\currentratio/1.2*\leftwidth}
            \node[bar, anchor=west, minimum width={\scaledratio cm}, 
                  fill=\modelcolor!\modeltransparency, opacity=0.8] at (\centralgap,\groupy) {};
                        
            \pgfmathsetmacro{\scaledsize}{\currentsize/4.5*\rightwidth}
            \node[bar, anchor=east, minimum width={\scaledsize cm}, 
                      preaction={fill=\modelcolor!\modeltransparency, opacity=0.8}, 
                  ] at (-\centralgap,\groupy) {};

            \pgfmathsetmacro{\ciratiolowerscaled}{\currentciratiolower/1.2*\leftwidth}
            \pgfmathsetmacro{\ciratioupperscaled}{\currentciratioupper/1.2*\leftwidth}
            
            \draw[ci, line width=0.8pt] 
                (\ciratiolowerscaled + \centralgap, \groupy - \ciheight/2) -- 
                (\ciratiolowerscaled + \centralgap, \groupy + \ciheight/2);
            \draw[ci, line width=0.8pt] 
                (\ciratioupperscaled + \centralgap, \groupy - \ciheight/2) -- 
                (\ciratioupperscaled + \centralgap, \groupy + \ciheight/2);
            
            \draw[ci, line width=0.6pt] 
                (\ciratiolowerscaled + \centralgap, \groupy) -- 
                (\ciratioupperscaled + \centralgap, \groupy);

            \pgfmathsetmacro{\cisizelowerscaled}{-\currentcisizelower/4.5*\rightwidth}
            \pgfmathsetmacro{\cisizeupperscaled}{-\currentcisizeupper/4.5*\rightwidth}
            
            \draw[ci, line width=0.8pt] 
                (\cisizelowerscaled - \centralgap, \groupy - \ciheight/2) -- 
                (\cisizelowerscaled - \centralgap, \groupy + \ciheight/2);
            \draw[ci, line width=0.8pt] 
                (\cisizeupperscaled - \centralgap, \groupy - \ciheight/2) -- 
                (\cisizeupperscaled - \centralgap, \groupy + \ciheight/2);
            
            \draw[ci, line width=0.6pt] 
                (\cisizelowerscaled - \centralgap, \groupy) -- 
                (\cisizeupperscaled - \centralgap, \groupy);
            
            \pgfmathsetmacro{\entrycount}{\entrycount+1}
            \xdef\entrycount{\entrycount}
        \fi
    }
}

\draw[black!30, line width=0.8pt] (0,0.5) -- (0,-13.4);

\draw[black!30, dashed, line width=0.8pt] (1.0/1.2*\leftwidth+\centralgap,1) -- (1.0/1.2*\leftwidth+\centralgap,-13.4);

\draw[->, line width=0.8pt] (\centralgap,1) -- (\leftwidth+\centralgap,1) node[midway, above, font=\LARGE] {Ratio (\%)};
\foreach \x/\label in {0/0, 0.25/25, 0.5/50, 0.75/75, 1.0/100} {
    \pgfmathsetmacro{\xpos}{\x/1.2*\leftwidth+\centralgap}
    \draw[gray] (\xpos,1.1) -- (\xpos,0.9);
    \node[scale bar, anchor=south] at (\xpos,0.6) {\normalsize \label};
}

\draw[->, line width=0.8pt] (-\centralgap,1) -- (-\rightwidth-\centralgap,1) node[midway, above, font=\LARGE] {Model Size (log)};
\foreach \x/\label in {0/0, 1/1, 2/2, 3/3, 4/4} {
    \pgfmathsetmacro{\xpos}{-\x/4.5*\rightwidth-\centralgap}
    \draw[gray] (\xpos,1.1) -- (\xpos,0.9);
    \node[scale bar, anchor=south] at (\xpos,0.6) {\normalsize $10^{\label}$};
}

\def\legx{-5}
\def\legy{2.07}

\def\modelcolor{color_orcaid}
\node[bar, anchor=west, minimum width={1 cm}, 
                  fill=\modelcolor!100, opacity=0.8] at (\legx, \legy) {};
\node[anchor=west] at (\legx+1, \legy) {\toolname};

\def\modelcolor{color_dt}
\node[bar, anchor=west, minimum width={1 cm}, 
                  fill=\modelcolor!30, opacity=0.8] at (\legx+2.72, \legy) {};
\node[anchor=west] at (\legx+3.74, \legy) {DT};

\def\modelcolor{color_cubist}
\node[bar, anchor=west, minimum width={1 cm}, 
                  fill=\modelcolor!30, opacity=0.8] at (\legx+4.8, \legy) {};
\node[anchor=west] at (\legx+5.8, \legy) {Cubist};

\def\modelcolor{color_rulefit}
\node[bar, anchor=west, minimum width={1 cm}, 
                  fill=\modelcolor!30, opacity=0.8] at (\legx+7.2, \legy) {};
\node[anchor=west] at (\legx+8.3, \legy) {RuleFit:};

\end{tikzpicture}

%% file: images/plot_MSEs.tex
\begin{tikzpicture}[
    font=\sffamily\footnotesize,
    bar/.style={minimum height=0.01cm, minimum width=0.25cm, rounded corners=1pt},
    label/.style={anchor=center, inner sep=1pt},
    scale bar/.style={anchor=center, inner sep=1pt, font=\LARGE},
    title/.style={font=\sffamily\bfseries, anchor=center},
    ci/.style={line width=1.5pt, color=gray, cap=round}
]

\definecolor{color_orcaid}{RGB}{31, 119, 180}  
\definecolor{color_dt}{RGB}{255, 127, 14}      
\definecolor{color_cubist}{RGB}{44, 160, 44}   
\definecolor{color_rulefit}{RGB}{214, 39, 40}  

\def\models{{"Orcaid","Orcaid","Orcaid","Orcaid","Orcaid","Orcaid","Orcaid","Orcaid","Orcaid","DT","DT","DT","DT","DT","DT","DT","DT","DT","Cubist","Cubist","Cubist","Cubist","Cubist","Cubist","Cubist","Cubist","Cubist","RuleFit","RuleFit","RuleFit","RuleFit","RuleFit","RuleFit","RuleFit","RuleFit","RuleFit"}}
\def\envs{{"MC","P","LL","IP","IDP","R","S","H","HC","MC","P","LL","IP","IDP","R","S","H","HC","MC","P","LL","IP","IDP","R","S","H","HC","MC","P","LL","IP","IDP","R","S","H","HC"}}
\def\mse{{17,24,71,15,11,3,59,51,50,20,37,102,7,44,7,58,56,72,22,21,64,8,11,2,36,47,43,36,104,100,13,44,8,65,77,81}}
\def\cilower{{10.1,21.7,68.7,12.7,12.0,3.0,49.8,48.7,48.7,10.8,34.7,99.7,4.7,39.7,7.0,48.8,51.4,72.0,15.1,18.7,61.7,5.7,9.7,2.0,29.1,43.7,37.4,24.5,97.1,95.4,8.4,40.7,8.0,55.8,72.4,78.7}}
\def\ciupper{{23.9,26.3,73.3,17.3,12.0,3.0,68.2,53.3,53.3,29.2,39.3,104.3,9.3,44.3,7.0,67.2,60.6,72.0,28.9,23.3,66.3,10.3,14.3,2.0,42.9,48.3,46.6,47.5,110.9,104.6,17.6,45.3,8.0,74.2,81.6,83.3}}

\def\environments{{"MC","P","LL","IP","IDP","R","S","H","HC"}}

\def\envgap{0.2}
\def\barwidth{0.3}
\def\centergap{0.7}
\def\maxmse{110}
\def\chartheight{2.5}
\def\labelwidth{1.5}
\def\modelgap{0.01}
\def\ciwidth{0.15}

\foreach \envindex in {0,...,8} {
    \pgfmathsetmacro{\envname}{\environments[\envindex]}
    \def\modelcount{0}
    
    \pgfmathsetmacro{\xpos}{\centergap+(\envindex)*(\barwidth*4+\modelgap*3+\envgap)}
    
    \node[label, anchor=north, align=center, font=\large] 
        at (\xpos+0.35, -0.2) {\envname};
    
    \ifnum\envindex>0{
        \draw[gray!50, line width=0.5pt, dash pattern=on 2pt off 2pt] 
            (\xpos-0.25, 0) -- (\xpos-0.25, \chartheight);
    }
    \fi
    
    \foreach \i in {0,...,35} {
        \pgfmathsetmacro{\currentenv}{\envs[\i]}
        \pgfmathsetmacro{\currentmodel}{\models[\i]}
        \pgfmathsetmacro{\currentmse}{\mse[\i]}
        \pgfmathsetmacro{\currentcilower}{\cilower[\i]}
        \pgfmathsetmacro{\currentciupper}{\ciupper[\i]}
        
        \edef\tempmodel{\currentmodel}
        \edef\tempenv{\currentenv}
        
        \ifx\tempenv\envname
            \pgfmathsetmacro{\groupx}{\xpos + \modelcount*(\barwidth+\modelgap)}
            
            \pgfmathsetmacro{\barheight}{\currentmse/\maxmse*\chartheight}

            \pgfmathsetmacro{\cilowerheight}{\currentcilower/\maxmse*\chartheight}
            \pgfmathsetmacro{\ciupperheight}{\currentciupper/\maxmse*\chartheight}
            
            \def\modelcolor{color_rulefit}
            \def\modeltransparency{30}
            
            \def\testorcaid{Orcaid}
            \ifx\tempmodel\testorcaid
                \def\modelcolor{color_orcaid}
                \def\modeltransparency{100}
            \else
                \def\testdt{DT}
                \ifx\tempmodel\testdt
                    \def\modelcolor{color_dt}
                \else
                    \def\testcubist{Cubist}
                    \ifx\tempmodel\testcubist
                        \def\modelcolor{color_cubist}
                    \else
                        \def\modelcolor{color_rulefit}
                    \fi
                \fi
            \fi

            \node[bar, anchor=south, minimum width={\barwidth cm}, 
                  minimum height={\barheight cm}, 
                  fill=\modelcolor!\modeltransparency, opacity=0.8] 
                at (\groupx, 0) {};

            \draw[ci, line width=0.8pt] 
                (\groupx - \ciwidth/2, \ciupperheight) -- 
                (\groupx + \ciwidth/2, \ciupperheight);
            
            \draw[ci, line width=0.8pt] 
                (\groupx - \ciwidth/2, \cilowerheight) -- 
                (\groupx + \ciwidth/2, \cilowerheight);
            
            \draw[ci, line width=0.6pt] 
                (\groupx, \cilowerheight) -- 
                (\groupx, \ciupperheight);
            
            \pgfmathsetmacro{\modelcount}{\modelcount+1}
            \xdef\modelcount{\modelcount}
        \fi
    }
}

\draw[->, line width=0.8pt] (0,0) -- (0,\chartheight) 
    node[midway, above, sloped, font=\LARGE] {MSE\_test};
    
\foreach \y/\label in {0/0, \chartheight*0.25/25, \chartheight*0.5/50, \chartheight*0.75/75} {
    \draw[gray!30, line width=0.5pt] (-0.05,\y) -- (13.35,\y);
    \node[scale bar, anchor=south] at (0.25,\y) {\normalsize \label};
}

\def\legx{1}
\def\legy{\chartheight+0.3}

\def\modelcolor{color_orcaid}
\node[bar, anchor=west, minimum width={0.8 cm}, minimum height={0.4 cm},
      fill=\modelcolor!100, opacity=0.8] at (\legx,\legy) {};
\node[anchor=west] at (\legx+1,\legy) {\toolname};

\def\modelcolor{color_dt}
\node[bar, anchor=west, minimum width={0.8 cm}, minimum height={0.4 cm},
      fill=\modelcolor!30, opacity=0.8] at (\legx+2.6,\legy) {};
\node[anchor=west] at (\legx+3.6,\legy) {DT};

\def\modelcolor{color_cubist}
\node[bar, anchor=west, minimum width={0.8 cm}, minimum height={0.4 cm},
      fill=\modelcolor!30, opacity=0.8] at (\legx+5,\legy) {};
\node[anchor=west] at (\legx+6,\legy) {Cubist};

\def\modelcolor{color_rulefit}
\node[bar, anchor=west, minimum width={0.8 cm}, minimum height={0.4 cm},
      fill=\modelcolor!30, opacity=0.8] at (\legx+7.5,\legy) {};
\node[anchor=west] at (\legx+8.5,\legy) {RuleFit};

\end{tikzpicture}

%% file: legible.tex
\paragraph{Policy Improvement.}
 We adapt \legible \cite{DBLP:conf/ijcai/TapplerLTB25} to continuous state and action spaces. \legible\ transforms rules to related situations, e.g., symmetric situations, then locally overrides policy decisions with selected transformed rules, yielding a new policy $\pi'$. If $\pi'$ achieves a higher return than $\pi$, it exposes a weakness in $\pi$ and improves it. This testing approach may be ineffective for perfect policies, but as in model-based software testing, the quality of surrogate models matters; we therefore study which surrogates are most effective.
 
 To test for policy improvements, we (1) transform rules produced by surrogates by mirroring them along selected axes, and (2) enforce them by overriding RL policy decisions to identify weaknesses. We confine to symmetry, as this is the most relevant property in continuous environments. App.\ \appref{sec:app_transformation_improvement} includes the formalization of transformations, the complete results and an extra analysis.

Every surrogate was evaluated using four mirroring strategies (1 or 2 mirrored regions for 2 axes, and 1 mirrored region for 1 or 4 axes) with a random selection of $50$ transformations for each strategy. Table~\ref{tab:results_legible} shows how often the transformations led to better performance (\textbf{Rate}) and the maximum relative improvement w.r.t. the RL policy (\textbf{Best}). For Cubist and RuleFit, only single-action environments can be evaluated, since they build one model per action, which prevents mirroring across regions that contain different models. RL policies for \pendulum\ and \mountainCar\ are optimal and do not improve; thus, they are excluded. For the other environments, \toolname-based testing finds weaknesses more consistently than other methods, and the maximum improvement is higher or comparable. 

We can thus answer {\bf RQ3} positively as we could use \toolname models to improve the RL policies -- in general, more effectively than with other models.
Furthermore, as \toolname models are smaller, their mirrored rules contain fewer parameters. This makes it easier to understand both the mirrored region and how to improve the RL policy.

%% file: conclusion.tex
We presented \toolname, the first \emph{XRL method} for constructing interpretable rule-based surrogate models of deep RL policies with \emph{continuous actions over mixed continuous-discrete state spaces using oblique decisions}.  Empirically, \toolname achieves high fidelity and strong task performance while using few parameters, and it is typically more interpretable than standard DT (effectively a Q-value-free variant of VIPER), Cubist, and RuleFit at comparable performance. 

\noindent
\textbf{Discussion.} Our current evaluation spans tasks up to 
$17$-dimensional state and $6$-dimensional action spaces (\textbf{HC}); further scaling may require dimensionality reduction or feature selection. We model deterministic policies and collect data via environment queries (DAgger); this can limit fidelity for stochastic policies and risks model drift without query access, although the core oblique-tree learning and simplification steps remain applicable. Extending leaves with independent linear-Gaussian models per action dimension (diagonal covariance) is a natural next step toward supporting stochastic policies. Beyond quantitative metrics, we used an LLM-as-a-Judge protocol to assess interpretability; establishing agreement with human judgments remains important future work.

Our experiments suggest that rule models can match or improve the original policy, so they may be further improved by training them in the environment, i.e., re-optimization~\cite{DhebarDNZF24}.
Such models could be deployed in place of deep RL policies while offering transparency.
Finally, we plan to compose learned rules with environment models to verify compliance with requirements based on \cite{DBLP:conf/nips/0001LDL23}.

%% file: appendixDagger.tex
\begin{algorithm}[tb]
\caption{Dagger-like Rule Learning Loop}
\label{algo:dagger}
\small
\begin{algorithmic}[1]
\State $M_j \leftarrow \text{FitModel}(M_{j-1},X_{train}, y_{train}, \text{depth}=j)$
\State $\text{rew}_{prev} \leftarrow \text{Evaluate}(M_j)$

\While {true} \Comment{Trajectories collection and refinement}
    \State $D_{extra} \leftarrow \emptyset$
    \For{$1$ to \texttt{n\_episodes}}
        \State $\tau_{obs}, a_{model}, a_{expert} \leftarrow \text{CollectEpisode}(M_j)$ \Comment{Following the policy from $M_j$}
        \State $d \leftarrow \|a_{model} - a_{expert}\|_2^2$ \Comment{Distance between policies}
        \State $D_{extra} \leftarrow D_{extra} \cup \{\tau_{obs}[d >  \|a_{max}-a_{min}\|/2]\}$ \Comment{Add dissimilar samples}
    \EndFor
    \If{($D_{extra} = \emptyset$)}
        \State \textbf{break}
    \EndIf
    \State $X_{clust}, w_{clust} \leftarrow \text{Cluster}(D_{extra}, \pi_{expert})$ \Comment{Cluster data}
    \State $X, w \leftarrow \text{ResampleWithNoise}(X_{clust}, w_{clust})$ \Comment{Augment data with Gaussian noise}
    \State $Y \leftarrow \pi_{expert}(X_{clustered})$ \Comment{Get expert labels for new data}
    
    \State $(X, y, w)_{train}^+ \leftarrow (X, y, w)_{train} \cup (X, Y, w)$
    \State $M'_j \leftarrow \text{FitModel}(M_{j-1}, X_{train}^+, y_{train}^+, w_{train}^+,j)$
    \State $\text{rew} \leftarrow \text{Evaluate}(M'_j)$
    \If{($\text{rew} < \text{rew}_{prev}$)}
        \State \textbf{break}
    \EndIf
    \State $M_j \leftarrow M'_j$
    \State $(X, y, w)_{train} \leftarrow (X, y, w)_{train}^+$
    \State $\text{rew}_{prev} \leftarrow \text{rew}$
\EndWhile
\end{algorithmic}
\end{algorithm}

\section{Dagger-like Iterative Rule Learning}
\label{sec:appDagger}

Based on the Dataset Aggregation (DAgger) approach, we iteratively refine the model by collecting new data points that are annotated with the RL policy. We filter in the points at which the distilled model deviates from the original RL policy. This way, we iteratively build a policy that tries to learn from previous errors. 

Algorithm~\ref{algo:dagger} consists of two main stages for each iteration:

\begin{compactenum}[1.]
    \item \textbf{Episode collection and labeling}: New environment trajectories are generated using the current model $M_{j}$. The actions from this model are compared to the ones that the original RL agent would have taken in those states. The actions from the RL agent are considered expert demonstrations.
    
    \item \textbf{Model refinement}: The model is re-trained on an augmented dataset consisting of the original data plus the newly collected samples, where the action of the learned model deviates from the pre-trained RL agent. This refinement step is repeated until performance degrades or no more useful data can be collected.
\end{compactenum}

\subsection{Key Components}

\subsubsection{Trajectory Collection}
\begin{compactitem}[•]
    \item The current model $M_j$ is used to generate a trajectory in the environment.
    \item At each step, an observation $o_t$ is collected.
    \item The learned model's action $a^{M_j}_t = M_j(o_t)$ is computed and stored.
\end{compactitem}

\begin{compactitem}[•]
    \item The expert RL model is queried for the expert action $a^{\pi}_t = \pi_{expert}(o_t)$, which is also stored.
    \item The action $a^{M_j}_t$ is executed in the environment.
    \item The process repeats until the episode terminates.
\end{compactitem}

This generates a trajectory of observation-actions triples: $\{\dots, (o_t, a^{M_j}_t, a^{\pi}_t), \dots\}$.

\subsubsection{Data Filtering and Augmentation}
To ensure we do not add unnecessary data points to the training dataset, we include these steps:
\begin{compactenum}[1.]
    \item \textbf{Disagreement filtering:} For each point in the new trajectory, the squared Euclidean distance between the learned action and the expert action is computed:
    $$
    d(o_t) = \|a^{\pi}_t - a^{M_j}_t\|_2^2
    $$
    Only observations where this distance is significant ($d >  |\mathcal{A}|/2$) are kept, where $|\mathcal{A}|$ is the absolute difference between the maximum and the minimum possible actions $\|a_{max}-a_{min}\|$ (scalar values). This places the focus on the states where the RL policy and the learned model differ the most.
    \item \textbf{Clustering:} The filtered points are clustered to reduce redundancy and represent different regions of the state space.
    \item \textbf{Noise augmentation:} Each cluster is used as a sample, to which Gaussian noise is added. This improves model robustness and potential model drift.
\end{compactenum}

\subsubsection{Model fitting}
After adding the new data points to the training dataset, the model is fitted to this new dataset. The retraining process is designed to be efficient.

Instead of training from scratch until depth $j$, the algorithm starts from the previous model of depth $j-1$ by extending it by one depth. It extends that model, optimizing oblique splits as described in Section~\ref{sec:tree_building}. This increases the tree's depth to $j$. This only applies to \toolname.

\subsubsection{Termination}
The refinement loop contains an early stopping mechanism. After each refinement attempt, the average reward of the new model $M'_j$ is estimated.
\begin{compactitem}[•]
    \item If the average reward of $M'_j$ is worse than the previous model of the same depth $M_j$, the refinement is stopped.
    \item The algorithm discards $M'_j$ and reverts to the previous model $M_j$.
    \item This prevents the latest iteration from worsening the policy performance.
\end{compactitem}

%% file: appendix_legible.tex
\subsection{Method for Transformation-based Improvement \& Evaluation}
In our evaluation for \textbf{RQ3}, we present an application of \toolname-generated rules to evaluate and improve reinforcement learning (RL) policies. Following the approach proposed for rules with discrete conditions in~\cite{DBLP:conf/ijcai/TapplerLTB25}, we generate alternative rules $\mathcal{R}'$ by adapting rules $\mathcal{R}$ learned from a given policy. By enforcing $\mathcal{R}'$, that is, by overriding the decisions of a policy $\pi$ according to $\mathcal{R}'$, we can identify weaknesses in $\pi$: if enforcing $\mathcal{R}'$ leads to a higher return, this indicates that $\pi$ makes suboptimal decisions in the corresponding situations. The alternative rules $\mathcal{R}'$ capture counterfactual decisions, which we generate by exploiting domain knowledge, such as known symmetries in the environment.

This application further enhances the explainability provided by \toolname. Instead of solely constructing a rule-based surrogate model, \toolname-guided policy evaluation identifies a concise set of rules that pinpoint situations in which the evaluated policy makes suboptimal decisions. While a complete formalization of this evaluation approach is beyond the scope of this paper, we focus on exploiting domain symmetries by mirroring rules across selected axes. 

\paragraph{Rule Adaptation for \toolname. }
Let $c_i \Rightarrow \hat{\pi}_i$ be a rule $i$, where $c_i$ is a region and let $MI \subseteq \{1,\ldots,n_c\}$ be a set of indices. We define $\textsc{Mirror}(c_i \Rightarrow \hat{\pi}_i,MI) = m_i \Rightarrow \hat{\pi}_i$ as function mirroring the conditions $c_i$ across the axes given by $MI$. To this end, \textsc{Mirror} inverts the coefficients for these axes, i.e., for every $c_{\mathbf{w},b}$ in $c_i$, we create $c_{\mathbf{w'},b}$ such that for every $m\in MI$: $\mathbf{w'}[m]$ = $-\mathbf{w}[m]$. We do not invert $\hat{\pi}_i$ because we assume that if symmetries exist in an environment, the same linear model captures the required dependencies between states and actions. It suffices to mirror the conditions and apply the same local policy.
Note that we assume that $MI$ does not contain discrete features, because mirroring one-hot-encoded features generally does not make sense.

\paragraph{Rule Adaptation for DTs.}
For DTs, we consider individual paths from the root to a leaf. The function  $\textsc{MirrorDT}((C,l),MI)$ takes a list of axis-aligned conditions $C$ as input and a leaf-value $l$. It mirrors every $c \in C$ that applies to a feature with an index in $MI$, by inverting the relation and comparison value of $c$. Additionally, it multiplies $l$ by $-1$, as DTs predict a constant value rather than using a linear function. 

\paragraph{Rule Adaptation for CUBIST.}
For CUBIST, we combine the above transformation approaches. Like DT, CUBIST uses axis-aligned conditions; therefore, we mirror them analogously. As for \toolname, we leave the linear model for action prediction unchanged. 

\paragraph{Rule Adaptation for RuleFit.}
RuleFit uses axis-aligned rules, which we mirror in the same way as paths in DTs. However, RuleFit rules cannot be used on their own. For this reason, we create a tuple $(C,R)$ from mirroring RuleFit rules, $C$ is a list of mirrored conditions, and $R$ is a copy of the original RuleFit models transformed by replacing one list of rules with the mirrored list $C$.

\paragraph{Mirrored Rule Selection.}
Given a surrogate model, we generally cannot exhaustively mirror and test all possible rule-axis combinations, especially considering that multiple rules should be mirrored at once along multiple axes. For this reason, we apply a random selection approach parameterized by $rs$, $as$, and $nm$, where $rs$ specifies how many rules we select and mirror at once, $as$ specifies along how many axes we mirror at most, and $nm$ specifies the overall number of selections. For instance, if $rs = 2$ and $as = 2$, and $nm = 50,$ we randomly select $50$ pairs of rules and mirror them along $2$ or $1$ randomly selected axes. 

Note that for DTs, we consider paths in the tree, and we ensure that all selected axes-rule combinations make sense, i.e., that a rule includes conditions on the selected axes. While we use both DTs and rules, from now on, we simply refer to sets of rules. Let $\mathcal{MR}$ be a set of $nm$ mirrored rule sets.

\paragraph{Enforcing Rules.}
Given an RL policy $\pi$, and a rule set $\mathcal{R} \in \mathcal{MR}$, and state $s$, we define $\textsc{Enforce}(\pi,\mathcal{R},s)$ as follows: if there is an $r\in \mathcal{R}$ that triggers in $s$, i.e., its conditions evaluate to true, we $r$ to predict the action value and return it from \textsc{Enforce}. Otherwise, we return $\pi(s)$. If several rules trigger, we select one of them at random. In the case of RuleFit, we check for each rule if the corresponding conditions $C$ trigger and apply the transformed model $R$ if they do.

Note that by the rule construction for \toolname above, mirrored rule conditions do not overlap, which is harder to guarantee for the other models.

\paragraph{Rule-based Evaluation \& Improvement.}
We propose to combine, mirror, and enforce sets of rules to determine regions where the policy under consideration could be improved. 
To implement this form rule-based evaluation, we first select $\mathcal{MR}$ using the random selection described above. Then, we execute $n_{eval}$ episodes with $\pi$ and without any interference to collect samples of the average undiscounted return $G_\pi$. That is, we determine a baseline for the return.

After that, for every $\mathcal{R} \in \mathcal{MR}$, we perform another $n_{eval}$ episodes with $\pi$ while enforcing the rules in $\mathcal{R}$ via \textsc{Enforce} to collect samples of $G_{\pi,\mathcal{R}}$. Overall, we get $|\mathcal{MR}| = nm$ sets of samples. Based on these, we statistically test if and how many $\mathcal{R}$ improve the RL policy $\pi$. For this purpose, we use a variation of Dunnett's test~\cite{Dunnett}. Since the basic version of Dunnett's test for multiple comparisons assumes equal variances, we use a non-parametric bootstrapped version, which is a non-parametric adaptation of the test proposed by Alver and Zhang~\cite{pb_dunnett}. With this test, we determine for every $\mathcal{R}$ if it yields a significantly larger return than the baseline policy $\pi$, where we generally apply a significance threshold of $p=0.05$.

\subsection{Experiments}

The results reported in Table~\ref{tab:results_legible} are different than the ones from this section because the first ones were only conducted for a single model configuration (fixed maximum depth or number of rules). The ones in this section present are cumulative, increasing the depth or the number of rules. For example, for \invDoublePendulum, Table~\ref{tab:results_legible} shows no improvement for the baselines, but in this section, we find that improvements are possible with smaller models. This occurs when the mirrored regions are too small, which is equivalent to having large models.

Tables~\ref{tab:mirrored_orcaid_ll}-~\ref{tab:mirrored_dt_hc} present the results for this analysis. For each environment, we tested four mirroring configurations, with varying numbers of selected mirrored regions $rs$ and number of selected axes, i.e., features, $as$.
The tables are labeled at the top with ``comb-$as$ \& 
random-$rs$-$nm$'' to denote the mirroring configuration.
In each case, we set the number of mirroring evaluations to $nm=25$. For each evaluation with mirrored rules, we ran the environment for $n_{eval} = 500$ episodes in parallel, whereas for the base agent, we performed $n_{eval} =1,500$ episodes. A significant improvement is detected using a bootstrapped version of Dunnett's test as previously mentioned. That is why the base agent is run for more episodes.

To limit the number of tests, we followed an incremental approach. For a given $rs$ and a given $as$, we tested whether there was an improvement for the smallest model. If there was, we stopped; if there was not, we continued to the next bigger model by increasing the maximum depth (\toolname and DT) or the maximum number of rules (Cubist and RuleFit). The results are therefore cumulative. Rates indicate that an improvement was found at that maximum depth or maximum number of rules, or at a lower one. The best improvement follows the same idea.

We can see that for many environments (\invPendulum, \swimmer, \hopper), an \toolname model of depth 1 often finds a weakness. Recall that an \toolname model with depth 1 divides the state space into just two parts. That is, enforcing a mirrored rule in one part of the space and letting the RL policy decide actions in the other part improves performance compared to having the RL policy for the entire state space. This also means that we do not need to learn a perfect \toolname model to improve the original RL policy.

Another example of this situation appears with Cubist for \invDoublePendulum\ (Table~\ref{tab:mirrored_cubist_idp}), where improvements are found with smaller depths than the one that maximizes the reward ratio.

The results for a similar analysis leading to the identification of a weakness could help a user in, for example, the following two ways:
\begin{compactenum}[1.]
    \item A user could enforce the mirrored rule that improves the RL policy.
    \item A user could generate demonstrations that satisfy the rule and retrain the RL model.
\end{compactenum}

The following section shows how to understand the results with an example.

\begin{figure}
    \centering
    \begin{minipage}{.5\textwidth}
      \centering
      \includegraphics[width=0.6\linewidth]{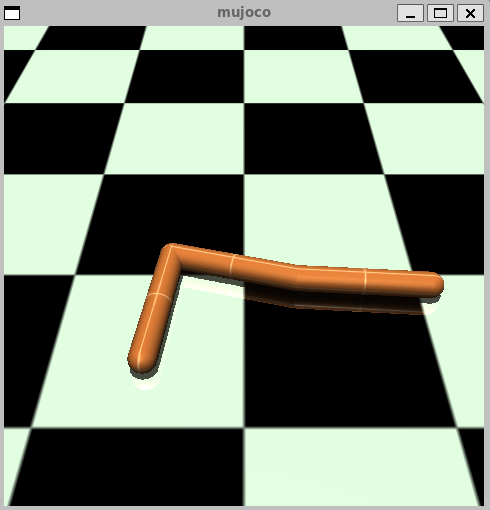}
      \captionof{figure}{Swimmer in Region 0 from the \toolname learned model with depth 1.}
        \label{fig:swimmer_original}
    \end{minipage}%
    \begin{minipage}{.5\textwidth}
      \centering
      \includegraphics[width=0.6\linewidth]{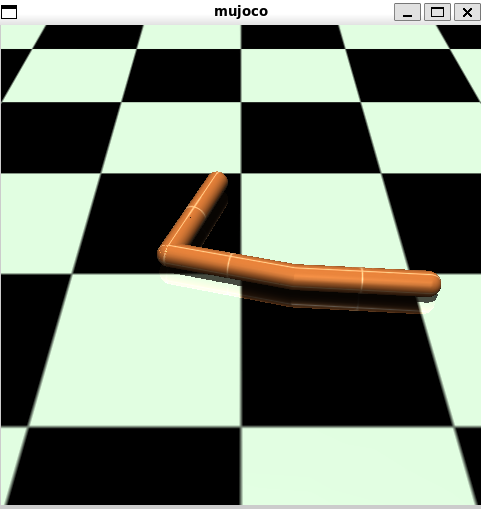}
        \captionof{figure}{Swimmer in a mirrored region from the \toolname learned model with depth 1.}
        \label{fig:swimmer_mirrored}
    \end{minipage}
\end{figure}

\subsection{Explanation example for Swimmer}

Table~\ref{tab:mirrored_orcaid_s} shows that weaknesses were often found with an \toolname model of depth 1. One of such models has the following condition $C$ to split the state space into two regions:

\begin{align*}
C \equiv &\ 0.2\cdot \mathit{motor1\_rot} +  0.7\cdot\mathit{motor2\_rot} \\
& + 0.2\cdot \mathit{slider1dot}  \\
& - 0.6\cdot\mathit{slider2dot} + 0.2\cdot\mathit{motor1\_rotdot} \geq 0.4  
\end{align*}

Region 0 satisfies $C$ and has the following torques:

\begin{minipage}[t]{.48\textwidth}
    \begin{align*}
    torque1 & = 0.7\cdot \mathit{free\_body\_rot} \\
    & +\! 0.4\!\cdot\!\mathit{motor1\_rot}\!-\!0.8\!\cdot\!\mathit{motor2\_rot}\\
    & + \mathit{slider1dot} + 0.5\cdot\mathit{slider2dot} \\
    & -0.4\cdot\mathit{free\_body\_rotdot}\\
    & + 0.4\cdot\mathit{motor1\_rotdot} - 1.4
    \end{align*}
\end{minipage}
\hfill
\begin{minipage}[t]{.48\textwidth}
    \begin{align*}
    torque2 = &-0.4\cdot\mathit{free\_body\_rot} + \mathit{motor1\_rot}  \\
    & - 2.1\cdot\mathit{motor2\_rot} + 1.1\cdot\mathit{slider1dot} \\
    & - 0.4\cdot\mathit{slider2dot} + 0.4
    \end{align*}
    \vfill
\end{minipage}

Region 1 does not satisfy $C$ and has the following torques:

\begin{minipage}[t]{.49\textwidth}
    \begin{align*}
    torque1 & =  0.3\cdot\mathit{free\_body\_rot} \\
    & - 0.8\cdot\mathit{motor1\_rot}  \\
    & + \mathit{motor2\_rot} - 0.6\cdot\mathit{slider1dot} + 0.4
    \end{align*}
\end{minipage}
\hfill
\begin{minipage}[t]{.49\textwidth}
    \begin{align*}
    torque2 & = 1.1\cdot\mathit{free\_body\_rot}  \\
    &\! -1.4\!\cdot\!\mathit{motor1\_rot}\! -\! 0.4\!\cdot\!\mathit{motor2\_rot} \\
    & + 0.9\cdot\mathit{slider1dot} - 0.1
    \end{align*}
\end{minipage}

The feature in $C$ with the highest positive weight is $\mathit{motor2\_rot}$ (angle of the second rotor), and the feature with the highest negative weight is $\mathit{slider2dot}$ (velocity of the tip along the y-axis). Thus, Region 0 corresponds to states in which the Swimmer is folded and has low upward velocity, possibly during the power stroke (thrust-generating phase). Fig.~\ref{fig:swimmer_original} shows this type of position when setting the state with our interactive tool (cf. App~\ref{sec:appTool}). The torque for this region is positive for both rotors, confirming that the Swimmer is generating thrust.

On the other hand, Region 1 covers states in which the Swimmer is extended with high upward motion, during a recovery stroke (reset phase). Here, therefore, the torques are negative, as one can see in their formulas.

Clearly, the Swimmer is a symmetric environment. Mirroring $C$ by the axis given by a feature (e.g., $\mathit{motor2\_rot}$), we get
\begin{align*}
C' \equiv &\ 0.2\cdot \mathit{motor1\_rot} \mathbf{-}  0.7\cdot\mathit{motor2\_rot} \\
& + 0.2\cdot \mathit{slider1dot}  \\
& - 0.6\cdot\mathit{slider2dot} + 0.2\cdot\mathit{motor1\_rotdot} \geq 0.4.
\end{align*}

As explained previously, the actions (torques) remain the same, as the linear regression should already capture variations in their inputs. In Fig.~\ref{fig:swimmer_mirrored}, we can now see what the Swimmer would look like when $C'$ is satisfied. It makes sense that this mirrored region corresponding to the satisfaction of $C'$ behaves the same way as the original Region 0. That is, this mirrored region should also correspond to a thrust-generating phase, since the states within it are the same; only the angle has been mirrored, without affecting the region's general purpose.

When we enforce this mirrored condition on the RL agent and observe a reward improvement, we can suspect that the RL agent has a weakness, since it has not fully learned this symmetry. Thus, we could try retraining with demonstrations that satisfy this condition. Also, we could incorporate this rule on top of the RL agent to enforce this behavior in production, and the user would know exactly how it behaves in this region. This would enhance the trust in the agent.

\clearpage

\begin{table}
        \scriptsize
        \centering
        \setlength{\tabcolsep}{1.7pt}
        \renewcommand{\arraystretch}{.4}
\parbox{.62\linewidth}{
\centering
\input{tables/rule_mirroring/DT_LL}
}
\parbox{.37\linewidth}{
\centering
\input{tables/rule_mirroring/orcaid_LL}
}
\end{table}

\begin{table}
        \scriptsize
        \centering
        \setlength{\tabcolsep}{1.7pt}
        \renewcommand{\arraystretch}{.4}
\parbox{.45\linewidth}{
\centering
\input{tables/rule_mirroring/orcaid_invPendulum}
}
\parbox{.5\linewidth}{
\centering
\input{tables/rule_mirroring/Cubist_invPendulum}
}
\end{table}

\begin{table}
        \scriptsize
        \centering
        \setlength{\tabcolsep}{1.7pt}
        \renewcommand{\arraystretch}{.4}
\parbox{.55\linewidth}{
\centering
\input{tables/rule_mirroring/DT_invPendulum}
}
\parbox{.42\linewidth}{
\centering
\input{tables/rule_mirroring/rulefit_invPendulum}
}
\end{table}

\begin{table}
        \scriptsize
        \centering
        \setlength{\tabcolsep}{1.7pt}
        \renewcommand{\arraystretch}{.4}
\parbox{.35\linewidth}{
\centering
\input{tables/rule_mirroring/orcaid_invDoublePendulum}
}
\parbox{.62\linewidth}{
\centering
\input{tables/rule_mirroring/Cubist_invDoublePendulum}
}
\end{table}

\begin{table}
        \scriptsize
        \centering
        \setlength{\tabcolsep}{1.7pt}
        \renewcommand{\arraystretch}{.4}
\parbox{.49\linewidth}{
\centering
\input{tables/rule_mirroring/DT_invDoublePendulum}
}
\parbox{.49\linewidth}{
\centering
\input{tables/rule_mirroring/rulefit_invDoublePendulum}
}
\end{table}

\begin{table}
        \scriptsize
        \centering
        \setlength{\tabcolsep}{1.7pt}
        \renewcommand{\arraystretch}{.4}
\parbox{.49\linewidth}{
\centering
\input{tables/rule_mirroring/orcaid_reacher}
}
\parbox{.49\linewidth}{
\centering
\input{tables/rule_mirroring/DT_reacher}
}
\end{table}

\begin{table}
        \scriptsize
        \centering
        \setlength{\tabcolsep}{1.7pt}
        \renewcommand{\arraystretch}{.4}
\parbox{.49\linewidth}{
\centering
\input{tables/rule_mirroring/orcaid_swimmer}
}
\parbox{.49\linewidth}{
\centering
\input{tables/rule_mirroring/DT_swimmer}
}
\end{table}

\begin{table}
        \scriptsize
        \centering
        \setlength{\tabcolsep}{1.7pt}
        \renewcommand{\arraystretch}{.4}
\parbox{.49\linewidth}{
\centering
\input{tables/rule_mirroring/orcaid_hopper}
}
\parbox{.49\linewidth}{
\centering
\input{tables/rule_mirroring/DT_hopper}
}
\end{table}

\begin{table}
        \scriptsize
        \centering
        \setlength{\tabcolsep}{1.7pt}
        \renewcommand{\arraystretch}{.4}
\parbox{.49\linewidth}{
\centering
\input{tables/rule_mirroring/orcaid_halfCheetah}
}
\parbox{.49\linewidth}{
\centering
\input{tables/rule_mirroring/DT_halfCheetah}
}
\end{table}

\clearpage

%% file: tables/rule_mirroring/DT_LL.tex
\begin{tabular}{l|cccccccccc}
\toprule
& \multicolumn{10}{c}{comb-1 \& random-1-25} \\\hline
Max. Depth & ${\leq}2$ & ${\leq}4$ & ${\leq}6$ & ${\leq}8$ & ${\leq}10$ & ${\leq}12$ & ${\leq}14$ & ${\leq}16$ & ${\leq}18$ & ${\leq}20$ \\\hline
Rate (\%) & 44 & 56 & 67 & 78 & 78 & 89 & 100 & -- & -- & -- \\
Best (\%) & 2.6 & 2.9 & 3.1 & 3.1 & 3.1 & 4.9 & 4.9 & -- & -- & -- \\
\hline
& \multicolumn{10}{c}{comb-1 \& random-2-25} \\\hline
Rate (\%) & 22 & 44 & 44 & 56 & 78 & 89 & 89 & 89 & 89 & 100 \\
Best (\%) & 3.1 & 3.4 & 3.4 & 3.4 & 4.1 & 4.9 & 4.9 & 4.9 & 4.9 & 4.9 \\
\hline
& \multicolumn{10}{c}{comb-2 \& random-2-25} \\\hline
Rate (\%) & 44 & 56 & 78 & 89 & 100 & -- & -- & -- & -- & -- \\
Best (\%) & 3.3 & 3.3 & 3.3 & 3.3 & 4.3 & -- & -- & -- & -- & -- \\
\hline
& \multicolumn{10}{c}{comb-1 \& random-4-25} \\\hline
Rate (\%) & 0 & 22 & 22 & 44 & 78 & 100 & -- & -- & -- & -- \\
Best (\%) & 0.0 & 3.1 & 3.1 & 3.9 & 4.1 & 4.1 & -- & -- & -- & -- \\
\bottomrule
\end{tabular}
\caption{Performance by Region and Index Selection for DT for \lunarLander.}
\label{tab:mirrored_dt_ll}

%% file: tables/rule_mirroring/orcaid_LL.tex
\begin{tabular}{l|ccccc}
\toprule
& \multicolumn{5}{c}{comb-1 \& random-1-25} \\\hline
Max. Depth & ${\leq}1$ & ${\leq}2$ & ${\leq}3$ & ${\leq}4$ & ${\leq}5$ \\\hline
Rate (\%) & 22 & 44 & 78 & 89 & 89 \\
Best (\%) & 3.0 & 4.9 & 5.8 & 5.8 & 5.8 \\
\hline
& \multicolumn{5}{c}{comb-1 \& random-2-25} \\\hline
Rate (\%) & 22 & 78 & 78 & 100 & -- \\
Best (\%) & 3.7 & 4.8 & 4.8 & 4.8 & -- \\
\hline
& \multicolumn{5}{c}{comb-2 \& random-2-25} \\\hline
Rate (\%) & 44 & 56 & 56 & 56 & 67 \\
Best (\%) & 4.1 & 4.1 & 4.1 & 4.1 & 4.1 \\
\hline
& \multicolumn{5}{c}{comb-1 \& random-4-25} \\\hline
Rate (\%) & 0 & 22 & 56 & 78 & 89 \\
Best (\%) & 0.0 & 2.8 & 8.2 & 8.2 & 8.2 \\
\bottomrule
\end{tabular}
\caption{Performance by Region and Index Selection for \toolname for \lunarLander.}
\label{tab:mirrored_orcaid_ll}

%% file: tables/rule_mirroring/orcaid_invPendulum.tex
\begin{tabular}{l|ccccc}
\toprule
& \multicolumn{5}{c}{comb-1 \& random-1-25} \\\hline
Max. Depth & ${\leq}1$ & ${\leq}2$ & ${\leq}3$ & ${\leq}4$ & ${\leq}5$ \\\hline
Rate (\%) & 22 & 33 & 33 & 33 & 33 \\
Best (\%) & 11 & 15 & 15 & 15 & 15 \\
\hline
& \multicolumn{5}{c}{comb-1 \& random-2-25} \\\hline
Rate (\%) & 22 & 33 & 33 & 33 & 33 \\
Best (\%) & 11 & 18 & 18 & 18 & 18 \\
\hline
& \multicolumn{5}{c}{comb-2 \& random-2-25} \\\hline
Rate (\%) & 22 & 22 & 22 & 22 & 22 \\
Best (\%) & 12 & 12 & 12 & 12 & 12 \\
\hline
& \multicolumn{5}{c}{comb-1 \& random-4-25} \\\hline
Rate (\%) & 22 & 22 & 22 & 22 & 22 \\
Best (\%) & 10 & 10 & 10 & 10 & 10 \\
\bottomrule
\end{tabular}
\caption{Performance by Region and Index Selection for \toolname for \invPendulum.}
\label{tab:mirrored_orcaid_ip}

%% file: tables/rule_mirroring/Cubist_invPendulum.tex
\begin{tabular}{l|ccccccccc}
\toprule
& \multicolumn{9}{c}{comb-1 \& random-1-25} \\\hline
Max.\#rules  & ${\leq}2$ & ${\leq}3$ & ${\leq}4$ & ${\leq}5$ & ${\leq}6$ & ${\leq}7$ & ${\leq}8$ & ${\leq}9$ & ${\leq}10$ \\\hline
Rate (\%) & 0 & 11 & 11 & 33 & 33 & 33 & 33 & 33 & 33 \\
Best (\%) & 0.0 & 20 & 20 & 24 & 24 & 24 & 24 & 24 & 24 \\
\hline
& \multicolumn{9}{c}{comb-1 \& random-2-25} \\\hline
Rate (\%) & 0 & 11 & 11 & 33 & 33 & 33 & 33 & 33 & 33 \\
Best (\%) & 0.0 & 21 & 21 & 24 & 24 & 24 & 24 & 24 & 24 \\
\hline
& \multicolumn{9}{c}{comb-2 \& random-2-25} \\\hline
Rate (\%) & 0 & 11 & 11 & 33 & 33 & 33 & 33 & 33 & 33 \\
Best (\%) & 0.0 & 19 & 19 & 24 & 24 & 24 & 24 & 24 & 24 \\
\hline
& \multicolumn{9}{c}{comb-1 \& random-4-25} \\\hline
Rate (\%) & 0 & 11 & 11 & 33 & 33 & 33 & 33 & 33 & 33 \\
Best (\%) & 0.0 & 21 & 21 & 24 & 24 & 24 & 24 & 24 & 24 \\
\bottomrule
\end{tabular}
\caption{Performance by Region and Index Selection for Cubist for \invPendulum.}
\label{tab:mirrored_cubist_ip}

%% file: tables/rule_mirroring/DT_invPendulum.tex
\begin{tabular}{l|cccccccccc}
\toprule
& \multicolumn{10}{c}{comb-1 \& random-1-25} \\\hline
Max. Depth & ${\leq}1$ & ${\leq}2$ & ${\leq}3$ & ${\leq}4$ & ${\leq}5$ & ${\leq}6$ & ${\leq}7$ & ${\leq}8$ & ${\leq}9$ & ${\leq}10$ \\\hline
Rate (\%) & 0 & 0 & 0 & 0 & 11 & 11 & 22 & 22 & 22 & 22 \\
Best (\%) & 0.0 & 0.0 & 0.0 & 0.0 & 17 & 17 & 17 & 17 & 17 & 17 \\
\hline
& \multicolumn{10}{c}{comb-1 \& random-2-25} \\\hline
Rate (\%) & 0 & 0 & 0 & 0 & 11 & 11 & 22 & 22 & 33 & 33 \\
Best (\%) & 0.0 & 0.0 & 0.0 & 0.0 & 13 & 13 & 19 & 19 & 19 & 19 \\
\hline
& \multicolumn{10}{c}{comb-2 \& random-2-25} \\\hline
Rate (\%) & 0 & 0 & 0 & 0 & 11 & 22 & 22 & 22 & 22 & 22 \\
Best (\%) & 0.0 & 0.0 & 0.0 & 0.0 & 16 & 19 & 19 & 19 & 19 & 19 \\
\hline
& \multicolumn{10}{c}{comb-1 \& random-4-25} \\\hline
Rate (\%) & 0 & 0 & 0 & 0 & 22 & 22 & 22 & 22 & 33 & 33 \\
Best (\%) & 0.0 & 0.0 & 0.0 & 0.0 & 21 & 21 & 21 & 21 & 21 & 21 \\
\bottomrule
\end{tabular}
\caption{Performance by Region and Index Selection for DT for \invPendulum.}
\label{tab:mirrored_dt_ip}

%% file: tables/rule_mirroring/rulefit_invPendulum.tex
\begin{tabular}{l|ccccccc}
\toprule
& \multicolumn{7}{c}{comb-1 \& random-1-25} \\\hline
Max.\#rules  & ${\leq}2$ & ${\leq}4$ & ${\leq}6$ & ${\leq}8$ & ${\leq}10$ & ${\leq}12$ & ${\leq}14$ \\\hline
Rate (\%) & 25 & 25 & 25 & 25 & 50 & 50 & 50 \\
Best (\%) & 8.5 & 8.5 & 8.5 & 8.5 & 13 & 13 & 13 \\
\hline
& \multicolumn{7}{c}{comb-1 \& random-2-25} \\\hline
Rate (\%) & 0 & 0 & 0 & 0 & 25 & 50 & 50 \\
Best (\%) & 0.0 & 0.0 & 0.0 & 0.0 & 17 & 17 & 17 \\
\hline
& \multicolumn{7}{c}{comb-2 \& random-2-25} \\\hline
Rate (\%) & 0 & 0 & 0 & 0 & 0 & 0 & 25 \\
Best (\%) & 0.0 & 0.0 & 0.0 & 0.0 & 0.0 & 0.0 & 14 \\
\hline
& \multicolumn{7}{c}{comb-1 \& random-4-25} \\\hline
Rate (\%) & 0 & 0 & 0 & 0 & 0 & 0 & 25 \\
Best (\%) & 0.0 & 0.0 & 0.0 & 0.0 & 0.0 & 0.0 & 6.9 \\
\bottomrule
\end{tabular}
\caption{Performance by Region and Index Selection for RuleFit for \invPendulum.}
\label{tab:mirrored_rulefit_ip}

%% file: tables/rule_mirroring/orcaid_invDoublePendulum.tex
\begin{tabular}{l|cccc}
\toprule
& \multicolumn{4}{c}{comb-1 \& random-1-25} \\\hline
Max. Depth & ${\leq}1$ & ${\leq}2$ & ${\leq}3$ & ${\leq}4$ \\\hline
Rate (\%) & 0 & 33 & 50 & 50 \\
Best (\%) & 0.0 & 4.6 & 4.6 & 4.6 \\
\hline
& \multicolumn{4}{c}{comb-1 \& random-2-25} \\\hline
Rate (\%) & 0 & 33 & 50 & 50 \\
Best (\%) & 0.0 & 4.8 & 4.8 & 4.8 \\
\hline
& \multicolumn{4}{c}{comb-2 \& random-2-25} \\\hline
Rate (\%) & 0 & 0 & 67 & 67 \\
Best (\%) & 0.0 & 0.0 & 5.0 & 5.0 \\
\hline
& \multicolumn{4}{c}{comb-1 \& random-4-25} \\\hline
Rate (\%) & 0 & 0 & 50 & 50 \\
Best (\%) & 0.0 & 0.0 & 4.0 & 4.0 \\
\bottomrule
\end{tabular}
\caption{Performance by Region and Index Selection for \toolname for \invDoublePendulum.}
\label{tab:mirrored_orcaid_idp}

%% file: tables/rule_mirroring/Cubist_invDoublePendulum.tex
\begin{tabular}{l|cccccccccc}
\toprule
& \multicolumn{10}{c}{comb-1 \& random-1-25} \\\hline
Max.\#rules & ${\leq}10$ & ${\leq}20$ & ${\leq}40$ & ${\leq}50$ & ${\leq}70$ & ${\leq}80$ & ${\leq}100$ & ${\leq}110$ & ${\leq}130$ & ${\leq}150$ \\\hline
Rate (\%) & 17 & 33 & 33 & 50 & 67 & 83 & 83 & 83 & 83 & 83 \\
Best (\%) & 3.4 & 5.0 & 5.0 & 5.0 & 5.0 & 5.0 & 5.0 & 5.0 & 5.0 & 5.0 \\
\hline
& \multicolumn{10}{c}{comb-1 \& random-2-25} \\\hline
Rate (\%) & 17 & 17 & 33 & 50 & 50 & 50 & 67 & 67 & 67 & 67 \\
Best (\%) & 2.8 & 2.8 & 4.8 & 4.8 & 4.8 & 4.8 & 4.8 & 4.8 & 4.8 & 4.8 \\
\hline
& \multicolumn{10}{c}{comb-2 \& random-2-25} \\\hline
Rate (\%) & 17 & 33 & 33 & 50 & 50 & 50 & 50 & 67 & 67 & 67 \\
Best (\%) & 3.2 & 5.0 & 5.0 & 5.0 & 5.0 & 5.0 & 5.0 & 5.0 & 5.0 & 5.0 \\
\hline
& \multicolumn{10}{c}{comb-1 \& random-4-25} \\\hline
Rate (\%) & 17 & 33 & 33 & 33 & 33 & 50 & 50 & 67 & 67 & 67 \\
Best (\%) & 3.7 & 5.0 & 5.0 & 5.0 & 5.0 & 5.0 & 5.0 & 5.0 & 5.0 & 5.0 \\
\bottomrule
\end{tabular}
\caption{Performance by Region and Index Selection for Cubist for \invDoublePendulum.}
\label{tab:mirrored_cubist_idp}

%% file: tables/rule_mirroring/DT_invDoublePendulum.tex
\begin{tabular}{l|cccccccccc}
\toprule
& \multicolumn{10}{c}{comb-1 \& random-1-25} \\\hline
Max. Depth & ${\leq}2$ & ${\leq}4$ & ${\leq}6$ & ${\leq}8$ & ${\leq}10$ & ${\leq}12$ & ${\leq}14$ & ${\leq}16$ & ${\leq}18$ & ${\leq}20$ \\\hline
Rate (\%) & 0 & 0 & 0 & 17 & 17 & 17 & 17 & 17 & 17 & 17 \\
Best (\%) & 0.0 & 0.0 & 0.0 & 2.8 & 2.8 & 2.8 & 2.8 & 2.8 & 2.8 & 2.8 \\
\hline
& \multicolumn{10}{c}{comb-1 \& random-2-25} \\\hline
Rate (\%) & 0 & 0 & 0 & 0 & 0 & 0 & 17 & 17 & 17 & 17 \\
Best (\%) & 0.0 & 0.0 & 0.0 & 0.0 & 0.0 & 0.0 & 2.8 & 2.8 & 2.8 & 2.8 \\
\hline
& \multicolumn{10}{c}{comb-2 \& random-2-25} \\\hline
Rate (\%) & 0 & 0 & 0 & 0 & 0 & 0 & 0 & 0 & 17 & 17 \\
Best (\%) & 0.0 & 0.0 & 0.0 & 0.0 & 0.0 & 0.0 & 0.0 & 0.0 & 3.2 & 3.2 \\
\hline
& \multicolumn{10}{c}{comb-1 \& random-4-25} \\\hline
Rate (\%) & 0 & 0 & 0 & 0 & 0 & 0 & 0 & 0 & 0 & 0 \\
Best (\%) & 0.0 & 0.0 & 0.0 & 0.0 & 0.0 & 0.0 & 0.0 & 0.0 & 0.0 & 0.0 \\
\bottomrule
\end{tabular}
\caption{Performance by Region and Index Selection for DT for \invDoublePendulum.}
\label{tab:mirrored_dt_idp}

%% file: tables/rule_mirroring/rulefit_invDoublePendulum.tex
\begin{tabular}{l|ccccccccc}
\toprule
& \multicolumn{9}{c}{comb-1 \& random-1-25} \\\hline
Max.\#rules  & ${\leq}3$ & ${\leq}6$ & ${\leq}9$ & ${\leq}12$ & ${\leq}15$ & ${\leq}18$ & ${\leq}21$ & ${\leq}24$ & ${\leq}27$ \\\hline
Rate (\%) & 0 & 0 & 67 & 67 & 67 & 67 & 67 & 67 & 67 \\
Best (\%) & 0.0 & 0.0 & 2.1 & 2.1 & 2.1 & 2.1 & 2.1 & 2.1 & 2.1 \\
\hline
& \multicolumn{9}{c}{comb-1 \& random-2-25} \\\hline
Rate (\%) & 33 & 33 & 33 & 33 & 67 & 67 & 67 & 67 & 67 \\
Best (\%) & 2.1 & 2.1 & 2.1 & 2.1 & 2.2 & 2.2 & 2.2 & 2.2 & 2.2 \\
\hline
& \multicolumn{9}{c}{comb-2 \& random-2-25} \\\hline
Rate (\%) & 33 & 33 & 33 & 33 & 33 & 33 & 33 & 33 & 33 \\
Best (\%) & 2.1 & 2.1 & 2.1 & 2.1 & 2.1 & 2.1 & 2.1 & 2.1 & 2.1 \\
\hline
& \multicolumn{9}{c}{comb-1 \& random-4-25} \\\hline
Rate (\%) & 0 & 0 & 0 & 0 & 0 & 0 & 0 & 0 & 0 \\
Best (\%) & 0.0 & 0.0 & 0.0 & 0.0 & 0.0 & 0.0 & 0.0 & 0.0 & 0.0 \\
\bottomrule
\end{tabular}
\caption{Performance by Region and Index Selection for RuleFit for \invDoublePendulum.}
\label{tab:mirrored_rulefit_idp}

%% file: tables/rule_mirroring/orcaid_reacher.tex
\begin{tabular}{l|ccccc}
\toprule
& \multicolumn{5}{c}{comb-1 \& random-1-25} \\\hline
Max. Depth & ${\leq}1$ & ${\leq}2$ & ${\leq}3$ & ${\leq}4$ & ${\leq}5$ \\\hline
Rate (\%) & 0 & 0 & 0 & 0 & 0 \\
Best (\%) & 0.0 & 0.0 & 0.0 & 0.0 & 0.0 \\
\hline
& \multicolumn{5}{c}{comb-1 \& random-2-25} \\\hline
Rate (\%) & 0 & 0 & 0 & 0 & 0 \\
Best (\%) & 0.0 & 0.0 & 0.0 & 0.0 & 0.0 \\
\hline
& \multicolumn{5}{c}{comb-2 \& random-2-25} \\\hline
Rate (\%) & 0 & 0 & 0 & 0 & 0 \\
Best (\%) & 0.0 & 0.0 & 0.0 & 0.0 & 0.0 \\
\hline
& \multicolumn{5}{c}{comb-1 \& random-4-25} \\\hline
Rate (\%) & 0 & 0 & 0 & 0 & 0 \\
Best (\%) & 0.0 & 0.0 & 0.0 & 0.0 & 0.0 \\
\bottomrule
\end{tabular}
\caption{Performance by Region and Index Selection for \toolname for \reacher.}
\label{tab:mirrored_orcaid_r}

%% file: tables/rule_mirroring/DT_reacher.tex
\begin{tabular}{l|cccccccccc}
\toprule
& \multicolumn{10}{c}{comb-1 \& random-1-25} \\\hline
Max. Depth & ${\leq}1$ & ${\leq}3$ & ${\leq}5$ & ${\leq}7$ & ${\leq}9$ & ${\leq}11$ & ${\leq}13$ & ${\leq}15$ & ${\leq}17$ & ${\leq}20$ \\\hline
Rate (\%) & 0 & 0 & 0 & 0 & 0 & 0 & 0 & 0 & 0 & 0 \\
Best (\%) & 0.0 & 0.0 & 0.0 & 0.0 & 0.0 & 0.0 & 0.0 & 0.0 & 0.0 & 0.0 \\
\hline
& \multicolumn{10}{c}{comb-1 \& random-2-25} \\\hline
Rate (\%) & 0 & 0 & 0 & 0 & 0 & 0 & 0 & 0 & 0 & 0 \\
Best (\%) & 0.0 & 0.0 & 0.0 & 0.0 & 0.0 & 0.0 & 0.0 & 0.0 & 0.0 & 0.0 \\
\hline
& \multicolumn{10}{c}{comb-2 \& random-2-25} \\\hline
Rate (\%) & 0 & 0 & 0 & 0 & 0 & 0 & 0 & 0 & 0 & 0 \\
Best (\%) & 0.0 & 0.0 & 0.0 & 0.0 & 0.0 & 0.0 & 0.0 & 0.0 & 0.0 & 0.0 \\
\hline
& \multicolumn{10}{c}{comb-1 \& random-4-25} \\\hline
Rate (\%) & 0 & 0 & 0 & 0 & 0 & 0 & 0 & 0 & 0 & 0 \\
Best (\%) & 0.0 & 0.0 & 0.0 & 0.0 & 0.0 & 0.0 & 0.0 & 0.0 & 0.0 & 0.0 \\
\bottomrule
\end{tabular}
\caption{Performance by Region and Index Selection for DT for \reacher.}
\label{tab:mirrored_dt_r}

%% file: tables/rule_mirroring/orcaid_swimmer.tex
\begin{tabular}{l|ccccc}
\toprule
& \multicolumn{5}{c}{comb-1 \& random-1-25} \\\hline
Max. Depth & ${\leq}1$ & ${\leq}2$ & ${\leq}3$ & ${\leq}4$ & ${\leq}5$ \\\hline
Rate (\%) & 22 & 67 & 78 & 89 & 100 \\
Best (\%) & 2.5 & 2.5 & 2.5 & 2.5 & 2.5 \\
\hline
& \multicolumn{5}{c}{comb-1 \& random-2-25} \\\hline
Rate (\%) & 22 & 56 & 67 & 89 & 89 \\
Best (\%) & 2.5 & 2.5 & 2.5 & 2.5 & 2.5 \\
\hline
& \multicolumn{5}{c}{comb-2 \& random-2-25} \\\hline
Rate (\%) & 11 & 44 & 67 & 67 & 67 \\
Best (\%) & 0.3 & 1.5 & 1.5 & 1.5 & 1.5 \\
\hline
& \multicolumn{5}{c}{comb-1 \& random-4-25} \\\hline
Rate (\%) & 0 & 44 & 67 & 89 & 89 \\
Best (\%) & 0.0 & 1.5 & 2.1 & 2.1 & 2.1 \\
\bottomrule
\end{tabular}
\caption{Performance by Region and Index Selection for \toolname for \swimmer.}
\label{tab:mirrored_orcaid_s}

%% file: tables/rule_mirroring/DT_swimmer.tex
\begin{tabular}{l|cccccccccc}
\toprule
& \multicolumn{10}{c}{comb-1 \& random-1-25} \\\hline
Max. Depth & ${\leq}1$ & ${\leq}2$ & ${\leq}3$ & ${\leq}4$ & ${\leq}5$ & ${\leq}6$ & ${\leq}7$ & ${\leq}8$ & ${\leq}9$ & ${\leq}10$ \\\hline
Rate (\%) & 0 & 0 & 33 & 33 & 44 & 56 & 78 & 89 & 89 & 89 \\
Best (\%) & 0.0 & 0.0 & 1.1 & 1.1 & 1.1 & 2.0 & 2.0 & 2.0 & 2.0 & 2.0 \\
\hline
& \multicolumn{10}{c}{comb-1 \& random-2-25} \\\hline
Rate (\%) & 0 & 0 & 33 & 33 & 33 & 44 & 78 & 78 & 78 & 89 \\
Best (\%) & 0.0 & 0.0 & 1.1 & 1.1 & 1.1 & 1.1 & 1.1 & 1.1 & 1.1 & 1.1 \\
\hline
& \multicolumn{10}{c}{comb-2 \& random-2-25} \\\hline
Rate (\%) & 0 & 0 & 22 & 22 & 33 & 56 & 78 & 89 & 89 & 100 \\
Best (\%) & 0.0 & 0.0 & 1.2 & 1.2 & 1.2 & 1.2 & 1.6 & 1.6 & 1.6 & 2.7 \\
\hline
& \multicolumn{10}{c}{comb-1 \& random-4-25} \\\hline
Rate (\%) & 0 & 0 & 33 & 33 & 56 & 78 & 78 & 89 & 89 & 89 \\
Best (\%) & 0.0 & 0.0 & 1.2 & 1.2 & 1.2 & 1.2 & 1.2 & 1.2 & 1.2 & 1.2 \\
\bottomrule
\end{tabular}
\caption{Performance by Region and Index Selection for DT for \swimmer.}
\label{tab:mirrored_dt_s}

%% file: tables/rule_mirroring/orcaid_hopper.tex
\begin{tabular}{l|ccccc}
\toprule
& \multicolumn{5}{c}{comb-1 \& random-1-25} \\\hline
Max. Depth & ${\leq}1$ & ${\leq}2$ & ${\leq}3$ & ${\leq}4$ & ${\leq}5$ \\\hline
Rate (\%) & 67 & 89 & 100 & -- & -- \\
Best (\%) & 5.8 & 5.8 & 5.8 & -- & -- \\
\hline
& \multicolumn{5}{c}{comb-1 \& random-2-25} \\\hline
Rate (\%) & 56 & 78 & 100 & -- & -- \\
Best (\%) & 3.6 & 3.6 & 3.6 & -- & -- \\
\hline
& \multicolumn{5}{c}{comb-2 \& random-2-25} \\\hline
Rate (\%) & 33 & 67 & 89 & 89 & 100 \\
Best (\%) & 2.8 & 4.3 & 4.3 & 4.3 & 4.3 \\
\hline
& \multicolumn{5}{c}{comb-1 \& random-4-25} \\\hline
Rate (\%) & 11 & 56 & 78 & 100 & -- \\
Best (\%) & 2.8 & 4.6 & 4.6 & 4.6 & -- \\
\bottomrule
\end{tabular}
\caption{Performance by Region and Index Selection for \toolname for \hopper.}
\label{tab:mirrored_orcaid_h}

%% file: tables/rule_mirroring/DT_hopper.tex
\begin{tabular}{l|cccccccccc}
\toprule
& \multicolumn{10}{c}{comb-1 \& random-1-25} \\\hline
Max. Depth & ${\leq}1$ & ${\leq}2$ & ${\leq}3$ & ${\leq}4$ & ${\leq}5$ & ${\leq}6$ & ${\leq}7$ & ${\leq}8$ & ${\leq}9$ & ${\leq}10$ \\\hline
Rate (\%) & 0 & 0 & 33 & 44 & 44 & 78 & 89 & 89 & 100 & -- \\
Best (\%) & 0.0 & 0.0 & 5.9 & 5.9 & 5.9 & 9.1 & 9.1 & 9.1 & 9.1 & -- \\
\hline
& \multicolumn{10}{c}{comb-1 \& random-2-25} \\\hline
Rate (\%) & 0 & 0 & 11 & 56 & 78 & 89 & 89 & 100 & -- & -- \\
Best (\%) & 0.0 & 0.0 & 4.9 & 4.9 & 4.9 & 6.5 & 6.5 & 6.5 & -- & -- \\
\hline
& \multicolumn{10}{c}{comb-2 \& random-2-25} \\\hline
Rate (\%) & 0 & 0 & 33 & 56 & 67 & 89 & 89 & 89 & 89 & 100 \\
Best (\%) & 0.0 & 0.0 & 5.9 & 5.9 & 5.9 & 5.9 & 5.9 & 5.9 & 5.9 & 5.9 \\
\hline
& \multicolumn{10}{c}{comb-1 \& random-4-25} \\\hline
Rate (\%) & 0 & 0 & 11 & 56 & 56 & 78 & 78 & 100 & -- & -- \\
Best (\%) & 0.0 & 0.0 & 3.9 & 5.7 & 5.7 & 5.7 & 5.7 & 5.7 & -- & -- \\
\bottomrule
\end{tabular}
\caption{Performance by Region and Index Selection for DT for \hopper.}
\label{tab:mirrored_dt_h}

%% file: tables/rule_mirroring/orcaid_halfCheetah.tex
\begin{tabular}{l|cccccc}
\toprule
& \multicolumn{6}{c}{comb-1 \& random-1-25} \\\hline
Max. Depth & ${\leq}1$ & ${\leq}2$ & ${\leq}3$ & ${\leq}4$ & ${\leq}5$ & ${\leq}6$ \\\hline
Rate (\%) & 0 & 0 & 100 & -- & -- & -- \\
Best (\%) & 0.0 & 0.0 & 2.9 & -- & -- & -- \\
\hline
& \multicolumn{6}{c}{comb-1 \& random-2-25} \\\hline
Rate (\%) & 0 & 67 & 100 & -- & -- & -- \\
Best (\%) & 0.0 & 2.6 & 2.6 & -- & -- & -- \\
\hline
& \multicolumn{6}{c}{comb-2 \& random-2-25} \\\hline
Rate (\%) & 0 & 0 & 33 & 67 & 67 & 67 \\
Best (\%) & 0.0 & 0.0 & 2.1 & 2.3 & 2.3 & 2.3 \\
\hline
& \multicolumn{6}{c}{comb-1 \& random-4-25} \\\hline
Rate (\%) & 0 & 0 & 0 & 0 & 67 & 67 \\
Best (\%) & 0.0 & 0.0 & 0.0 & 0.0 & 2.7 & 2.7 \\
\bottomrule
\end{tabular}
\caption{Performance by Region and Index Selection for \toolname for \halfCheetah.}
\label{tab:mirrored_orcaid_hc}

%% file: tables/rule_mirroring/DT_halfCheetah.tex
\begin{tabular}{l|cccccccccc}
\toprule
& \multicolumn{10}{c}{comb-1 \& random-1-25} \\\hline
Max. Depth & ${\leq}1$ & ${\leq}3$ & ${\leq}5$ & ${\leq}7$ & ${\leq}9$ & ${\leq}11$ & ${\leq}13$ & ${\leq}15$ & ${\leq}17$ & ${\leq}20$ \\\hline
Rate (\%) & 0 & 38 & 38 & 38 & 38 & 38 & 38 & 38 & 38 & 38 \\
Best (\%) & 0.0 & 2.6 & 2.6 & 2.6 & 2.6 & 2.6 & 2.6 & 2.6 & 2.6 & 2.6 \\
\hline
& \multicolumn{10}{c}{comb-1 \& random-2-25} \\\hline
Rate (\%) & 0 & 25 & 25 & 38 & 38 & 38 & 38 & 38 & 38 & 38 \\
Best (\%) & 0.0 & 2.4 & 2.4 & 2.6 & 2.6 & 2.6 & 2.6 & 2.6 & 2.6 & 2.6 \\
\hline
& \multicolumn{10}{c}{comb-2 \& random-2-25} \\\hline
Rate (\%) & 43 & 43 & 57 & 57 & 57 & 57 & 57 & 57 & 57 & 57 \\
Best (\%) & 2.6 & 2.6 & 2.6 & 2.6 & 2.6 & 2.6 & 2.6 & 2.6 & 2.6 & 2.6 \\
\hline
& \multicolumn{10}{c}{comb-1 \& random-4-25} \\\hline
Rate (\%) & -- & 0 & 0 & 14 & 14 & 14 & 29 & 43 & 43 & 43 \\
Best (\%) & -- & 0.0 & 0.0 & 2.7 & 2.7 & 2.7 & 2.7 & 2.7 & 2.7 & 2.7 \\
\bottomrule
\end{tabular}
\caption{Performance by Region and Index Selection for DT for \halfCheetah.}
\label{tab:mirrored_dt_hc}

%% file: appendixTool.tex
\clearpage

\begin{figure}
\centering
\begin{minipage}{.35\textwidth}
  \centering
  \includegraphics[width=0.98\linewidth]{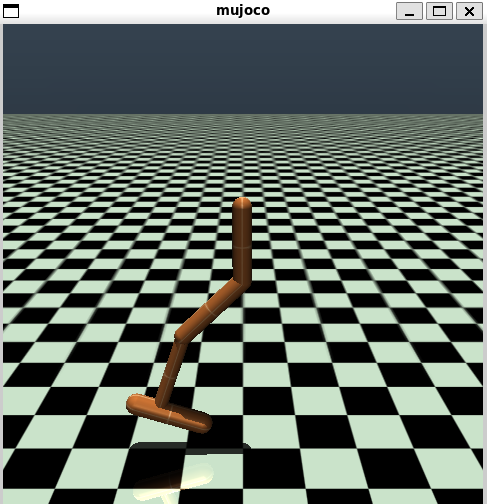}
    \captionof{figure}{Hopper environment whose state has been set manually.}
    \label{fig:hopper_tool}
\end{minipage}
\hfill
\begin{minipage}{.63\textwidth}
  \centering
  \includegraphics[width=.98\linewidth]{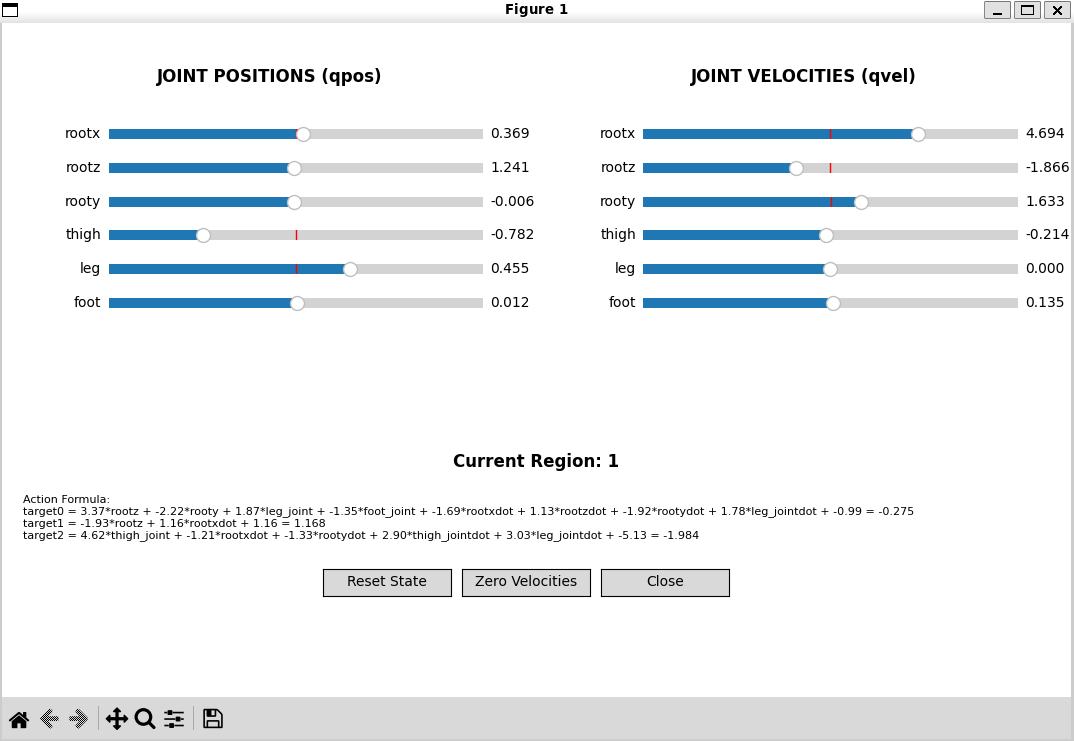}
    \captionof{figure}{User interface to modify the state of Hopper.}
    \label{fig:interface_hopper}
\end{minipage}
\end{figure}

\section{Interactive State Visualization Tool}
\label{sec:appTool}

The code in \texttt{interactive\_representation.py} implements an interactive tool for exploring the regions of a learned \toolname model. The tool enables users to manually manipulate the state of some Gymnasium environments and observe which region the model assigns the state to, together with its corresponding action formula. Fig.~\ref{fig:hopper_tool} displays the Hopper environment in a position set manually by the user via the interface in Fig.~\ref{fig:interface_hopper}. It shows that it is in Region 1 from an \toolname model whose linear models are displayed at the bottom of the interface.

\subsection{Core Functionality}

The tool's primary purpose is to facilitate the understanding of an \toolname learned model. It achieves this by:
\begin{compactenum}[1.]
    \item Providing a graphical interface to set any valid state of the environment (Fig.~\ref{fig:interface_hopper}).
    \item Instantaneously determining the \toolname region for the current state (Fig.~\ref{fig:interface_hopper}).
    \item Displaying the linear model that defines the policy within that region (Fig.~\ref{fig:interface_hopper}).
    \item Visually updating the environment's rendering to reflect the new state (Fig.~\ref{fig:hopper_tool}).
\end{compactenum}

\subsection{Workflow and User Interaction}

The typical workflow for a user of this tool is as follows:

\begin{compactenum}[1.]
    \item \textbf{Initialization:} The user runs the script selecting an \toolname model, which creates the environment and loads the model.
    \item \textbf{Exploration:} The user sees a window with the environment rendering and a separate control panel.
    \item \textbf{Manipulation:} By moving the sliders for \textbf{qpos} and \textbf{qvel}, the user can set the environment to any state. As they move a slider:
    \begin{compactenum}[(a)]
        \item The environment updates.
        \item The corresponding region of the current state for the \toolname updates, showing the conditions and the linear model for that region, as well as the output of that model given the current state.
    \end{compactenum}
    \item \textbf{Hypothesis Testing:} A researcher can use this to answer questions like:
    \begin{compactitem}[•]
        \item ``What is the policy at the edge of this region?"
        \item ``Does the model's behavior in this region match my physical intuition?"
        \item ``How much do I need to change a certain feature to move to a different region?"
    \end{compactitem}
\end{compactenum}

%% file: appendixLLM.tex
Quantitative metrics for model explainability, such as tree depth or number of rules, and model size, often fail to capture the semantic quality of an explanation. A model with a thousand simple but arbitrary rules is not necessarily more explainable than a model with a few complex but physically meaningful partitions. Human evaluation remains the gold standard but is costly and potentially biased.

To address this problem, we have taken the LLM-as-a-Judge framework. We have created a domain-specific evaluation questionnaire with associated rubrics to enable an LLM to act as a control expert and determine whether a model is globally explainable. We demonstrate this by comparing \toolname with decision trees, which are typically assumed to be explainable, and with Cubist, which has the most similar reward ratio w.r.t. \toolname.

\subsection{Methodology}
Our methodology consists of three main steps: (1) Context generation, (2) prompt and rubric design, and (3) execution and analysis.

\subsubsection{Evidence and Context Generation}
For a fair comparison, we standardize the ``case file" provided to the LLM judge for each model. The case file contains:
\begin{compactitem}[•]
    \item \textbf{Problem Definition.} A description of the control task copied directly from the Gymnasium Documentation.
    \item \textbf{Model Representation.} In order not to introduce bias into the model, we introduced the models as Model A and Model B, instead of naming them. Each feature in the model is named according to its documentation, representing a clear physical input of the model.
    \begin{compactitem}[-]
        \item \textbf{Model A (\toolname):} A textual description of the model with the definitions of the conditions to split the state space, the conditions for each region, and the linear models in each region.
        \item \textbf{Model B (Decision Tree):} The full text describing the tree in an indented manner.
        \item \textbf{Model B (Cubist):} The list of rules with their conditions and linear models per target, as a textual description.
    \end{compactitem}
\end{compactitem}

\subsubsection{Prompt and Rubric Design}
We first assigned a persona (a control expert), then provided the model's case file, and finally asked the LLM to evaluate the models based on the following questionnaire. Based on the list suggested by~\cite{contrerasolivas2025}, we included 7 explainability-related questions. Additionally, we designed 8 domain-specific questions, as shown below. We assigned a meaning to each question's score so that each LLM's score would have the same meaning, allowing us to later summarize them all together. Each criterion is scored on a 1-5 Likert scale (5: best, 1: worst). LLMs must also provide a justification for each score, forcing them to ``think'' more and produce more accurate results.

\paragraph{Principles:}
\begin{compactenum}[1.]
    \item \textbf{Robustness}: Is the explanation consistent and reliable across different input conditions, avoiding significant variations or failures when the model is subjected to slight perturbations in the environment?\\
    \textbf{Rubrics}: \\
    5: Explanation remains unchanged under slight perturbations. \\
    3: Slight perturbations cause noticeable changes in the explanation or in the operating modes it captures, while preserving core decision logic.\\
    1: Explanations fail or change drastically when subjected to slight perturbations.
    \item \textbf{Bias}: Does the explanation account for and reveal potential biases in the RL agent's decision-making, ensuring that it does not systematically favor or disadvantage certain actions? \\
    \textbf{Rubrics}: \\
    5: The explanation reveals any bias in the RL agent's decisions and provides insight into the cause for bias. \\
    3: The explanation reveals sources of bias, but fails to capture causes, e.g., resulting from the environment. \\
    1: The explanation completely ignores sources of systematic bias.
    \item \textbf{Transferability}: Can the insights derived from the explanation be applied to similar RL models, different environments, or other tasks, ensuring generalizability beyond a single use case? \\
    \textbf{Rubrics}: \\
    5: The explanation supports reasoning about policies with partially deviating behavior and about environment variations, e.g., local obstacles or additional features. \\
    3: The explanation supports reasoning about either policies with deviating behavior or slight environment variations, but not both. \\
    1: The explanation overfits to the given RL policy and environments. 
    \item \textbf{Human Comprehensibility}: Is the explanation structured in a way that aligns with human cognitive abilities, avoiding excessive complexity while maintaining interpretability?\\
    \textbf{Rubrics}: \\
    5: A user could trace what factors led to an action decision through a low number of intuitive steps. \\
    3: A user could trace what factors led to an action decision through either a large number of steps, requiring memorization of intermediate decisions, or steps requiring complex computation. \\
    1: Decisions are hard to trace, requiring a large number of steps, memorization of intermediate steps, and following unintuitive decision-making.
    \item \textbf{Transparency}: Does the explanation offer a clear and verifiable account of the RL agent's decision-making process? \\
    \textbf{Rubrics}: \\
    5: The explanation captures all decision-relevant details, and its complexity admits verification and validation. \\
    3: Some factors leading to a decision are omitted and verification and validation requires substantial engineering.\\
    1: The explanation obscures essential details, and verification and validation are infeasible. 
\end{compactenum}

\paragraph{Good explanations:}
\begin{compactenum}[1.]
\setcounter{enumi}{5}
    \item \textbf{Selective}: Does the explanation focus on the most relevant and influential factors affecting the agent's decision, avoiding unnecessary details that may overwhelm the user? \\
    \textbf{Rubrics}: \\
    5: The explanation includes mostly relevant details. \\
    3: The explanation contains relevant details, but also details that are not necessary. \\
    1: The explanation comprises a significant portion of unnecessary details, making it infeasible to discern what details are relevant. 
    \item \textbf{Facilitating Actions}: Does the explanation provide actionable insights that enable the user to modify inputs, adjust policies, or intervene effectively based on the RL system's decision-making process? \\
    5: The explanation provides sufficient insight for a user familiar with the environment to modify or monitor the agent's policy, e.g., adapting to state-space changes or behavior shifts. \\
    3: The explanation provides some insight, but modifications require familiarity with the control policy. \\
    1: The explanation does not provide insights that support modifications with reasonable effort.
\end{compactenum}

\paragraph{Domain-specific:}
\begin{compactenum}[1.]
\setcounter{enumi}{7}
    \item \textbf{Physical coherence}: Does the model's partitioning of the state space align with the system's meaningful physical regimes or modes? Or are the regions arbitrary, non-intuitive, and too small from a physics perspective? \\
    \textbf{Rubrics}: \\
    5: The partitions correspond to operating modes expected in the given environment, and a user can connect them to physical regimes/dynamics. \\
    3: Some partitions reflect expected operating modes, while some are arbitrary. Establishing a connection to the dynamics requires minor mathematical transformations. \\
    1: There is no intuitive connection between conditions in the explanation and the environment's dynamics and physical properties.
    \item \textbf{Granularity and partitioning}: How well does the model structure the state space into meaningful regions? Are these regions too large or too small? Are there any arbitrary or poorly defined regions? Is there a risk of overfitting due to excessive partitioning?\\
    \textbf{Rubrics}: \\
    5: Regions have clear, interpretable semantic meaning derived from shared constraints that align with the environment’s physical dynamics and are appropriately sized to balance complexity and interpretability.\\
    3: Some regions have partial semantic meaning, but others are small, fragmented, or unrelated to the environment’s dynamics. Interpretability is limited.\\
    1: Regions are arbitrary, tiny clusters with no semantic or structural coherence, making them completely uninterpretable.
    \item \textbf{Debugging}: After a failure, how easy would it be to diagnose the problem? Does the model's structure help to identify a certain physical regime, or does it give you a rule that failed in a particular small region of the state space?\\
    \textbf{Rubrics}: \\
    5: Failures can be traced to decisions in specific regions and the cause of failures can be deduced with moderate effort. \\
    3: Users need to consider various regions and alternative decisions to trace failures, and failure causes cannot be unambiguously deduced. \\
    1: Users need to invest substantial effort into tracing failures, e.g., studying the complete explanation, and deducing the causes of failures is hardly feasible.
    \item \textbf{Interpretability of Regions}: Do regions correspond to physical behaviors? Is it easy to interpret what each region represents in terms of system dynamics?\\
    \textbf{Rubrics}:\\
    5: Users knowing the environment understand the aspects of system dynamics that prompted the definition of most regions. \\
    3: Users knowing the environment understand the aspects of system dynamics that prompted the definition of several regions, while others do not correspond to system dynamics. \\
    1: Users who know the environment very well cannot establish a connection between system dynamics and regions in the explanation.
    \item \textbf{Global Interpretability}: Is the high-level decision-making intuitive, or is it obscured by complexity? \\
    \textbf{Rubrics}: \\
    5: Regions are appropriately sized to capture complex, meaningful behaviors without overfitting. They balance granularity and interpretability while maintaining clear semantic links to the environment's dynamics. \\
    3: Some regions are small or unrelated, reducing interpretability. Others capture partial semantic meaning but lack cohesion, with a mix of useful and fragmented partitions. \\
    1: Regions are tiny, arbitrary clusters with no clear meaning, making them completely uninterpretable and disconnected from the system's behavior.
    \item \textbf{Minimal Granularity}: Is the number of regions justified by the system's behavior and the environment's complexity? Are there too many regions, leading to unnecessary complexity, excessive noise, and a failure to capture the essential dynamics? \\
    \textbf{Rubrics}: \\
    5: The number of regions is minimal and justified by the environment's complexity and efficient policies. Regions capture essential dynamics without redundancy, balancing simplicity and completeness. \\
    3: The number of regions is larger than expected for the complexity of the environment and efficient policies, leading to unnecessary cognitive load. \\
    1: The number of regions is excessive, creating a high cognitive load. Regions are overly fragmented, with no clear justification for their quantity.
    \item \textbf{Stability Across Regions}: Are transitions between regions smooth and predictable?\\
    \textbf{Rubrics}: \\
    5: Regions adjacent to each other generally predict similar actions, especially on the borders. Sudden changes in the region's borders are justified by environmental dynamics. \\
    3: Transitioning between regions causes jumps unnecessarily, but the magnitude of jumps is generally small. Predicting the jumps and their magnitude is not always possible. \\
    1: The explanation generally shows non-smooth behavior with unpredictable jumps in the predicted actions.
    \item \textbf{Simplicity}: Is the number of parameters justified, or are there too many, making it unnecessarily harder to understand? \\
    \textbf{Rubrics}: \\
    5: The number of parameters is minimal and well justified, directly aligning with the environment's state-space structure and action prediction needs. Parameters are interpretable and contribute to clear, concise explanations. \\
    3: The number of parameters is larger than ideal, with some redundancy. Parameters could be reduced while still adequately explaining actions, but interpretability is moderate.\\
    1: The number of parameters is excessive, making the explanation infeasible to understand.
\end{compactenum}

\paragraph{Final:}
\begin{compactenum}[1.]
\setcounter{enumi}{15}
    \item \textbf{Verdict}: Declare the winning model. Base your conclusion on the previous answers and on the principles of control theory and robotics. An explainable controller should be understandable, trustworthy, and verifiable. Coherence, stability, and adaptability of the global policy are important measures of explainability in this context.
\end{compactenum}

\subsubsection{Execution and Analysis}
We used six SOTA LLMs available for free for a limited number of requests at \url{https://openrouter.ai/}:
\begin{compactitem}[•]
    \item \textit{xiaomi/mimo-v2-flash:free},
    \item \textit{meta-llama/llama-3.3-70b-instruct:free},
    \item \textit{deepseek/deepseek-r1-0528:free},
    \item \textit{z-ai/glm-4.5-air:free},
    \item \textit{qwen/qwen3-coder:free},
    \item \textit{openai/gpt-oss-120b:free}.
\end{compactitem}
Every surrogate for every LLM was evaluated at a low temperature of 0.2 to get more coherent and reproducible answers. The average scores and standard errors are shown in Table~\ref{tab:app_all_LLMs_vs_DT} and Table~\ref{tab:app_all_LLMs_vs_Cubist}.

\subsubsection{Model A: \toolname Case File}
The following lines show the case file provided to the LLMs to represent an \toolname model for Pendulum.
\begin{verbatim}
Model Representation:
C0: -0.24y - 0.97ang_vel <= -0.60
C1: -0.19x - 0.41y - 0.89ang_vel <= -0.87
C2: 0.30x - 0.39y - 0.87ang_vel <= -0.33
Region 0:
C0 & C1
Model: z = -4.94x + -4.66y + 6.07
Region 1:
C0 & not C1
Model: z=-1.11x-1.04y-1.41ang_vel+3.47
Region 2:
C2 & not C0
Model: z = 2.54x - 1.05y - 1.55
Region 3:
not C0 & not C2
Model: z = 2.74x + 3.20*y - 2.80
\end{verbatim}

\subsubsection{Model B: Decision Tree Case File}
The following text shows the case file provided to the LLMs to represent a DT model for Pendulum, simplified for readability purposes.
\begin{verbatim}
Model Representation:
|--- ang_vel <= 0.48
| |--- x <= 0.40
| | |--- y <= 0.66
| | | |--- x <= 0.33
| | | | |--- ang_vel <= 0.06
| | | | | |--- x <= 0.16
| | | | | | |--- x <= 0.13
| | | | | | | |--- value: [-1.96]

[... removed 204 lines ...]

| | | |--- x > 0.78
| | | | |--- ang_vel <= 0.48
| | | | | |--- value: [0.49]
| | | | |--- ang_vel > 0.48
| | | | | |--- ang_vel <= 0.51
| | | | | | |--- value: [-1.41]
| | | | | |--- ang_vel > 0.51
\end{verbatim}

\subsubsection{Full LLM Evaluation Prompt}
The following prompt was given to the LLMs for each environment. The prompt also included instructions to format their answers according to a given JSON structure, which is omitted for better readability.

\noindent\texttt{You are an expert in robotics and control systems, tasked with auditing the explainability and trustworthiness of two different surrogate models that were learned from a reinforcement learning model using imitation learning.}
\begin{center}
    \vspace{-4pt}
    \texttt{[Gymnasium Description]}
    \vspace{-4pt}
\end{center}
\noindent\texttt{Model A is defined as follows:}
\begin{center}
    \vspace{-4pt}
    \texttt{[Model A Case File]}
    \vspace{-4pt}
\end{center}
\noindent\texttt{Model B is defined as follows:}
\begin{center}
    \vspace{-4pt}
    \texttt{[Model B Case File]}
    \vspace{-4pt}
\end{center}
\noindent\texttt{Your goal is to determine which model is more understandable, verifiable, and ultimately more trustworthy for deployment. Please evaluate both using the following rubric.
Each criterion is scored on a 1-5 Likert scale, with 5 indicating a strong positive answer ('Excellent') and 1 indicating a strong negative answer ('Very Poor'). You are also required to provide a brief justification for each score.}
\begin{center}
    \vspace{-4pt}
    \texttt{[Questionnaire]}
    \vspace{-3pt}
\end{center}

\subsection{Results}
Table~\ref{tab:app_all_LLMs_vs_DT} shows the average score across the 6 LLMs per question and per environment when comparing \toolname with DT. One can see that \toolname scores better on most questions across all environments, except Hopper and Half Cheetah. For Hopper, the scores are mixed, and for Half Cheetah, DT shows slightly better results. However, \toolname and DT are not comparable for Half Cheetah since \toolname performs much better (reward ratio $\sim\!75\%$ vs $\sim\! 47\%$). Analogously, Table~\ref{tab:app_all_LLMs_vs_Cubist} shows the score average for the comparison between \toolname and Cubist, where \toolname outperforms Cubist across all environments, with the exception of Hopper, where results are mixed.

These results clearly indicate that \toolname might be more interpretable than DTs and Cubist, especially when the latter have similar performance to \toolname. This is even more noticeable in the examples when \toolname models are built from oblique trees with a small depth.

\begin{table*}[]
\centering
\tiny
\setlength{\tabcolsep}{1.75pt}
\renewcommand{\arraystretch}{1.2}
\begin{tabular}{ll|l:l:l:l:l:l:l:l:l:l:l:l:l:l:l}
Env. &  \multicolumn{2}{l}{Model} & \multicolumn{14}{c}{Question} \\
\hline
   & & 1           & 2           & 3           & 4           & 5           & 6           & 7           & 8           & 9           & 10          & 11          & 12          & 13          & 14          & 15          \\
\hline
\multirow{ 2}{*}{\mountainCar} & \toolnameAbbr & 3.7$\pm$0.8 & 3.5$\pm$0.8 & 3.8$\pm$0.8 & 4.7$\pm$0.5 & 4.8$\pm$0.4 & 4.8$\pm$0.4 & 4.7$\pm$0.5 & 3.8$\pm$1.3 & 4.3$\pm$1.0   & 4.7$\pm$0.5 & 4.3$\pm$1.0   & 4.5$\pm$0.8 & 4.7$\pm$0.8 & 3.8$\pm$1.0   & 4.7$\pm$0.8 \\
 & DT & 2.0$\pm$0.9   & 2.8$\pm$0.4 & 1.7$\pm$0.8 & 2.0$\pm$0.9   & 2.5$\pm$0.8 & 2.5$\pm$0.5 & 2.7$\pm$0.8 & 2.5$\pm$0.8 & 2.2$\pm$0.8 & 2.2$\pm$0.8 & 2.3$\pm$0.8 & 2.2$\pm$0.8 & 1.3$\pm$0.8 & 1.8$\pm$0.8 & 1.5$\pm$0.8 \\
\hline
\multirow{ 2}{*}{\pendulum} & \toolnameAbbr & 3.5$\pm$0.5 & 3.3$\pm$0.8 & 3.7$\pm$0.5 & 4.0$\pm$0.0     & 4.3$\pm$0.5 & 4.0$\pm$0.0     & 4.2$\pm$0.4 & 3.7$\pm$0.8 & 4.0$\pm$0.6   & 4.3$\pm$0.5 & 3.7$\pm$0.8 & 4.3$\pm$0.5 & 4.3$\pm$0.5 & 3.7$\pm$0.8 & 4.5$\pm$0.5 \\
 & DT & 2.2$\pm$0.4 & 2.0$\pm$0.0     & 1.8$\pm$0.4 & 1.5$\pm$0.5 & 2.2$\pm$0.8 & 1.5$\pm$0.5 & 2.0$\pm$0.6   & 2.0$\pm$0.6   & 1.3$\pm$0.5 & 1.7$\pm$0.5 & 1.5$\pm$0.8 & 1.3$\pm$0.5 & 1.2$\pm$0.4 & 1.7$\pm$0.8 & 1.2$\pm$0.4 \\
\hline
\multirow{ 2}{*}{\lunarLander} & \toolnameAbbr & 3.2$\pm$0.4 & 3.3$\pm$0.8 & 3.5$\pm$0.8 & 4.0$\pm$0.9   & 4.2$\pm$1.0   & 3.5$\pm$0.8 & 4.0$\pm$0.9   & 3.0$\pm$1.8   & 3.8$\pm$1.0   & 4.0$\pm$0.9   & 3.0$\pm$1.8   & 4.0$\pm$0.9   & 4.0$\pm$0.9   & 3.2$\pm$1.3 & 4.0$\pm$0.9 \\
   & DT & 1.5$\pm$0.8 & 2.5$\pm$0.8 & 1.7$\pm$0.8 & 1.3$\pm$0.5 & 2.2$\pm$1.0   & 1.8$\pm$1.0   & 1.7$\pm$0.8 & 2.5$\pm$0.8 & 1.2$\pm$0.4 & 1.5$\pm$0.8 & 1.8$\pm$1.0   & 1.2$\pm$0.4 & 1.2$\pm$0.4 & 1.3$\pm$0.5 & 1.3$\pm$0.5 \\
\hline
\multirow{ 2}{*}{\invPendulum} & \toolnameAbbr & 4.0$\pm$0.9   & 3.5$\pm$0.8 & 4.3$\pm$0.8 & 5.0$\pm$0.0     & 5.0$\pm$0.0     & 4.8$\pm$0.4 & 5.0$\pm$0.0     & 4.0$\pm$0.9   & 4.7$\pm$0.8 & 4.7$\pm$0.8 & 4.3$\pm$1.0   & 4.7$\pm$0.8 & 4.7$\pm$0.8 & 4.2$\pm$1.0   & 5.0$\pm$0.0 \\
  & DT & 2.0$\pm$0.9   & 2.8$\pm$0.4 & 1.7$\pm$0.8 & 1.7$\pm$0.8 & 2.7$\pm$0.5 & 2.3$\pm$0.8 & 2.3$\pm$0.8 & 1.8$\pm$1.0   & 1.5$\pm$0.8 & 1.5$\pm$0.8 & 1.7$\pm$1.0   & 1.5$\pm$0.8 & 1.3$\pm$0.8 & 1.7$\pm$0.8 & 1.3$\pm$0.8 \\
\hline
\multirow{ 2}{*}{\invDoublePendulum} & \toolnameAbbr & 3.7$\pm$0.8 & 3.3$\pm$1.0   & 3.7$\pm$0.8 & 4.0$\pm$0.9   & 4.2$\pm$0.8 & 3.8$\pm$1.0   & 3.8$\pm$1.0   & 3.8$\pm$1.0   & 4.0$\pm$0.9   & 4.2$\pm$0.8 & 4.0$\pm$0.9   & 4.0$\pm$0.9   & 4.5$\pm$0.8 & 3.7$\pm$1.2 & 4.2$\pm$0.8 \\
  & DT & 1.5$\pm$0.5 & 2.5$\pm$0.5 & 1.8$\pm$0.8 & 1.3$\pm$0.8 & 2.0$\pm$0.9   & 1.7$\pm$0.8 & 1.5$\pm$0.8 & 2.3$\pm$0.5 & 1.3$\pm$0.8 & 1.5$\pm$0.8 & 1.7$\pm$1.0   & 1.3$\pm$0.8 & 1.3$\pm$0.8 & 1.3$\pm$0.8 & 1.5$\pm$0.8 \\
\hline
\multirow{ 2}{*}{\reacher} & \toolnameAbbr  & 3.0$\pm$0.6   & 2.8$\pm$0.4 & 3.2$\pm$1.2 & 2.8$\pm$0.8 & 3.5$\pm$0.8 & 2.7$\pm$0.8 & 3.2$\pm$1.2 & 3.3$\pm$1.0   & 2.8$\pm$0.8 & 3.2$\pm$1.2 & 3.2$\pm$1.2 & 3.3$\pm$1.0   & 3.7$\pm$1.4 & 3.0$\pm$0.6   & 3.2$\pm$1.2 \\
    & DT & 2.5$\pm$1.2 & 3.2$\pm$0.8 & 2.3$\pm$1.4 & 2.5$\pm$1.4 & 3.5$\pm$0.8 & 2.8$\pm$1.2 & 3.3$\pm$1.0   & 3.5$\pm$1.0   & 2.3$\pm$1.5 & 3.2$\pm$1.5 & 3.0$\pm$1.4   & 2.5$\pm$1.6 & 2.0$\pm$1.7   & 2.3$\pm$1.5 & 2.3$\pm$1.5 \\
\hline
\multirow{ 2}{*}{\swimmer}  & \toolnameAbbr & 3.8$\pm$0.8 & 3.7$\pm$0.8 & 4.0$\pm$0.6   & 4.5$\pm$0.5 & 4.8$\pm$0.4 & 4.5$\pm$0.8 & 4.7$\pm$0.5 & 4.0$\pm$0.6   & 4.0$\pm$0.9   & 4.7$\pm$0.5 & 3.8$\pm$0.8 & 4.7$\pm$0.5 & 4.5$\pm$0.8 & 3.8$\pm$0.8 & 4.7$\pm$0.5\\
   & DT & 1.8$\pm$0.8 & 2.3$\pm$0.5 & 1.8$\pm$0.4 & 1.7$\pm$0.5 & 2.2$\pm$0.8 & 2.5$\pm$1.4 & 2.0$\pm$0.6   & 2.3$\pm$0.5 & 1.3$\pm$0.5 & 1.7$\pm$0.5 & 1.8$\pm$0.8 & 1.5$\pm$0.5 & 1.2$\pm$0.4 & 1.5$\pm$0.5 & 1.2$\pm$0.4 \\
\hline
\multirow{ 2}{*}{\hopper}  & \toolnameAbbr & 2.7$\pm$0.8 & 2.7$\pm$0.5 & 3.0$\pm$0.6   & 2.5$\pm$1.4 & 3.3$\pm$0.5 & 2.5$\pm$1.4 & 3.2$\pm$0.8 & 2.3$\pm$1.0   & 2.2$\pm$1.0   & 2.8$\pm$1.2 & 1.8$\pm$1.0   & 2.3$\pm$1.2 & 2.8$\pm$1.3 & 2.5$\pm$0.8 & 2.5$\pm$1.4 \\
   & DT & 2.7$\pm$1.0   & 2.5$\pm$1.0   & 2.5$\pm$1.0   & 3.2$\pm$1.2 & 3.5$\pm$1.2 & 3.5$\pm$1.4 & 3.3$\pm$1.6 & 3.3$\pm$1.5 & 2.3$\pm$1.2 & 3.0$\pm$1.4   & 3.2$\pm$1.3 & 2.3$\pm$1.2 & 1.8$\pm$1.3 & 2.3$\pm$1.2 & 2.5$\pm$1.0 \\
\hline
\multirow{ 2}{*}{\halfCheetah} & \toolnameAbbr & 2.7$\pm$0.8 & 2.7$\pm$1.4 & 2.8$\pm$1.3 & 2.5$\pm$1.0   & 2.8$\pm$0.8 & 2.5$\pm$1.0   & 2.7$\pm$1.2 & 2.8$\pm$1.3 & 2.7$\pm$1.4 & 2.8$\pm$1.3 & 2.7$\pm$1.4 & 2.5$\pm$1.4 & 2.7$\pm$1.4 & 2.5$\pm$1.0   & 2.2$\pm$1.0 \\
   & DT & 2.5$\pm$1.0   & 3.3$\pm$1.2 & 2.7$\pm$0.8 & 3.0$\pm$0.9   & 3.2$\pm$1.2 & 3.5$\pm$0.8 & 3.0$\pm$1.1   & 3.3$\pm$1.2 & 2.7$\pm$1.4 & 3.3$\pm$1.5 & 3.2$\pm$1.5 & 2.8$\pm$1.5 & 2.5$\pm$1.5 & 2.7$\pm$1.4 & 2.7$\pm$1.4
\end{tabular}
\caption{Average scores across the 6 LLMs for the 15 questions when asked to compare \toolname and Decision Trees, per environment.}
\label{tab:app_all_LLMs_vs_DT}
\end{table*}

\begin{table*}[]
\centering
\tiny
\setlength{\tabcolsep}{1.75pt}
\renewcommand{\arraystretch}{1.2}
\begin{tabular}{ll|l:l:l:l:l:l:l:l:l:l:l:l:l:l:l}
Env. &  \multicolumn{2}{l}{Model} & \multicolumn{14}{c}{Question} \\
\hline
   & & 1           & 2           & 3           & 4           & 5           & 6           & 7           & 8           & 9           & 10          & 11          & 12          & 13          & 14          & 15          \\
\hline
\multirow{ 2}{*}{\mountainCar} & \toolnameAbbr & 3.8$\pm$0.8 & 3.2$\pm$0.4 & 3.8$\pm$0.8 & 4.7$\pm$0.5 & 5.0$\pm$0.0     & 4.8$\pm$0.4 & 4.7$\pm$0.5 & 4.5$\pm$0.5 & 4.3$\pm$0.8 & 4.5$\pm$0.8 & 4.3$\pm$0.8 & 4.8$\pm$0.4 & 4.7$\pm$0.5 & 3.8$\pm$0.8 & 4.8$\pm$0.4 \\
  & DT & 2.5$\pm$0.8 & 3.3$\pm$1.0   & 2.2$\pm$0.8 & 2.3$\pm$0.5 & 3.0$\pm$0.0     & 2.8$\pm$0.4 & 2.7$\pm$0.5 & 2.7$\pm$0.5 & 2.5$\pm$0.5 & 2.8$\pm$0.8 & 2.7$\pm$0.8 & 2.3$\pm$0.5 & 1.5$\pm$0.8 & 1.7$\pm$0.5 & 1.8$\pm$0.8\\

\hline
\multirow{ 2}{*}{\pendulum} & \toolnameAbbr & 4.2$\pm$1.0   & 3.7$\pm$0.8 & 4.0$\pm$0.9   & 5.0$\pm$0.0     & 5.0$\pm$0.0     & 5.0$\pm$0.0     & 5.0$\pm$0.0     & 4.3$\pm$1.0   & 4.7$\pm$0.8 & 5.0$\pm$0.0     & 4.3$\pm$1.0   & 5.0$\pm$0.0     & 5.0$\pm$0.0     & 4.0$\pm$1.1   & 5.0$\pm$0.0 \\
  & DT & 2.2$\pm$1.0   & 3.0$\pm$0.0     & 2.0$\pm$0.9   & 1.2$\pm$0.4 & 2.3$\pm$0.8 & 2.0$\pm$0.9   & 2.5$\pm$0.8 & 1.8$\pm$1.0   & 1.2$\pm$0.4 & 1.7$\pm$1.0   & 1.5$\pm$0.8 & 1.3$\pm$0.8 & 1.0$\pm$0.0     & 1.3$\pm$0.5 & 1.0$\pm$0.0\\

\hline
\multirow{ 2}{*}{\lunarLander} & \toolnameAbbr & 3.5$\pm$0.5 & 3.0$\pm$0.6   & 3.5$\pm$0.5 & 3.5$\pm$0.5 & 3.8$\pm$0.8 & 3.8$\pm$0.4 & 3.7$\pm$0.5 & 3.8$\pm$0.8 & 3.7$\pm$0.5 & 3.8$\pm$0.8 & 3.7$\pm$1.0   & 3.8$\pm$0.8 & 3.8$\pm$0.8 & 3.7$\pm$0.5 & 4.0$\pm$0.6 \\
  & DT & 2.0$\pm$0.6   & 2.0$\pm$0.6   & 1.8$\pm$0.4 & 1.5$\pm$0.5 & 2.0$\pm$0.6   & 2.0$\pm$0.6   & 2.0$\pm$0.9   & 2.0$\pm$0.6   & 1.5$\pm$0.5 & 1.8$\pm$0.8 & 1.8$\pm$0.8 & 1.5$\pm$0.5 & 1.2$\pm$0.4 & 1.8$\pm$0.4 & 1.2$\pm$0.4 \\

\hline
\multirow{ 2}{*}{\invPendulum} & \toolnameAbbr & 4.0$\pm$0.9   & 3.0$\pm$1.1   & 3.8$\pm$0.8 & 4.7$\pm$0.8 & 5.0$\pm$0.0     & 4.7$\pm$0.5 & 4.8$\pm$0.4 & 3.8$\pm$1.6 & 3.7$\pm$1.6 & 4.5$\pm$0.8 & 4.3$\pm$1.0   & 4.5$\pm$0.8 & 4.3$\pm$1.0   & 3.5$\pm$1.5 & 4.7$\pm$0.8 \\
  & DT & 2.8$\pm$1.3 & 3.0$\pm$0.0     & 2.5$\pm$0.8 & 2.8$\pm$1.0   & 3.3$\pm$0.5 & 3.5$\pm$0.8 & 3.3$\pm$0.5 & 3.2$\pm$1.3 & 2.8$\pm$1.3 & 3.2$\pm$1.3 & 3.2$\pm$1.3 & 2.8$\pm$1.3 & 2.7$\pm$1.0   & 2.5$\pm$0.8 & 2.3$\pm$1.2 \\

\hline
\multirow{ 2}{*}{\invDoublePendulum} & \toolnameAbbr & 3.7$\pm$0.8 & 3.7$\pm$0.8 & 3.8$\pm$0.8 & 4.3$\pm$0.8 & 5.0$\pm$0.0     & 4.7$\pm$0.5 & 4.7$\pm$0.5 & 4.7$\pm$0.8 & 4.8$\pm$0.4 & 4.8$\pm$0.4 & 4.7$\pm$0.8 & 5.0$\pm$0.0     & 4.8$\pm$0.4 & 4.0$\pm$0.9   & 5.0$\pm$0.0 \\
  & DT & 1.7$\pm$0.8 & 2.3$\pm$0.8 & 1.8$\pm$0.8 & 1.2$\pm$0.4 & 1.5$\pm$0.5 & 1.7$\pm$0.8 & 1.5$\pm$0.5 & 1.5$\pm$0.5 & 1.0$\pm$0.0     & 1.0$\pm$0.0     & 1.2$\pm$0.4 & 1.0$\pm$0.0     & 1.0$\pm$0.0     & 1.2$\pm$0.4 & 1.0$\pm$0.0 \\

\hline
\multirow{ 2}{*}{\reacher} & \toolnameAbbr & 3.3$\pm$0.5 & 2.5$\pm$1.2 & 3.3$\pm$0.5 & 3.5$\pm$1.0   & 3.8$\pm$0.8 & 3.8$\pm$0.8 & 3.7$\pm$0.8 & 3.7$\pm$1.0   & 4.0$\pm$0.6   & 4.2$\pm$0.8 & 3.7$\pm$1.0   & 3.8$\pm$0.8 & 4.2$\pm$0.8 & 3.8$\pm$0.8 & 3.8$\pm$0.8 \\
  & DT & 1.7$\pm$0.8 & 2.0$\pm$1.1   & 1.8$\pm$0.8 & 1.5$\pm$0.8 & 2.2$\pm$0.8 & 2.5$\pm$1.2 & 2.0$\pm$0.9   & 2.3$\pm$0.8 & 1.7$\pm$0.8 & 1.7$\pm$0.8 & 2.3$\pm$0.8 & 1.7$\pm$0.8 & 1.3$\pm$0.8 & 1.8$\pm$0.8 & 1.3$\pm$0.8 \\

\hline
\multirow{ 2}{*}{\swimmer} & \toolnameAbbr & 4.0$\pm$0.6   & 3.0$\pm$0.6   & 3.8$\pm$0.4 & 4.8$\pm$0.4 & 5.0$\pm$0.0     & 4.8$\pm$0.4 & 4.8$\pm$0.4 & 4.2$\pm$0.4 & 4.7$\pm$0.8 & 4.8$\pm$0.4 & 4.3$\pm$0.8 & 4.8$\pm$0.4 & 4.7$\pm$0.8 & 4.3$\pm$0.8 & 5.0$\pm$0.0 \\
  & DT & 2.0$\pm$0.6   & 2.5$\pm$0.8 & 1.5$\pm$0.5 & 1.3$\pm$0.5 & 1.5$\pm$0.5 & 1.3$\pm$0.5 & 1.7$\pm$0.5 & 1.7$\pm$0.5 & 1.2$\pm$0.4 & 1.2$\pm$0.4 & 1.2$\pm$0.4 & 1.2$\pm$0.4 & 1.2$\pm$0.4 & 1.2$\pm$0.4 & 1.0$\pm$0.0 \\

\hline
\multirow{ 2}{*}{\hopper} & \toolnameAbbr &  3.0$\pm$0.0     & 2.8$\pm$0.4 & 2.9$\pm$0.5 & 3.2$\pm$0.5   & 2.9$\pm$1.0   & 3.5$\pm$1.0   & 3.5$\pm$0.8 & 2.5$\pm$1.0   & 3.0$\pm$0.8 & 2.7$\pm$1.0   & 3.5$\pm$1.0   & 2.9$\pm$1.0   & 2.5$\pm$1.0   & 2.7$\pm$1.0   & 2.5$\pm$1.0 \\
  & DT & 2.5$\pm$1.4 & 3.2$\pm$0.4 & 2.5$\pm$1.2 & 2.8$\pm$1.2 & 3.0$\pm$1.3   & 3.5$\pm$0.8 & 3.7$\pm$1.5 & 3.5$\pm$0.8 & 2.7$\pm$1.6 & 3.5$\pm$1.5 & 3.5$\pm$0.8 & 2.5$\pm$1.5 & 2.5$\pm$1.5 & 2.7$\pm$1.5 & 2.5$\pm$1.5 \\

\hline
\multirow{ 2}{*}{\halfCheetah} & \toolnameAbbr & 3.7$\pm$1.2 & 2.8$\pm$1.2 & 4.0$\pm$1.1   & 2.7$\pm$1.2 & 3.3$\pm$0.8 & 2.8$\pm$1.2 & 3.2$\pm$1.0   & 4.0$\pm$1.1   & 3.5$\pm$1.2 & 3.5$\pm$1.0   & 4.0$\pm$1.1   & 3.8$\pm$1.2 & 3.5$\pm$1.2 & 3.8$\pm$1.2 & 3.5$\pm$1.2\\
  & DT & 2.5$\pm$0.5 & 3.3$\pm$1.0   & 2.7$\pm$0.5 & 3.3$\pm$1.4 & 3.5$\pm$1.4 & 3.5$\pm$1.0   & 3.3$\pm$1.4 & 2.5$\pm$0.5 & 2.2$\pm$1.0   & 3.0$\pm$1.4   & 2.2$\pm$1.0   & 2.2$\pm$1.0   & 2.2$\pm$1.0   & 2.7$\pm$0.5 & 2.2$\pm$1.0
\end{tabular}
\caption{Average scores across the 6 LLMs for the 15 questions when asked to compare \toolname and Cubist, per environment.}
\label{tab:app_all_LLMs_vs_Cubist}
\end{table*}

\subsubsection{Qualitative Analysis}
LLM's justifications explained the scores. As an example, for Pendulum with \toolname depth 2 (Model A) and DT depth 7 (Model B), a judge explained its verdict towards \toolname as follows:
\begin{quote}
    ``Model A [{\toolname}]'s strategy is clear, sound, and physically motivated. It decomposes the complex pendulum problem into a small number of physically meaningful regimes, each governed by a simple, analyzable law. This structure provides a direct path to understanding, verification, and debugging. Its global policy is coherent and robust".
\end{quote}

\noindent On the other hand, the DT model was criticized:
\begin{quote}
    ``Model B [DT]'s strategy is non-existent. It is a high-resolution, data-driven lookup table that has memorized a solution. Its individual rules are simple, but the resulting global policy is a chaotic patchwork with no discernible physical basis. This lack of an overarching strategy makes it a black box, rendering it unverifiable, brittle, and ultimately untrustworthy for a real-world deployment". 
\end{quote}

Judges usually highlighted the small number of regions and parameters of \toolname. In contrast, they criticized the excessive number of conditions/rules in the DT and Cubist models. They suggested that \toolname regions could more likely be associated with different modes of operations, whereas DT and Cubist partitions seemed to be arbitrary and too granular.

Therefore, the results support our hypothesis that \toolname models can be interpretable. The LLM-as-a-Judge successfully identified and valued the semantic quality of \toolname's global strategy.

\input{tables/RL_hyperparams}

\input{tables/rule_learning_hyperparams}

\subsection{Limitations}
Our methodology depends on prompt design and may be sensitive to the specific LLM version. We mitigated these dependencies by using a subset of the given list in~\cite{contrerasolivas2025}, homogenizing the questions across all LLMs by providing explanations for the scores, and reducing randomness by setting a low temperature and using multiple LLMs. A more extensive analysis of different LLMs and learned models, together with a human evaluation, is an important direction for future work.

%% file: tables/RL_hyperparams.tex
\begin{table*}[ht!]
\centering
\footnotesize
\setlength{\tabcolsep}{1.5pt}
\begin{tabular}{c|ccccccccccc}
          & \# steps & Policy & Architecture & Act. fun. & Batch & Buffer & $\gamma$ & Grad. steps & Learn. rate & Train. freq. & Learn. starts\\
\hline
\mountainCar & 150k & MLP & [512,256,128,64] & ReLU & 256 & 20k & 0.95 & -1 & $10^{-4}$ & 1 & 100\\
\pendulum & 300k & MLP & [400, 300] & ReLU & 256 & 20k & 0.98 & 1 &  $10^{-4}$ & 1 & 1k\\
\lunarLander & 2000k & MLP & [400, 300] & ReLU & 256 & 200k & 0.98 & 1 &  $10^{-3}$ & 1 & 10k \\
\invPendulum & 750k & MLP & [400, 300] & ReLU & 256 & 20k & 0.98 & 1 &  $10^{-4}$ & 1 & 1k\\
\invDoublePendulum & 3000k & MLP & [400, 300] & ReLU & 256 & 1000k & 0.99 & 1 &  $10^{-5}$ & 1 & 10k\\
\reacher & 1000k & MLP & [400, 300] & ReLU & 256 & 200k & 0.99 & 1 &  $10^{-3}$ & 1 & 10k\\
\swimmer & 2000k & MLP & [400, 300] & ReLU & 256 & 1000k & 0.9999 & 1 &  $10^{-3}$ & 1 & 10k\\
\hopper & 3000k & MLP & [400, 300] & ReLU & 256 & 1000k & 0.99 & 1 &  $3\cdot10^{-4}$ & 2 & 25k\\
\halfCheetah & 250k & MLP & [400, 300] & ReLU & 256 & 1000k & 0.99 & 1 &  $10^{-3}$ & 1 & 10k
\end{tabular}
\caption{\textbf{Hyperparameters} per environment for \textbf{reinforcement learning} per environment.}
    \label{tab:rl_hyperparams}
\end{table*}

%% file: tables/rule_learning_hyperparams.tex
\begin{table}[h!]
\centering
\setlength{\tabcolsep}{1.3pt}
\footnotesize
\begin{tabular}{c|cccc|c} 
\textbf{Environment} & \multicolumn{4}{c|}{\textbf{all surrogates}} & \textbf{\toolname} \\
 & \#steps & \#clusters & \#samples & $\sigma$ & \texttt{max\_mse} \\
\hline
\mountainCar & 1k & 4 & 1k & 0.15 & 0.05\\
\pendulum & 20k & 15 & 5k & 0.15 & 0.2\\
\lunarLander & 50k & 30 & 5k & 0.15 & 0.05\\
\invPendulum & 20k & 25 & 5k & 0.15 & 0.45\\
\invDoublePendulum & 100k & 100 & 20k & 0.05 & 0.001\\
\reacher & 100k & 100 & 20k & 0.05 & 0.002\\
\swimmer & 100k & 100 & 20k & 0.15 & 0.05\\
\hopper & 100k & 400 & 20k & 0.02 & 0.01\\
\halfCheetah & 100k & 6k & 40k & 0.01 & 0.01
\end{tabular}
\caption{\textbf{Hyperparameters} per environment for \textbf{\toolname}.}
\label{tab:orcaid_hyperparams}
\end{table}

%% file: appendix.tex
\section{Hyperparameters}
\label{sec:Apphyperparams}

Table~\ref{tab:rl_hyperparams} shows the hyperparameters used for training an agent using RL for each environment. They are the ones that can be adjusted with Stable Baselines3 (SB3). Hyperparameters not listed in the table use their default values.

These hyperparameters were selected using RL Baselines3 Zoo and Hugging Face (\url{https://huggingface.co/}) models as references. However, they were slightly fine-tuned to reduce the number of training steps and speed up convergence. One can see that most of the values are similar across environments. Convergence was reached before the end of the training steps, ensuring the RL agent had sufficient performance.

Table~\ref{tab:orcaid_hyperparams} presents the hyperparameters for \toolname used for the experiments in the paper. Each of the hyperparameters is described as follows:
\begin{compactitem}[•]
    \item \#steps: Number of steps collected from the executions of the agent interacting with the environment.
    \item \texttt{max\_mse}: A leaf stops growing if its MSE does not exceed \texttt{max\_mse}. It appears in Alg.~\appref{algo:oblique_tree_full}. It is not scaled in the action range, so MSE values across domains cannot be compared unless one adjusts them by multiplying by $1/(h-l)$, where $h$ and $l$ are the highest and lowest action values. This adjustment is made in the body of the paper, not here, because the non-scaled MSE is the one provided to our tool that implements \toolname.
    \item \#clusters: Number of clusters used for clustering the original data collected from the executions. The algorithm that is used to cluster the points is K-means due to its high efficiency.
    \item \#samples: Number of data points sampled from the clusters, to which noise was added.
    \item $\sigma$: Standard deviation of the normal distribution function used to add noise to the sampled clusters.
\end{compactitem}

Recall that $\epsilon$, the relative error threshold to merge adjacent regions (stop condition no. 3, ``rel. improvement'' in the paper), is set to $0.05$ for all the experiments. Also, the MSE threshold to merge regions is set to $\tau=2\cdot$\texttt{max\_mse} for all the experiments. There are two main reasons to set a higher MSE threshold like this. First, this creates a more granular partition of the state space by allowing finer splits. Second, it provides flexibility to later combine smaller regions into larger ones. If both thresholds were too similar, we would have trouble finding suitable regions to merge.

\input{tables/all_performance_mse_results}

\section{Performance, MSE, model size, and time}
\label{sec:app_main_exp}

Table~\ref{tab:results_performance} represents the underlying data for Fig.~\appref{fig:plot_ratio_model_size} and Fig.~\appref{fig:plot_mses} of the paper. They include the mean and standard error for each metric. To compute the confidence intervals with a probability of $\alpha=0.05$ shown on the plots of the paper, we used the following formula since the number of samples was relatively small ($N=9$):
\[
\bar{x}\pm t_{0.05/2,8}\cdot SE,
\]
where $SE$ is the standard error, $SE=\sigma/\sqrt{N}$, $t_{0.05/2,8}$ is the value of the Student's t-distribution with $9-1=8$ degrees of freedom whose cumulative distribution function equals $1-\alpha/2=0.975$, i.e., $t_{0.05/2,8}\approx2.3$.

Table~\ref{tab:results_performance} also includes the maximum depth (MD for \toolname and Decision Trees) or maximum number of rules that were used (MNR for Cubist and RuleFit). The depth that is needed to reach an equal or better performance for \toolname in comparison to Decision Trees is much smaller. Since MD with MNR are not comparable, we compared model sizes in the paper, showing that \toolname model sizes are smaller than those of the other surrogates when the reward ratios were similar.

Tables~\ref{tab:orcaid_mc}-~\ref{tab:rulefit_hc} display RL reward/surrogate reward ratio in \%, MSE for the training and test sets, model size calculated as explained in the paper (Def.~\appref{def:number_params} and Section~\appref{sec:exp}), and the total training time, which includes iterative trainings and evaluations as for the DAgger methodology.

For \toolname, neither the training time nor the model size increases exponentially. This is thanks to the stopping conditions and the simplification of regions, which reduces the number of linear regressions and conditions, thus reducing the number of parameters.

Clearly, Decision Trees are the fastest to learn due to the efficient implementation from \textit{scikit-learn}. Cubist and \toolname have similar training times, and RuleFit lies in between. However, for Cubist and RuleFit, selecting the maximum number of rules requires trying several possibilities. For example, for the extreme case of swimmer (\swimmer), for \toolname, the best maximum depth was just $2$, and we tried up to $5$, whereas for Cubist we tried up to $150$, selecting $110$ as the best, and for RuleFit we tried up to $14$, selecting $14$ as the best. This adds an extra computation time that is not accounted for per single training maximum depth (resp. maximum number of rules).

For Decision Trees, one can see that they tend to overfit: the MSE train keeps decreasing, while the MSE test remains the same or slightly decreases. This seems to be avoided by \toolname, where MSE train and test remain closer. This is probably thanks to the measures that \toolname includes to avoid overfitting, such as early stopping, merging, and feature removal.

\clearpage

\begin{table}
        \tiny
        \centering
        \setlength{\tabcolsep}{1.7pt}
        \renewcommand{\arraystretch}{.4}
\parbox{.49\linewidth}{
\centering
\input{tables/rule_learning/orcaid_mountainCar}
}
\parbox{.49\linewidth}{
\centering
\input{tables/rule_learning/DT_mountainCar}
}
\end{table}

\begin{table}
        \tiny
        \centering
        \setlength{\tabcolsep}{1.7pt}
        \renewcommand{\arraystretch}{.4}
\parbox{.54\linewidth}{
\centering
\input{tables/rule_learning/cubist_mountainCar}
}
\parbox{.45\linewidth}{
\centering
\input{tables/rule_learning/rulefit_mountainCar}
}
\end{table}

\begin{table}
        \tiny
        \centering
        \setlength{\tabcolsep}{1.7pt}
        \renewcommand{\arraystretch}{.4}
\parbox{.49\linewidth}{
\centering
\input{tables/rule_learning/orcaid_pendulum}
}
\parbox{.49\linewidth}{
\centering
\input{tables/rule_learning/dt_pendulum}
}
\end{table}

\begin{table}
        \tiny
        \centering
        \setlength{\tabcolsep}{1.7pt}
        \renewcommand{\arraystretch}{.4}
\parbox{.49\linewidth}{
\centering
\input{tables/rule_learning/cubist_pendulum}
}
\parbox{.49\linewidth}{
\centering
\input{tables/rule_learning/rulefit_pendulum}
}
\end{table}

\begin{table}
        \tiny
        \centering
        \setlength{\tabcolsep}{1.7pt}
        \renewcommand{\arraystretch}{.4}
\parbox{.49\linewidth}{
\centering
\input{tables/rule_learning/orcaid_lunarLander}
}
\parbox{.49\linewidth}{
\centering
\input{tables/rule_learning/dt_lunarLander}
}
\end{table}

\begin{table}
        \tiny
        \centering
        \setlength{\tabcolsep}{1.7pt}
        \renewcommand{\arraystretch}{.4}
\parbox{.5\linewidth}{
\centering
\input{tables/rule_learning/cubist_lunarLander}
}
\parbox{.48\linewidth}{
\centering
\input{tables/rule_learning/rulefit_lunarLander}
}
\end{table}

\begin{table}
        \tiny
        \centering
        \setlength{\tabcolsep}{1.7pt}
        \renewcommand{\arraystretch}{.4}
\parbox{.49\linewidth}{
\centering
\input{tables/rule_learning/orcaid_invPendulum}
}
\parbox{.49\linewidth}{
\centering
\input{tables/rule_learning/dt_invPendulum}
}
\end{table}

\begin{table}
        \tiny
        \centering
        \setlength{\tabcolsep}{1.7pt}
        \renewcommand{\arraystretch}{.4}
\parbox{.52\linewidth}{
\centering
\input{tables/rule_learning/cubist_invPendulum}
}
\parbox{.47\linewidth}{
\centering
\input{tables/rule_learning/rulefit_invPendulum}
}
\end{table}

\begin{table}
        \tiny
        \centering
        \setlength{\tabcolsep}{1.7pt}
        \renewcommand{\arraystretch}{.4}
\parbox{.49\linewidth}{
\centering
\input{tables/rule_learning/orcaid_invDoublePendulum}
}
\parbox{.49\linewidth}{
\centering
\input{tables/rule_learning/dt_invDoublePendulum}
}
\end{table}

\begin{table}
        \tiny
        \centering
        \setlength{\tabcolsep}{1.7pt}
        \renewcommand{\arraystretch}{.4}
\parbox{.52\linewidth}{
\centering
\input{tables/rule_learning/cubist_invDoublePendulum}
}
\parbox{.47\linewidth}{
\centering
\input{tables/rule_learning/rulefit_invDoublePendulum}
}
\end{table}

\begin{table}
        \tiny
        \centering
        \setlength{\tabcolsep}{1.7pt}
        \renewcommand{\arraystretch}{.4}
\parbox{.49\linewidth}{
\centering
\input{tables/rule_learning/orcaid_reacher}
}
\parbox{.49\linewidth}{
\centering
\input{tables/rule_learning/dt_reacher}
}
\end{table}

\begin{table}
        \tiny
        \centering
        \setlength{\tabcolsep}{1.7pt}
        \renewcommand{\arraystretch}{.4}
\parbox{.52\linewidth}{
\centering
\input{tables/rule_learning/cubist_reacher}
}
\parbox{.47\linewidth}{
\centering
\input{tables/rule_learning/rulefit_reacher}
}
\end{table}

\begin{table}
        \tiny
        \centering
        \setlength{\tabcolsep}{1.7pt}
        \renewcommand{\arraystretch}{.4}
\parbox{.49\linewidth}{
\centering
\input{tables/rule_learning/orcaid_swimmer}
}
\parbox{.49\linewidth}{
\centering
\input{tables/rule_learning/dt_swimmer}
}
\end{table}

\begin{table}
        \tiny
        \centering
        \setlength{\tabcolsep}{1.7pt}
        \renewcommand{\arraystretch}{.4}
\parbox{.49\linewidth}{
\centering
\input{tables/rule_learning/cubist_swimmer}
}
\parbox{.49\linewidth}{
\centering
\input{tables/rule_learning/rulefit_swimmer}
}
\end{table}

\begin{table}
        \tiny
        \centering
        \setlength{\tabcolsep}{1.7pt}
        \renewcommand{\arraystretch}{.4}
\parbox{.49\linewidth}{
\centering
\input{tables/rule_learning/orcaid_hopper}
}
\parbox{.49\linewidth}{
\centering
\input{tables/rule_learning/dt_hopper}
}
\end{table}

\begin{table}
        \tiny
        \centering
        \setlength{\tabcolsep}{1.7pt}
        \renewcommand{\arraystretch}{.4}
\parbox{.49\linewidth}{
\centering
\input{tables/rule_learning/cubist_hopper}
}
\parbox{.49\linewidth}{
\centering
\input{tables/rule_learning/rulefit_hopper}
}
\end{table}

\begin{table}
        \tiny
        \centering
        \setlength{\tabcolsep}{1.7pt}
        \renewcommand{\arraystretch}{.4}
\parbox{.49\linewidth}{
\centering
\input{tables/rule_learning/orcaid_halfCheetah}
}
\parbox{.49\linewidth}{
\centering
\input{tables/rule_learning/dt_halfCheetah}
}
\end{table}

\begin{table}
        \tiny
        \centering
        \setlength{\tabcolsep}{1.7pt}
        \renewcommand{\arraystretch}{.4}
\parbox{.49\linewidth}{
\centering
\input{tables/rule_learning/cubist_halfCheetah}
}
\parbox{.49\linewidth}{
\centering
\input{tables/rule_learning/rulefit_halfCheetah}
}
\end{table}

\clearpage

%% file: tables/all_performance_mse_results.tex
\begin{table*}[]
\scriptsize
\setlength{\tabcolsep}{1.2pt}
\centering
\begin{tabular}{l|llll|llll|llll|llll}
             & \multicolumn{4}{c}{\toolname} & \multicolumn{4}{c}{Decision Tree}     & \multicolumn{4}{c}{Cubist}            & \multicolumn{4}{c}{RuleFit} \\
Env.  &  MD & \%   & MSE test   & Size  & MD & \% & MSE test & Size & MNR &  \% & MSE test & Size & MNR & \% & MSE test & Size \\
             \hline
\mountainCar  & 3 & 98$\pm$1 & 17$\pm$3 & 24$\pm$1 & 5 & 99$\pm$0 & 20$\pm$4 & 185$\pm$3 & 8 & 97$\pm$2 & 22$\pm$3 & 49$\pm$0 & 12 & 82$\pm$5 & 36$\pm$5 & 61$\pm$6\\
\pendulum & 3 & 92$\pm$1 & 24$\pm$1 & 59$\pm$2 & 10 & 93$\pm$1 & 37$\pm$1 & 1625$\pm$40 & 30 & 95$\pm$1 & 21$\pm$1 & 250$\pm$1 & 10 & 60$\pm$1 & 104$\pm$3 & 43$\pm$4\\
\lunarLander & 4 & 90$\pm$4 & 71$\pm$1 & 316$\pm$5 & 14 & 39$\pm$7 & 102$\pm$1 & 4099$\pm$75 & 50 & 88$\pm$4 & 64$\pm$1 & 1161$\pm$9 & 21 & 79$\pm$7 & 100$\pm$2 & 137$\pm$10\\
\invPendulum & 2 & 84$\pm$10 & 15$\pm$1 & 19$\pm$1 & 8 & 85$\pm$9 & 7$\pm$1 & 848$\pm$44 & 4 & 82$\pm$12 & 8$\pm$1 & 27$\pm$0 & 14 & 78$\pm$6 & 13$\pm$2 & 61$\pm$7\\
\invDoublePendulum & 3 & 103$\pm$1 & 12$\pm$0 & 146$\pm$4 & 16 & 5$\pm$0 & 42$\pm$1 & 4147$\pm$88 & 120 & 69$\pm$14 & 12$\pm$1 & 1595$\pm$6 & 27 & 7$\pm$1 & 43$\pm$1 & 94$\pm$12\\
\reacher & 4 & 80$\pm$1 & 3$\pm$0 & 457$\pm$2 & 16 & 56$\pm$1 & 7$\pm$0 & 3328$\pm$34 & 150 & 81$\pm$1 & 2$\pm$0 & 4820$\pm$47 & 30 & 42$\pm$1 & 8$\pm$0 & 212$\pm$6\\
\swimmer & 2 & 90$\pm$2 & 59$\pm$4 & 62$\pm$3 & 10 & 95$\pm$1 & 58$\pm$4 & 2384$\pm$86 & 110 & 101$\pm$1 & 36$\pm$3 & 2761$\pm$80 & 14 & 87$\pm$2 & 65$\pm$4 & 109$\pm$5\\
\hopper & 4 & 89$\pm$2 & 51$\pm$1 & 720$\pm$10 & 18 & 17$\pm$3 & 56$\pm$2 & 4238$\pm$67 & 20 & 86$\pm$8 & 46$\pm$1 & 850$\pm$6 & 28 & 85$\pm$3 & 77$\pm$2 & 287$\pm$7\\
\halfCheetah & 6 & 78$\pm$2 & 51$\pm$1 & 3215$\pm$130 & 20 & 47$\pm$4 & 72$\pm$0 & 6032$\pm$194 & 70 & 85$\pm$1 & 42$\pm$2 & 7500$\pm$97 & 32 & 46$\pm$2 & 81$\pm$1 & 721$\pm$21
\end{tabular}
    \caption{Maximum depth for \toolname and Decision Trees (MD), Maximum number of rules for Cubist and RuleFit (MNR), ratio RL rewards/surrogate model rewards, MSE test ($\times10^{-3}$), and model size for \toolname and baselines for all environments.}
    \label{tab:results_performance}
\end{table*}

%% file: tables/rule_learning/orcaid_mountainCar.tex
\begin{tabular}{l|ccccc}
& \multicolumn{5}{c}{Maximum depth}  \\
 & 1 & 2 & 3 & 4 & 5\\\hline
Ratio (\%) & 55$\pm$4 & 91$\pm$4 & 98$\pm$1 & 98$\pm$1 & 98$\pm$1\\
MSE train & 49$\pm$3 & 21$\pm$4 & 11$\pm$2 & 10$\pm$2 & 10$\pm$2\\
MSE test & 56$\pm$7 & 26$\pm$5 & 17$\pm$3 & 17$\pm$4 & 16$\pm$3\\
Model size & 10$\pm$0 & 18$\pm$1 & 24$\pm$1 & 29$\pm$3 & 31$\pm$3\\
Time (s) & 30$\pm$2 & 27$\pm$3 & 20$\pm$2 & 12$\pm$3 & 8$\pm$2\\
\bottomrule
\end{tabular}
\caption{Ratio between the rewards from the surrogate model and the RL policy, MSE for the training and test sets ($\times 10^{-3}$), model size, and learning time for \toolname for \mountainCar.}
\label{tab:orcaid_mc}

%% file: tables/rule_learning/DT_mountainCar.tex
\begin{tabular}{|ccccc}
\multicolumn{5}{c}{Maximum depth}  \\
 1 & 2 & 3 & 4 & 5\\\hline
 15$\pm$0 & 27$\pm$3 & 45$\pm$7 & 57$\pm$9 & 99$\pm$0\\
 105$\pm$6 & 68$\pm$5 & 35$\pm$4 & 19$\pm$2 & 7$\pm$1\\
 109$\pm$8 & 71$\pm$5 & 46$\pm$4 & 33$\pm$3 & 20$\pm$4\\
 6$\pm$0 & 16$\pm$0 & 40$\pm$0 & 92$\pm$1 & 185$\pm$3\\
 16$\pm$2 & 11$\pm$1 & 7$\pm$1 & 6$\pm$1 & 4$\pm$1\\
\bottomrule
\end{tabular}
\caption{Ratio between the rewards from the surrogate model and the RL policy, MSE for the training and test sets ($\times 10^{-3}$), model size, and learning time for DT for \mountainCar.}
\label{tab:dt_mc}

%% file: tables/rule_learning/cubist_mountainCar.tex
\begin{tabular}{l|ccccccc}
& \multicolumn{7}{c}{Maximum number of rules}  \\
 & 1 & 2 & 4 & 5 & 7 & 8 & 10\\\hline
Ratio (\%) & 32$\pm$1 & 46$\pm$1 & 74$\pm$5 & 94$\pm$3 & 97$\pm$2 & 97$\pm$2 & 100$\pm$1\\
MSE train & 98$\pm$3 & 92$\pm$6 & 33$\pm$4 & 26$\pm$2 & 17$\pm$2 & 16$\pm$2 & 12$\pm$2\\
MSE test & 96$\pm$6 & 91$\pm$8 & 40$\pm$4 & 34$\pm$3 & 24$\pm$3 & 22$\pm$3 & 17$\pm$3\\
Model size & 3$\pm$0 & 8$\pm$0 & 21$\pm$0 & 28$\pm$0 & 42$\pm$1 & 49$\pm$0 & 62$\pm$1\\
Time (min) & 3$\pm$1 & 2$\pm$0 & 1$\pm$0 & 1$\pm$0 & 1$\pm$0 & 1$\pm$0 & 1$\pm$0\\
\bottomrule
\end{tabular}
\caption{Ratio between the rewards from the surrogate model and the RL policy, MSE for the training and test sets ($\times 10^{-3}$), model size, and learning time for Cubist for \mountainCar.}
\label{tab:cubist_mc}

%% file: tables/rule_learning/rulefit_mountainCar.tex
\begin{tabular}{|ccccccc}
\multicolumn{7}{c}{Maximum number of rules}  \\
 2 & 4 & 6 & 8 & 10 & 12 & 16\\\hline
 72$\pm$6 & 69$\pm$7 & 80$\pm$5 & 69$\pm$4 & 80$\pm$4 & 82$\pm$5 & 80$\pm$4\\
 40$\pm$5 & 43$\pm$8 & 38$\pm$6 & 45$\pm$5 & 36$\pm$6 & 33$\pm$5 & 32$\pm$4\\
 46$\pm$6 & 49$\pm$8 & 43$\pm$5 & 55$\pm$6 & 43$\pm$6 & 36$\pm$5 & 37$\pm$5\\
 39$\pm$8 & 39$\pm$7 & 39$\pm$4 & 35$\pm$3 & 49$\pm$6 & 61$\pm$6 & 76$\pm$8\\
 16$\pm$3 & 11$\pm$1 & 9$\pm$1 & 11$\pm$1 & 10$\pm$1 & 10$\pm$1 & 13$\pm$2\\
\bottomrule
\end{tabular}
\caption{Ratio between the rewards from the surrogate model and the RL policy, MSE for the training and test sets ($\times 10^{-3}$), model size, and learning time for RuleFit for \mountainCar.}
\label{tab:rulefit_mc}

%% file: tables/rule_learning/orcaid_pendulum.tex
\begin{tabular}{l|ccccc}
& \multicolumn{5}{c}{Maximum depth}  \\
 & 1 & 2 & 3 & 4 & 5\\\hline
Ratio (\%) & 65$\pm$1 & 88$\pm$1 & 92$\pm$1 & 92$\pm$1 & 92$\pm$1\\
MSE train & 73$\pm$2 & 36$\pm$1 & 21$\pm$1 & 18$\pm$1 & 18$\pm$1\\
MSE test & 77$\pm$2 & 40$\pm$1 & 24$\pm$1 & 22$\pm$2 & 22$\pm$1\\
Model size & 13$\pm$0 & 32$\pm$0 & 59$\pm$2 & 92$\pm$5 & 120$\pm$11\\
Time (s) & 39$\pm$2 & 64$\pm$8 & 82$\pm$7 & 82$\pm$10 & 59$\pm$18\\
\bottomrule
\end{tabular}
\caption{Ratio between the rewards from the surrogate model and the RL policy, MSE for the training and test sets ($\times 10^{-3}$), model size, and learning time for \toolname for \pendulum.}
\label{tab:orcaid_p}

%% file: tables/rule_learning/DT_pendulum.tex
\begin{tabular}{|ccccccc}
\multicolumn{7}{c}{Maximum depth}  \\
 1 & 2 & 4 & 5 & 7 & 8 & 10\\\hline
 48$\pm$0 & 54$\pm$0 & 69$\pm$1 & 75$\pm$2 & 89$\pm$1 & 91$\pm$1 & 93$\pm$1\\
 192$\pm$4 & 133$\pm$4 & 71$\pm$1 & 50$\pm$2 & 24$\pm$1 & 20$\pm$1 & 17$\pm$1\\
 192$\pm$3 & 138$\pm$4 & 82$\pm$1 & 64$\pm$2 & 43$\pm$1 & 40$\pm$1 & 37$\pm$1\\
 .01$\pm$.0 & .02$\pm$.0 & .1$\pm$.0 & .2$\pm$.0 & .8$\pm$.0 & 1.0$\pm$.0 & 1.6$\pm$.0\\
 2$\pm$0 & 2$\pm$0 & 3$\pm$0 & 3$\pm$0 & 2$\pm$0 & 2$\pm$0 & 2$\pm$0\\
\bottomrule
\end{tabular}
\caption{Ratio between the rewards from the surrogate model and the RL policy, MSE for the training and test sets ($\times 10^{-3}$), model size, and learning time for DT for \pendulum.}
\label{tab:dt_p}

%% file: tables/rule_learning/cubist_pendulum.tex
\begin{tabular}{l|ccccccc}
& \multicolumn{7}{c}{Maximum number of rules}  \\
 & 1 & 5 & 10 & 15 & 20 & 25 & 30\\\hline
Ratio (\%) & 47$\pm$0 & 66$\pm$1 & 78$\pm$1 & 86$\pm$1 & 91$\pm$1 & 93$\pm$1 & 95$\pm$1\\
MSE train & 201$\pm$5 & 70$\pm$3 & 37$\pm$1 & 28$\pm$1 & 23$\pm$1 & 19$\pm$1 & 15$\pm$1\\
MSE test & 203$\pm$5 & 75$\pm$4 & 44$\pm$2 & 35$\pm$1 & 28$\pm$1 & 24$\pm$1 & 21$\pm$1\\
Model size & 4$\pm$0 & 34$\pm$1 & 74$\pm$1 & 115$\pm$1 & 159$\pm$1 & 205$\pm$1 & 250$\pm$1\\
Time (s) & 24$\pm$2 & 29$\pm$2 & 25$\pm$2 & 29$\pm$4 & 35$\pm$4 & 26$\pm$3 & 29$\pm$4\\
\bottomrule
\end{tabular}
\caption{Ratio between the rewards from the surrogate model and the RL policy, MSE for the training and test sets ($\times 10^{-3}$), model size, and learning time for Cubist for \pendulum.}
\label{tab:cubist_p}

%% file: tables/rule_learning/rulefit_pendulum.tex
\begin{tabular}{|cccccc}
\multicolumn{6}{c}{Maximum number of rules}  \\
 2 & 4 & 6 & 8 & 10 & 12\\\hline
 59$\pm$1 & 61$\pm$2 & 58$\pm$1 & 60$\pm$1 & 60$\pm$1 & 61$\pm$1\\
 122$\pm$9 & 105$\pm$7 & 114$\pm$5 & 103$\pm$4 & 101$\pm$3 & 100$\pm$4\\
 123$\pm$9 & 111$\pm$7 & 116$\pm$4 & 105$\pm$5 & 104$\pm$3 & 104$\pm$4\\
 27$\pm$6 & 35$\pm$8 & 27$\pm$2 & 42$\pm$6 & 43$\pm$4 & 51$\pm$4\\
 5$\pm$0 & 5$\pm$0 & 4$\pm$0 & 5$\pm$1 & 5$\pm$1 & 6$\pm$1\\
\bottomrule
\end{tabular}
\caption{Ratio between the rewards from the surrogate model and the RL policy, MSE for the training and test sets ($\times 10^{-3}$), model size, and learning time for RuleFit for \pendulum.}
\label{tab:rulefit_p}

%% file: tables/rule_learning/orcaid_lunarLander.tex
\begin{tabular}{l|ccccc}
& \multicolumn{5}{c}{Maximum depth}  \\
 & 1 & 2 & 3 & 4 & 5\\\hline
Ratio (\%) & 38$\pm$5 & 49$\pm$5 & 72$\pm$4 & 90$\pm$4 & 92$\pm$3\\
MSE train & 100$\pm$1 & 85$\pm$1 & 70$\pm$1 & 58$\pm$1 & 47$\pm$1\\
MSE test & 102$\pm$1 & 88$\pm$1 & 77$\pm$1 & 71$\pm$1 & 67$\pm$1\\
Model size & 27$\pm$1 & 63$\pm$1 & 141$\pm$2 & 316$\pm$5 & 625$\pm$12\\
Time (min) & 2$\pm$0 & 3$\pm$0 & 5$\pm$0 & 9$\pm$1 & 11$\pm$1\\
\bottomrule
\end{tabular}
\caption{Ratio between the rewards from the surrogate model and the RL policy, MSE for the training and test sets ($\times 10^{-3}$), model size, and learning time for \toolname for \lunarLander.}
\label{tab:orcaid_ll}

%% file: tables/rule_learning/DT_lunarLander.tex
\begin{tabular}{|ccccccc}
\multicolumn{7}{c}{Maximum depth}  \\
 2 & 4 & 8 & 10 & 14 & 16 & 20\\\hline
 -130$\pm$8 & -31$\pm$11 & 23$\pm$6 & 39$\pm$7 & 39$\pm$7 & 38$\pm$7 & 39$\pm$6\\
 170$\pm$1 & 130$\pm$1 & 68$\pm$1 & 58$\pm$1 & 54$\pm$1 & 54$\pm$1 & 55$\pm$1\\
 174$\pm$2 & 141$\pm$3 & 106$\pm$1 & 102$\pm$1 & 102$\pm$1 & 102$\pm$1 & 102$\pm$1\\
 .04$\pm$.0 & .2$\pm$.0 & 1.7$\pm$.0 & 2.7$\pm$.0 & 4.1$\pm$.1 & 4.6$\pm$.1 & 5.1$\pm$.1\\
 22$\pm$5 & 27$\pm$6 & 51$\pm$18 & 66$\pm$17 & 75$\pm$19 & 68$\pm$19 & 59$\pm$19\\
\bottomrule
\end{tabular}
\caption{Ratio between the rewards from the surrogate model and the RL policy, MSE for the training and test sets ($\times 10^{-3}$), model size, and learning time for DT for \lunarLander.}
\label{tab:dt_ll}

%% file: tables/rule_learning/cubist_lunarLander.tex
\begin{tabular}{l|ccccccc}
& \multicolumn{7}{c}{Maximum number of rules}  \\
 & 1 & 2 & 5 & 20 & 30 & 40 & 60\\\hline
Ratio (\%) & 42$\pm$2 & -1$\pm$6 & 43$\pm$8 & 63$\pm$8 & 74$\pm$8 & 85$\pm$5 & 90$\pm$4\\
MSE train & 108$\pm$1 & 104$\pm$1 & 79$\pm$1 & 56$\pm$1 & 51$\pm$1 & 46$\pm$1 & 40$\pm$1\\
MSE test & 110$\pm$2 & 107$\pm$2 & 84$\pm$2 & 68$\pm$2 & 66$\pm$2 & 64$\pm$1 & 64$\pm$1\\
Model size & 14$\pm$1 & 31$\pm$1 & 92$\pm$2 & 429$\pm$6 & 668$\pm$7 & 905$\pm$8 & 1416$\pm$10\\
Time (min) & 16$\pm$2 & 33$\pm$5 & 16$\pm$2 & 10$\pm$1 & 14$\pm$1 & 15$\pm$2 & 12$\pm$2\\
\bottomrule
\end{tabular}
\caption{Ratio between the rewards from the surrogate model and the RL policy, MSE for the training and test sets ($\times 10^{-3}$), model size, and learning time for Cubist for \lunarLander.}
\label{tab:cubist_ll}

%% file: tables/rule_learning/rulefit_lunarLander.tex
\begin{tabular}{|ccccccc}
\multicolumn{7}{c}{Maximum number of rules}  \\
 3 & 6 & 9 & 15 & 18 & 21 & 27\\\hline
72$\pm$10 & 74$\pm$9 & 66$\pm$11 & 85$\pm$6 & 77$\pm$10 & 79$\pm$7 & 80$\pm$6\\
 106$\pm$1 & 104$\pm$2 & 102$\pm$1 & 100$\pm$1 & 96$\pm$1 & 97$\pm$1 & 93$\pm$1\\
 110$\pm$2 & 106$\pm$2 & 105$\pm$2 & 104$\pm$3 & 101$\pm$2 & 100$\pm$2 & 98$\pm$2\\
 47$\pm$6 & 60$\pm$5 & 78$\pm$6 & 107$\pm$6 & 133$\pm$9 & 137$\pm$10 & 186$\pm$8\\
 2$\pm$0 & 2$\pm$0 & 3$\pm$0 & 2$\pm$0 & 2$\pm$0 & 2$\pm$0 & 2$\pm$1\\
\bottomrule
\end{tabular}
\caption{Ratio between the rewards from the surrogate model and the RL policy, MSE for the training and test sets ($\times 10^{-3}$), model size, and learning time for RuleFit for \lunarLander.}
\label{tab:rulefit_ll}

%% file: tables/rule_learning/orcaid_invPendulum.tex
\begin{tabular}{l|ccccc}
& \multicolumn{5}{c}{Maximum depth}  \\
 & 1 & 2 & 3 & 4 & 5\\\hline
Ratio (\%) & 92$\pm$10 & 84$\pm$10 & 84$\pm$10 & 84$\pm$10 & 85$\pm$10\\
MSE train & 18$\pm$2 & 14$\pm$1 & 14$\pm$1 & 14$\pm$1 & 15$\pm$1\\
MSE test & 19$\pm$2 & 15$\pm$1 & 15$\pm$1 & 15$\pm$1 & 16$\pm$2\\
Model size & 13$\pm$1 & 19$\pm$1 & 25$\pm$2 & 28$\pm$4 & 36$\pm$9\\
Time (s) & 63$\pm$12 & 40$\pm$7 & 31$\pm$11 & 23$\pm$7 & 22$\pm$8\\
\bottomrule
\end{tabular}
\caption{Ratio between the rewards from the surrogate model and the RL policy, MSE for the training and test sets ($\times 10^{-3}$), model size, and learning time for \toolname for \invPendulum.}
\label{tab:orcaid_ip}

%% file: tables/rule_learning/cubist_invPendulum.tex
\begin{tabular}{l|ccccccc}
& \multicolumn{7}{c}{Maximum number of rules}  \\
 & 1 & 2 & 4 & 5 & 7 & 8 & 10\\\hline
Ratio (\%) & 101$\pm$5 & 63$\pm$9 & 82$\pm$12 & 85$\pm$11 & 91$\pm$10 & 91$\pm$10 & 98$\pm$6\\
MSE train & 18$\pm$1 & 12$\pm$1 & 8$\pm$1 & 7$\pm$1 & 7$\pm$1 & 6$\pm$1 & 6$\pm$1\\
MSE test & 17$\pm$1 & 12$\pm$1 & 8$\pm$1 & 8$\pm$1 & 8$\pm$1 & 7$\pm$1 & 7$\pm$1\\
Model size & 4$\pm$0 & 12$\pm$0 & 27$\pm$0 & 35$\pm$1 & 50$\pm$1 & 59$\pm$1 & 77$\pm$1\\
Time (min) & 4$\pm$0 & 2$\pm$0 & 2$\pm$0 & 2$\pm$0 & 2$\pm$0 & 1$\pm$0 & 1$\pm$0\\
\bottomrule
\end{tabular}
\caption{Ratio between the rewards from the surrogate model and the RL policy, MSE for the training and test sets ($\times 10^{-3}$), model size, and learning time for Cubist for \invPendulum.}
\label{tab:cubist_ip}

%% file: tables/rule_learning/rulefit_invPendulum.tex
\begin{tabular}{|ccccccc}
\multicolumn{7}{c}{Maximum number of rules}  \\
 2 & 4 & 6 & 8 & 10 & 12 & 14\\\hline
 75$\pm$8 & 73$\pm$7 & 76$\pm$7 & 80$\pm$5 & 77$\pm$4 & 78$\pm$5 & 78$\pm$6\\
 15$\pm$1 & 14$\pm$1 & 13$\pm$1 & 13$\pm$1 & 13$\pm$1 & 13$\pm$1 & 11$\pm$1\\
 16$\pm$2 & 15$\pm$2 & 14$\pm$2 & 15$\pm$2 & 13$\pm$2 & 15$\pm$2 & 13$\pm$2\\
 19$\pm$3 & 24$\pm$4 & 33$\pm$4 & 35$\pm$4 & 42$\pm$3 & 40$\pm$4 & 61$\pm$7\\
 10$\pm$2 & 13$\pm$3 & 9$\pm$1 & 12$\pm$2 & 10$\pm$1 & 11$\pm$1 & 12$\pm$2\\
\bottomrule
\end{tabular}
\caption{Ratio between the rewards from the surrogate model and the RL policy, MSE for the training and test sets ($\times 10^{-3}$), model size, and learning time for RuleFit for \invPendulum.}
\label{tab:rulefit_ip}

%% file: tables/rule_learning/orcaid_invDoublePendulum.tex
\begin{tabular}{l|cccc}
& \multicolumn{4}{c}{Maximum depth}  \\
 & 1 & 2 & 3 & 4\\\hline
Ratio (\%) & 19$\pm$5 & 43$\pm$12 & 103$\pm$1 & 93$\pm$10\\
MSE train & 35$\pm$1 & 19$\pm$0 & 12$\pm$0 & 8$\pm$0\\
MSE test & 35$\pm$1 & 19$\pm$1 & 12$\pm$0 & 10$\pm$0\\
Model size & 23$\pm$1 & 62$\pm$2 & 146$\pm$4 & 313$\pm$8\\
Time (min) & 4$\pm$0 & 3$\pm$0 & 3$\pm$0 & 5$\pm$1\\
\bottomrule
\end{tabular}
\caption{Ratio between the rewards from the surrogate model and the RL policy, MSE for the training and test sets ($\times 10^{-3}$), model size, and learning time for \toolname for \invDoublePendulum.}
\label{tab:orcaid_idp}

%% file: tables/rule_learning/cubist_invDoublePendulum.tex
\begin{tabular}{l|ccccccc}
& \multicolumn{7}{c}{Maximum number of rules}  \\
 & 10 & 30 & 50 & 80 & 100 & 120 & 150\\\hline
Ratio (\%) & 47$\pm$10 & 59$\pm$10 & 61$\pm$9 & 58$\pm$9 & 51$\pm$7 & 69$\pm$7 & 57$\pm$6\\
MSE train & 24$\pm$1 & 15$\pm$1 & 13$\pm$1 & 11$\pm$0 & 10$\pm$1 & 9$\pm$1 & 8$\pm$0\\
MSE test & 25$\pm$1 & 16$\pm$1 & 14$\pm$1 & 12$\pm$1 & 12$\pm$1 & 12$\pm$1 & 11$\pm$1\\
M. size ($\cdot10^3$) & .1$\pm$.0 & .4$\pm$.0 & .6$\pm$.0 & 1.1$\pm$.0 & 1.3$\pm$.0 & 1.6$\pm$.0 & 2.0$\pm$.0\\
Time (min) & 2$\pm$1 & 3$\pm$1 & 3$\pm$1 & 3$\pm$1 & 3$\pm$0 & 4$\pm$1 & 5$\pm$1\\
\bottomrule
\end{tabular}
\caption{Ratio between the rewards from the surrogate model and the RL policy, MSE for the training and test sets ($\times 10^{-3}$), model size, and learning time for Cubist for \invDoublePendulum.}
\label{tab:cubist_idp}

%% file: tables/rule_learning/rulefit_invDoublePendulum.tex
\begin{tabular}{|ccccccc}
\multicolumn{7}{c}{Maximum number of rules}  \\
 3 & 6 & 9 & 15 & 18 & 21 & 27\\\hline
 5$\pm$0 & 6$\pm$1 & 6$\pm$1 & 6$\pm$1 & 6$\pm$1 & 6$\pm$1 & 7$\pm$1\\
 51$\pm$2 & 49$\pm$3 & 48$\pm$2 & 47$\pm$2 & 46$\pm$1 & 45$\pm$2 & 42$\pm$1\\
 51$\pm$2 & 49$\pm$2 & 48$\pm$2 & 47$\pm$2 & 46$\pm$1 & 45$\pm$2 & 43$\pm$1\\
 32$\pm$9 & 46$\pm$12 & 44$\pm$5 & 57$\pm$6 & 66$\pm$5 & 81$\pm$6 & 94$\pm$12\\
 3$\pm$0 & 3$\pm$0 & 3$\pm$0 & 4$\pm$1 & 3$\pm$0 & 3$\pm$0 & 4$\pm$1\\
\bottomrule
\end{tabular}
\caption{Ratio between the rewards from the surrogate model and the RL policy, MSE for the training and test sets ($\times 10^{-3}$), model size, and learning time for RuleFit for \invDoublePendulum.}
\label{tab:ruleft_idp}

%% file: tables/rule_learning/orcaid_reacher.tex
\begin{tabular}{l|ccccc}
& \multicolumn{5}{c}{Maximum depth}  \\
 & 1 & 2 & 3 & 4 & 5\\\hline
Ratio (\%) & 56$\pm$3 & 63$\pm$2 & 71$\pm$1 & 80$\pm$1 & 82$\pm$1\\
MSE train & 6$\pm$0 & 5$\pm$0 & 3$\pm$0 & 3$\pm$0 & 2$\pm$0\\
MSE test & 6$\pm$0 & 5$\pm$0 & 3$\pm$0 & 3$\pm$0 & 2$\pm$0\\
Model size & 39$\pm$1 & 95$\pm$1 & 215$\pm$2 & 457$\pm$2 & 956$\pm$12\\
Time (min) & 1$\pm$0 & 1$\pm$0 & 2$\pm$0 & 5$\pm$1 & 13$\pm$2\\
\bottomrule
\end{tabular}
\caption{Ratio between the rewards from the surrogate model and the RL policy, MSE for the training and test sets ($\times 10^{-3}$), model size, and learning time for \toolname for \reacher.}
\label{tab:orcaid_r}

%% file: tables/rule_learning/cubist_reacher.tex
\begin{tabular}{l|ccccccc}
& \multicolumn{7}{c}{Maximum number of rules}  \\
 & 10 & 30 & 50 & 80 & 100 & 120 & 150\\\hline
Ratio (\%) & 62$\pm$1 & 73$\pm$1 & 75$\pm$2 & 78$\pm$1 & 79$\pm$1 & 80$\pm$1 & 81$\pm$1\\
MSE train & 5$\pm$0 & 3$\pm$0 & 2$\pm$0 & 2$\pm$0 & 2$\pm$0 & 2$\pm$0 & 2$\pm$0\\
MSE test & 5$\pm$0 & 3$\pm$0 & 3$\pm$0 & 3$\pm$0 & 2$\pm$0 & 2$\pm$0 & 2$\pm$0\\
M. size ($\cdot10^3$) & .3$\pm$.0 & .9$\pm$.0 & 1.5$\pm$.0 & 2.5$\pm$.0 & 3.2$\pm$.0 & 3.9$\pm$.0 & 4.8$\pm$.0\\
Time (s) & 64$\pm$8 & 56$\pm$4 & 58$\pm$4 & 70$\pm$8 & 66$\pm$6 & 64$\pm$2 & 66$\pm$2\\
\bottomrule
\end{tabular}
\caption{Ratio between the rewards from the surrogate model and the RL policy, MSE for the training and test sets ($\times 10^{-3}$), model size, and learning time for Cubist for \reacher.}
\label{tab:cubist_r}

%% file: tables/rule_learning/rulefit_reacher.tex
\begin{tabular}{|ccccccc}
\multicolumn{7}{c}{Maximum number of rules}  \\
2 & 6 & 10 & 16 & 20 & 24 & 30\\\hline
36$\pm$1 & 38$\pm$1 & 39$\pm$1 & 39$\pm$1 & 39$\pm$1 & 40$\pm$1 & 42$\pm$1\\
9$\pm$0 & 8$\pm$0 & 8$\pm$0 & 8$\pm$0 & 8$\pm$0 & 8$\pm$0 & 8$\pm$0\\
9$\pm$0 & 8$\pm$0 & 8$\pm$0 & 8$\pm$0 & 8$\pm$0 & 8$\pm$0 & 8$\pm$0\\
42$\pm$4 & 78$\pm$9 & 109$\pm$14 & 132$\pm$9 & 160$\pm$9 & 185$\pm$5 & 212$\pm$6\\
5$\pm$0 & 5$\pm$0 & 5$\pm$0 & 5$\pm$0 & 6$\pm$0 & 6$\pm$0 & 6$\pm$0\\
\bottomrule
\end{tabular}
\caption{Ratio between the rewards from the surrogate model and the RL policy, MSE for the training and test sets ($\times 10^{-3}$), model size, and learning time for RuleFit for \reacher.}
\label{tab:ruefit_r}

%% file: tables/rule_learning/orcaid_swimmer.tex
\begin{tabular}{l|ccccc}
& \multicolumn{5}{c}{Maximum depth}  \\
 & 1 & 2 & 3 & 4 & 5\\\hline
Ratio (\%) & 81$\pm$2 & 90$\pm$2 & 92$\pm$2 & 93$\pm$1 & 95$\pm$2\\
MSE train & 69$\pm$4 & 58$\pm$4 & 51$\pm$4 & 45$\pm$4 & 41$\pm$3\\
MSE test & 69$\pm$4 & 59$\pm$4 & 52$\pm$4 & 47$\pm$4 & 45$\pm$3\\
Model size & 28$\pm$1 & 62$\pm$3 & 120$\pm$8 & 244$\pm$18 & 460$\pm$32\\
Time (min) & 3$\pm$0 & 4$\pm$0 & 5$\pm$1 & 8$\pm$1 & 16$\pm$2\\
\bottomrule
\end{tabular}
\caption{Ratio between the rewards from the surrogate model and the RL policy, MSE for the training and test sets ($\times 10^{-3}$), model size, and learning time for \toolname for \swimmer.}
\label{tab:orcaid_s}

%% file: tables/rule_learning/cubist_swimmer.tex
\begin{tabular}{l|ccccccc}
& \multicolumn{7}{c}{Maximum number of rules}  \\
 & 10 & 30 & 50 & 80 & 100 & 120 & 150\\\hline
Ratio (\%) & 95$\pm$1 & 98$\pm$1 & 99$\pm$1 & 101$\pm$1 & 101$\pm$1 & 101$\pm$1 & 101$\pm$1\\
MSE train & 47$\pm$4 & 37$\pm$3 & 33$\pm$3 & 28$\pm$3 & 26$\pm$3 & 25$\pm$3 & 23$\pm$3\\
MSE test & 49$\pm$4 & 41$\pm$3 & 39$\pm$3 & 37$\pm$3 & 36$\pm$3 & 36$\pm$3 & 36$\pm$3\\
M. size ($\cdot10^3$) & .2$\pm$.0 & .7$\pm$.0 & 1.2$\pm$.0 & 2.0$\pm$.1 & 2.5$\pm$.1 & 3.0$\pm$.1 & 3.8$\pm$.2\\
Time (min) & 11$\pm$1 & 11$\pm$1 & 9$\pm$1 & 11$\pm$1 & 10$\pm$1 & 11$\pm$1 & 11$\pm$1\\
\bottomrule
\end{tabular}
\caption{Ratio between the rewards from the surrogate model and the RL policy, MSE for the training and test sets ($\times 10^{-3}$), model size, and learning time for Cubist for \swimmer.}
\label{tab:cubist_s}

%% file: tables/rule_learning/rulefit_swimmer.tex
\begin{tabular}{|ccccccc}
\multicolumn{7}{c}{Maximum number of rules}  \\
 2 & 4 & 6 & 8 & 10 & 12 & 14\\\hline
82$\pm$3 & 83$\pm$2 & 86$\pm$2 & 86$\pm$2 & 85$\pm$2 & 86$\pm$2 & 87$\pm$2\\
 69$\pm$3 & 70$\pm$3 & 67$\pm$4 & 66$\pm$4 & 68$\pm$4 & 66$\pm$4 & 65$\pm$4\\
 70$\pm$4 & 70$\pm$4 & 68$\pm$4 & 67$\pm$4 & 68$\pm$4 & 66$\pm$4 & 65$\pm$4\\
 51$\pm$5 & 49$\pm$5 & 76$\pm$4 & 87$\pm$7 & 89$\pm$5 & 96$\pm$6 & 109$\pm$5\\
 61$\pm$8 & 58$\pm$7 & 55$\pm$4 & 69$\pm$11 & 64$\pm$8 & 62$\pm$11 & 60$\pm$9\\
\bottomrule
\end{tabular}
\caption{Ratio between the rewards from the surrogate model and the RL policy, MSE for the training and test sets ($\times 10^{-3}$), model size, and learning time for RuleFit for \swimmer.}
\label{tab:rulefit_s}

%% file: tables/rule_learning/orcaid_hopper.tex
\begin{tabular}{l|ccccccc}
& \multicolumn{7}{c}{Maximum depth}  \\
 & 1 & 2 & 3 & 4 & 5 & 6 & 8\\\hline
Ratio (\%) & 72$\pm$6 & 74$\pm$3 & 83$\pm$4 & 90$\pm$2 & 88$\pm$3 & 89$\pm$3 & 88$\pm$2\\
MSE train & 79$\pm$2 & 68$\pm$2 & 57$\pm$1 & 46$\pm$1 & 36$\pm$1 & 29$\pm$1 & 24$\pm$1\\
MSE test & 80$\pm$1 & 69$\pm$2 & 59$\pm$1 & 51$\pm$1 & 44$\pm$1 & 41$\pm$1 & 39$\pm$1\\
M. size ($\cdot10^3$) & .05$\pm$.0 & .1$\pm$.0 & .3$\pm$.0 & .7$\pm$.0 & 1.5$\pm$.0 & 2.6$\pm$.1 & 4.1$\pm$.1\\
Time (min) & 3$\pm$0 & 3$\pm$0 & 7$\pm$1 & 18$\pm$2 & 26$\pm$2 & 84$\pm$13 & 47$\pm$11\\
\bottomrule
\end{tabular}
\caption{Ratio between the rewards from the surrogate model and the RL policy, MSE for the training and test sets ($\times 10^{-3}$), model size, and learning time for \toolname for \hopper.}
\label{tab:orcaid_h}

%% file: tables/rule_learning/cubist_hopper.tex
\begin{tabular}{l|ccccccc}
& \multicolumn{7}{c}{Maximum number of rules}  \\
 & 10 & 20 & 40 & 50 & 70 & 80 & 100\\\hline
Ratio (\%) & 85$\pm$4 & 86$\pm$8 & 94$\pm$2 & 94$\pm$2 & 96$\pm$1 & 90$\pm$3 & 94$\pm$1\\
MSE train & 55$\pm$2 & 42$\pm$1 & 32$\pm$1 & 29$\pm$1 & 25$\pm$1 & 23$\pm$1 & 21$\pm$1\\
MSE test & 58$\pm$2 & 46$\pm$1 & 38$\pm$1 & 37$\pm$1 & 35$\pm$1 & 34$\pm$1 & 34$\pm$1\\
M. size ($\cdot10^3$) & .4$\pm$.0 & .9$\pm$.0 & 1.8$\pm$.0 & 2.2$\pm$.0 & 3.2$\pm$.0 & 3.7$\pm$.0 & 4.7$\pm$.0\\
Time (min) & 17$\pm$2 & 22$\pm$5 & 19$\pm$3 & 19$\pm$3 & 16$\pm$2 & 16$\pm$1 & 22$\pm$3\\
\bottomrule
\end{tabular}
\caption{Ratio between the rewards from the surrogate model and the RL policy, MSE for the training and test sets ($\times 10^{-3}$), model size, and learning time for Cubist for \hopper.}
\label{tab:cubist_h}

%% file: tables/rule_learning/rulefit_hopper.tex
\begin{tabular}{|ccccccc}
\multicolumn{7}{c}{Maximum number of rules}  \\
  2 & 8 & 14 & 20 & 26 & 32 & 40\\\hline
 77$\pm$6 & 76$\pm$6 & 87$\pm$4 & 85$\pm$5 & 83$\pm$4 & 87$\pm$2 & 86$\pm$3\\
 85$\pm$2 & 81$\pm$2 & 79$\pm$1 & 77$\pm$1 & 76$\pm$1 & 77$\pm$1 & 76$\pm$1\\
 86$\pm$2 & 81$\pm$2 & 79$\pm$2 & 78$\pm$2 & 78$\pm$1 & 78$\pm$1 & 77$\pm$2\\
 65$\pm$6 & 139$\pm$14 & 191$\pm$13 & 217$\pm$9 & 305$\pm$8 & 318$\pm$15 & 342$\pm$22\\
 1$\pm$0 & 1$\pm$0 & 2$\pm$0 & 1$\pm$0 & 2$\pm$0 & 2$\pm$0 & 2$\pm$0\\
\bottomrule
\end{tabular}
\caption{Ratio between the rewards from the surrogate model and the RL policy, MSE for the training and test sets ($\times 10^{-3}$), model size, and learning time for RuleFit for \hopper.}
\label{tab:rulefit_h}

%% file: tables/rule_learning/orcaid_halfCheetah.tex
\begin{tabular}{l|cccccc}
& \multicolumn{6}{c}{Maximum depth}  \\
 & 1 & 2 & 3 & 4 & 5 & 6\\\hline
Ratio (\%) & 18$\pm$5 & 32$\pm$8 & 50$\pm$5 & 60$\pm$5 & 75$\pm$4 & 84$\pm$2\\
MSE train & 93$\pm$1 & 81$\pm$1 & 71$\pm$1 & 62$\pm$1 & 53$\pm$1 & 47$\pm$1\\
MSE test & 93$\pm$1 & 82$\pm$1 & 72$\pm$1 & 63$\pm$1 & 56$\pm$1 & 51$\pm$1\\
Model size & 101$\pm$1 & 219$\pm$4 & 478$\pm$9 & 1011$\pm$20 & 2019$\pm$70 & 3215$\pm$130\\
Time (min) & 9$\pm$4 & 13$\pm$2 & 15$\pm$2 & 19$\pm$2 & 25$\pm$6 & 30$\pm$12\\
\bottomrule
\end{tabular}
\caption{Ratio between the rewards from the surrogate model and the RL policy, MSE for the training and test sets ($\times 10^{-3}$), model size, and learning time for \toolname for \halfCheetah.}
\label{tab:orcaid_hc}

%% file: tables/rule_learning/cubist_halfCheetah.tex
\begin{tabular}{l|ccccccc}
& \multicolumn{7}{c}{Maximum number of rules}  \\
 & 10 & 20 & 40 & 50 & 70 & 80 & 100\\\hline
Ratio (\%) & 57$\pm$10 & 77$\pm$3 & 80$\pm$3 & 83$\pm$1 & 85$\pm$1 & 85$\pm$1 & 86$\pm$1\\
MSE train & 60$\pm$2 & 51$\pm$2 & 42$\pm$2 & 40$\pm$2 & 35$\pm$1 & 33$\pm$1 & 31$\pm$0\\
MSE test & 62$\pm$3 & 54$\pm$3 & 46$\pm$2 & 45$\pm$2 & 42$\pm$2 & 40$\pm$1 & 39$\pm$1\\
M. size ($\cdot10^3$) & 1.1$\pm$.0 & 2.2$\pm$.0 & 4.3$\pm$.0 & 5.4$\pm$.1 & 7.5$\pm$.1 & 8.6$\pm$.1 & 10.7$\pm$.2\\
Time (min) & 18$\pm$1 & 28$\pm$10 & 20$\pm$1 & 26$\pm$7 & 21$\pm$1 & 21$\pm$2 & 23$\pm$1\\
\bottomrule
\end{tabular}
\caption{Ratio between the rewards from the surrogate model and the RL policy, MSE for the training and test sets ($\times 10^{-3}$), model size, and learning time for Cubist for \halfCheetah.}
\label{tab:cubist_hc}

%% file: tables/rule_learning/rulefit_halfCheetah.tex
\begin{tabular}{|ccccccc}
\multicolumn{7}{c}{Maximum number of rules}  \\
 2 & 8 & 14 & 20 & 26 & 32 & 40\\\hline
36$\pm$2 & 40$\pm$2 & 45$\pm$1 & 45$\pm$1 & 46$\pm$2 & 46$\pm$2 & 48$\pm$2\\
93$\pm$1 & 89$\pm$1 & 85$\pm$0 & 84$\pm$1 & 83$\pm$1 & 81$\pm$0 & 80$\pm$0\\
93$\pm$1 & 89$\pm$1 & 85$\pm$1 & 84$\pm$0 & 83$\pm$1 & 81$\pm$1 & 81$\pm$1\\
183$\pm$10 & 271$\pm$4 & 405$\pm$14 & 507$\pm$8 & 579$\pm$11 & 721$\pm$21 & 820$\pm$18\\
 2$\pm$0 & 1$\pm$0 & 1$\pm$0 & 2$\pm$0 & 2$\pm$0 & 2$\pm$0 & 2$\pm$0\\
\bottomrule
\end{tabular}
\caption{Ratio between the rewards from the surrogate model and the RL policy, MSE for the training and test sets ($\times 10^{-3}$), model size, and learning time for RuleFit for \halfCheetah.}
\label{tab:rulefit_hc}

%% file: appendixAblation.tex
\input{tables/ablation_study}

In this section, we learn different \toolname models by modifying the hyperparameters shown in Appendix~\ref{sec:Apphyperparams}, Table~\ref{tab:orcaid_hyperparams}:
\begin{compactitem}[•]
    \item Number of steps collected during the interaction between the original RL policy and the environment (\#steps).
    \item Noise added to the clusters when sampling ($\sigma$).
    \item MSE threshold to stop growing a branch (\texttt{max\_mse}).
    \item Number of clusters ($k$).
    \item Number of samples that we sample from the clusters (\#samples).
\end{compactitem}
We also learn models without certain components of our approach:
\begin{compactitem}[•]
    \item Instead of linear regression in the head of the rules, we assign the mean of the observed actions for that region per action dimension.
    \item We do not include the DAgger loop.
    \item We do not simplify adjacent regions.
\end{compactitem}
We keep all other components fixed using the values from Appendix~\ref{sec:Apphyperparams}, Table~\ref{tab:orcaid_hyperparams}, as \textit{default} and change only one of the mentioned hyperparameters or components per learned model. For this study, we selected the environments \pendulum\ and \swimmer\ to include both a simpler and a more complex environment. The default parameters for
\begin{compactitem}[•]
    \item \pendulum\ are \#steps=20k, $\sigma=0.15$, \texttt{max\_mse}=0.2, $k=15$, and \#samples=1k; and for
    \item \swimmer\ are \#steps=100k, $\sigma=0.15$, \texttt{max\_mse}=0.05, $k=100$, and \#samples=20k.
\end{compactitem} 

The maximum depth selected for \pendulum\ in the paper was $3$, and $2$ for \swimmer.

Table~\ref{tab:abl_study_pendulum_ratio} shows \toolname's reward ratio for \pendulum\  when varying the different components as described, keeping the column \textit{default} as a reference for the \toolname model with the hyperparameters and approach as presented in the paper. Table~\ref{tab:abl_study_pendulum_size} shows \toolname's model sizes for every ablation. Similarly, Table~\ref{tab:abl_study_swimmer_ratio} and Table~\ref{tab:abl_study_swimmer_size} show \toolname's reward ratios and model sizes for \swimmer\ for every ablation.

Reducing the \texttt{max\_mse} makes \toolname adjust better or equally to the resampled data. However, this comes at the cost of needing more regions as the number of merged regions decreases, and it also carries a potential for overfitting. This is also what happens when no simplification is included, i.e., the model still learns, but its size grows.

Using fewer clusters causes \toolname to miss trajectory segments (points become over-grouped), reducing performance. The models become smaller since the data points are more concentrated around the clusters. A similar situation occurs when the number of samples is insufficient, i.e., some important points may be excluded from the training process. The model is smaller but has worse performance.

Removing leaf regressions reduces adaptability, requiring deeper trees to match the original performance. Linear regressions help train models with strong performance more quickly by avoiding the need to create more regions.

The absence of the DAgger loop makes the model learn more slowly, needing more depth to reach the original performance. This is emphasized in \swimmer, where it is not able to learn a good enough model.

Finally, having no linear regression and no oblique decisions creates models similar to DTs. Hence, the comparisons to DTs in the main part of the paper constitute another ablation study.

\clearpage
\vfill

%% file: tables/ablation_study.tex
\begin{table}[]
    \centering
    \footnotesize
    \setlength{\tabcolsep}{2.2pt}
    \begin{tabular}{l|c|cccccccc}
          max. depth & default & \#steps=1k & $\sigma\!=\!0.05$ & \texttt{max\_mse}$=\!0.05$ & $k=5$ & \#samples$=\!1k$ & No lin. reg. & No DAgger & No simpl. \\
         \hline 
          1 & 65 & 63 & 60 & 65 & 55 & 62 & 60 & 65 & 65\\
          2 & 88 & 70 & 87 & 88 & 57 & 65 & 65 & 88 & 87\\
          3 & 92 & 84 & 90 & 92 & 68 & 76 & 78 & 92 & 90\\
          4 & 92 & 86 & 92 & 94 & 80 & 89 & 85 & 92 & 92\\
         \hline
    \end{tabular}
    \caption{Ablation study for \pendulum. Ratio (\%) between the rewards from the surrogate model and the RL policy. Only one hyperparameter is modified per column.}
    \label{tab:abl_study_pendulum_ratio}
\end{table}

\begin{table}[]
    \centering
    \footnotesize
    \setlength{\tabcolsep}{2.2pt}
    \begin{tabular}{l|c|cccccccc}
          max. depth & selected & \#steps=1k & $\sigma\!=\!0.05$ & \texttt{max\_mse}$=\!0.05$ & $k=5$ & \#samples$=\!1k$ & No lin. reg. & No DAgger & No simpl. \\
         \hline 
          1 & 13 & 13 & 13 & 13 & 13 & 13 & 13 & 13 & 13 \\
          2 & 32 & 29 & 30 & 32 & 32 & 30 & 32 & 30 & 32\\
          3 & 59 & 51 & 50 & 65 & 59 & 49 & 67 & 55 & 69\\
          4 & 92 & 80 & 79 & 120 & 92 & 75 & 121 & 88 & 130\\
         \hline
    \end{tabular}
    \caption{Ablation study for \pendulum. Model size. Only one hyperparameter is modified per column.}
    \label{tab:abl_study_pendulum_size}
\end{table}

\begin{table}[]
    \centering
    \footnotesize
    \setlength{\tabcolsep}{2.2pt}
    \begin{tabular}{l|c|cccccccc}
          max. depth & selected & \#steps=5k & $\sigma\!=\!0.05$ & \texttt{max\_mse}$=\!0.01$ & $k=5$ & \#samples$=\!5k$ & No lin. reg. & No DAgger & No simpl. \\
         \hline 
          1 & 81 & 78 & 80 & 81 & 60 & 68 & 78 & 70 & 81\\
          2 & 90 & 82 & 89 & 91 & 70 & 72 & 83 & 72 & 89\\
          3 & 92 & 86 & 92 & 93 & 72 & 75 & 88 & 73 & 91\\
         \hline
    \end{tabular}
    \caption{Ablation study for \swimmer. Ratio (\%) between the rewards from the surrogate model and the RL policy. Only one hyperparameter is modified per column.}
    \label{tab:abl_study_swimmer_ratio}
\end{table}

\begin{table}[]
    \centering
    \footnotesize
    \setlength{\tabcolsep}{2.2pt}
    \begin{tabular}{l|c|cccccccc}
          max. depth & selected & \#steps=5k & $\sigma\!=\!0.05$ & \texttt{max\_mse}$=\!0.01$ & $k=5$ & \#samples$=\!5k$ & No lin. reg. & No DAgger & No simpl. \\
         \hline 
          1 & 28 & 20 & 26 & 30 & 22 & 18 & 32 & 18 & 32\\
          2 & 62 & 42 & 58 & 72 & 50 & 45 & 78 & 38 & 78\\
          3 & 120 & 85 & 115 & 150 & 100 & 85 & 150 & 65 & 155\\
         \hline
    \end{tabular}
    \caption{Ablation study for \swimmer. Model size. Only one hyperparameter is modified per column.}
    \label{tab:abl_study_swimmer_size}
\end{table}

%% file: main.bib
@article{MilaniTVF24,
  author       = {Stephanie Milani and
                  Nicholay Topin and
                  Manuela Veloso and
                  Fei Fang},
  title        = {Explainable Reinforcement Learning: {A} Survey and Comparative Review},
  journal      = {{ACM} Comput. Surv.},
  volume       = {56},
  number       = {7},
  pages        = {168:1--168:36},
  year         = {2024},
  url          = {https://doi.org/10.1145/3616864},
  doi          = {10.1145/3616864},
  bibsource    = {dblp computer science bibliography, https://dblp.org}
}

@inproceedings{DBLP:conf/ijcai/TapplerLTB25,
  author       = {Martin Tappler and
                  Ignacio D. Lopez{-}Miguel and
                  Sebastian Tschiatschek and
                  Ezio Bartocci},
  title        = {Rule-Guided Reinforcement Learning Policy Evaluation and Improvement},
  booktitle    = {{IJCAI} 2025},
  pages        = {6254--6262},
  publisher    = {ijcai.org},
  year         = {2025},
  url          = {https://doi.org/10.24963/ijcai.2025/696},
  doi          = {10.24963/IJCAI.2025/696},
  bibsource    = {dblp computer science bibliography, https://dblp.org}
}

@article{DBLP:journals/nature/MnihKSRVBGRFOPB15,
  author       = {Volodymyr {Mnih {\em et al.}}},
  title        = {Human-level control through deep reinforcement learning},
  journal      = {Nat.},
  volume       = {518},
  number       = {7540},
  pages        = {529--533},
  year         = {2015},
  url          = {https://doi.org/10.1038/nature14236},
  doi          = {10.1038/NATURE14236},
  bibsource    = {dblp computer science bibliography, https://dblp.org}
}

@article{DBLP:journals/nature/SilverHMGSDSAPL16,
  author       = {David {Silver {\em et al.}}},
  title        = {Mastering the game of {Go} with deep neural networks and tree search},
  journal      = {Nat.},
  volume       = {529},
  number       = {7587},
  pages        = {484--489},
  year         = {2016},
  url          = {https://doi.org/10.1038/nature16961},
  doi          = {10.1038/NATURE16961},
  bibsource    = {dblp computer science bibliography, https://dblp.org}
}

@article{DBLP:journals/nature/DegraveFBNTCEHA22,
  author       = {Jonas {Degrave {\em et al.}}},
  title        = {Magnetic control of tokamak plasmas through deep reinforcement learning},
  journal      = {Nat.},
  volume       = {602},
  number       = {7897},
  pages        = {414--419},
  year         = {2022},
  url          = {https://doi.org/10.1038/s41586-021-04301-9},
  doi          = {10.1038/S41586-021-04301-9},
  bibsource    = {dblp computer science bibliography, https://dblp.org}
}

@article{DBLP:journals/corr/abs-2307-09288,
    author = {{Touvron et al.}},
  comauthor       = {Hugo Touvron and
                  Louis Martin and
                  Kevin Stone and
                  Peter Albert and
                  Amjad Almahairi and
                  Yasmine Babaei and
                  Nikolay Bashlykov and
                  Soumya Batra and
                  Prajjwal Bhargava and
                  Shruti Bhosale and
                  Dan Bikel and
                  Lukas Blecher and
                  Cristian Canton{-}Ferrer and
                  Moya Chen and
                  Guillem Cucurull and
                  David Esiobu and
                  Jude Fernandes and
                  Jeremy Fu and
                  Wenyin Fu and
                  Brian Fuller and
                  Cynthia Gao and
                  Vedanuj Goswami and
                  Naman Goyal and
                  Anthony Hartshorn and
                  Saghar Hosseini and
                  Rui Hou and
                  Hakan Inan and
                  Marcin Kardas and
                  Viktor Kerkez and
                  Madian Khabsa and
                  Isabel Kloumann and
                  Artem Korenev and
                  Punit Singh Koura and
                  Marie{-}Anne Lachaux and
                  Thibaut Lavril and
                  Jenya Lee and
                  Diana Liskovich and
                  Yinghai Lu and
                  Yuning Mao and
                  Xavier Martinet and
                  Todor Mihaylov and
                  Pushkar Mishra and
                  Igor Molybog and
                  Yixin Nie and
                  Andrew Poulton and
                  Jeremy Reizenstein and
                  Rashi Rungta and
                  Kalyan Saladi and
                  Alan Schelten and
                  Ruan Silva and
                  Eric Michael Smith and
                  Ranjan Subramanian and
                  Xiaoqing Ellen Tan and
                  Binh Tang and
                  Ross Taylor and
                  Adina Williams and
                  Jian Xiang Kuan and
                  Puxin Xu and
                  Zheng Yan and
                  Iliyan Zarov and
                  Yuchen Zhang and
                  Angela Fan and
                  Melanie Kambadur and
                  Sharan Narang and
                  Aur{\'{e}}lien Rodriguez and
                  Robert Stojnic and
                  Sergey Edunov and
                  Thomas Scialom},
  title        = {Llama 2: Open Foundation and Fine-Tuned Chat Models},
  journal      = {CoRR},
  volume       = {abs/2307.09288},
  year         = {2023},
  url          = {https://doi.org/10.48550/arXiv.2307.09288},
  doi          = {10.48550/ARXIV.2307.09288},
  eprinttype    = {arXiv},
  eprint       = {2307.09288},
  bibsource    = {dblp computer science bibliography, https://dblp.org}
}

@inproceedings{DBLP:journals/corr/LillicrapHPHETS15,
  author       = {Timothy P. {Lillicrap {\em et al.}}},
  editor       = {Yoshua Bengio and
                  Yann LeCun},
  title        = {Continuous control with deep reinforcement learning},
  booktitle    = {4th International Conference on Learning Representations, {ICLR} 2016,
                  San Juan, Puerto Rico, May 2-4, 2016, Conference Track Proceedings},
  year         = {2016},
  url          = {http://arxiv.org/abs/1509.02971},
  bibsource    = {dblp computer science bibliography, https://dblp.org}
}

@inproceedings{DBLP:conf/rss/PalejaNSRCG22,
  author       = {Rohan R. Paleja and
                  Yaru Niu and
                  Andrew Silva and
                  Chace Ritchie and
                  Sugju Choi and
                  Matthew C. Gombolay},
  comeditor       = {Kris Hauser and
                  Dylan A. Shell and
                  Shoudong Huang},
  title        = {Learning Interpretable, High-Performing Policies for Autonomous Driving},
  booktitle    = {Robotics: Science and Systems XVIII, 2022},
  year         = {2022},
  url          = {https://doi.org/10.15607/RSS.2022.XVIII.068},
  doi          = {10.15607/RSS.2022.XVIII.068},
  bibsource    = {dblp computer science bibliography, https://dblp.org}
}

@article{DhebarDNZF24,
  author       = {Yashesh D. Dhebar and
                  Kalyanmoy Deb and
                  Subramanya Nageshrao and
                  Ling Zhu and
                  Dimitar P. Filev},
  title        = {Toward Interpretable-{AI} Policies Using Evolutionary Nonlinear Decision
                  Trees for Discrete-Action Systems},
  journal      = {{IEEE} Trans. Cybern.},
  volume       = {54},
  number       = {1},
  pages        = {50--62},
  year         = {2024},
  url          = {https://doi.org/10.1109/TCYB.2022.3180664},
  doi          = {10.1109/TCYB.2022.3180664},
  bibsource    = {dblp computer science bibliography, https://dblp.org}
}

@article{DAI2022107932,
title = {Enhanced Oblique Decision Tree Enabled Policy Extraction for Deep Reinforcement Learning in Power System Emergency Control},
journal = {Electric Power Systems Research},
volume = {209},
pages = {107932},
year = {2022},
issn = {0378-7796},
doi = {https://doi.org/10.1016/j.epsr.2022.107932},
url = {https://www.sciencedirect.com/science/article/pii/S0378779622001626},
author = {Yuxin {Dai {\em et al.}}}
}

@article{HeinUR18,
  author       = {Daniel Hein and
                  Steffen Udluft and
                  Thomas A. Runkler},
  title        = {Interpretable policies for reinforcement learning by genetic programming},
  journal      = {Eng. Appl. Artif. Intell.},
  volume       = {76},
  pages        = {158--169},
  year         = {2018},
  url          = {https://doi.org/10.1016/j.engappai.2018.09.007},
  doi          = {10.1016/J.ENGAPPAI.2018.09.007},
  bibsource    = {dblp computer science bibliography, https://dblp.org}
}

@inproceedings{SilvaGKJS20,
  author       = {Andrew Silva and
                  Matthew C. Gombolay and
                  Taylor W. Killian and
                  Ivan Dario Jimenez Jimenez and
                  Sung{-}Hyun Son},
  comeditor       = {Silvia Chiappa and
                  Roberto Calandra},
  title        = {Optimization Methods for Interpretable Differentiable Decision Trees
                  Applied to Reinforcement Learning},
  booktitle    = {
                  {AISTATS} 2020},
  series       = {Proceedings of Machine Learning Research},
  volume       = {108},
  pages        = {1855--1865},
  publisher    = {{PMLR}},
  year         = {2020},
  url          = {http://proceedings.mlr.press/v108/silva20a.html},
  bibsource    = {dblp computer science bibliography, https://dblp.org}
}

@book{BreimanFOS84,
  author       = {Leo Breiman and
                  J. H. Friedman and
                  Richard A. Olshen and
                  C. J. Stone},
  title        = {Classification and Regression Trees},
  publisher    = {Wadsworth},
  year         = {1984},
  isbn         = {0-534-98053-8},
  bibsource    = {dblp computer science bibliography, https://dblp.org}
}

@inproceedings{BastaniPS18,
  author       = {Osbert Bastani and
                  Yewen Pu and
                  Armando Solar{-}Lezama},
  editor       = {Samy Bengio and
                  Hanna M. Wallach and
                  Hugo Larochelle and
                  Kristen Grauman and
                  Nicol{\`{o}} Cesa{-}Bianchi and
                  Roman Garnett},
  title        = {Verifiable Reinforcement Learning via Policy Extraction},
  booktitle    = {NeurIPS 2018},
  pages        = {2499--2509},
  year         = {2018},
  url          = {https://proceedings.neurips.cc/paper/2018/hash/e6d8545daa42d5ced125a4bf747b3688-Abstract.html},
  bibsource    = {dblp computer science bibliography, https://dblp.org}
}

@inproceedings{McCalmonLA022,
  author       = {Joe McCalmon and
                  Thai Le and
                  Sarra M. Alqahtani and
                  Dongwon Lee},
  comeditor       = {Piotr Faliszewski and
                  Viviana Mascardi and
                  Catherine Pelachaud and
                  Matthew E. Taylor},
  title        = {{CAPS:} Comprehensible Abstract Policy Summaries for Explaining Reinforcement
                  Learning Agents},
  booktitle    = { {AAMAS} 2022},
  pages        = {889--897},
  publisher    = {
                  {IFAAMAS}},
  year         = {2022},
  url          = {https://www.ifaamas.org/Proceedings/aamas2022/pdfs/p889.pdf},
  doi          = {10.5555/3535850.3535950},
  bibsource    = {dblp computer science bibliography, https://dblp.org}
}

@inproceedings{TopinV19,
  author       = {Nicholay Topin and
                  Manuela Veloso},
  title        = {Generation of Policy-Level Explanations for Reinforcement Learning},
  booktitle    = {{AAAI}
                  2019},
  pages        = {2514--2521},
  publisher    = {{AAAI} Press},
  year         = {2019},
  url          = {https://doi.org/10.1609/aaai.v33i01.33012514},
  doi          = {10.1609/AAAI.V33I01.33012514},
  bibsource    = {dblp computer science bibliography, https://dblp.org}
}

@inproceedings{SreedharanSK20,
  author       = {Sarath Sreedharan and
                  Siddharth Srivastava and
                  Subbarao Kambhampati},
  comeditor       = {J. Christopher Beck and
                  Olivier Buffet and
                  J{\"{o}}rg Hoffmann and
                  Erez Karpas and
                  Shirin Sohrabi},
  title        = {{TLdR}: Policy Summarization for Factored {SSP} Problems Using Temporal
                  Abstractions},
  booktitle    = {ICAPS 2020},
  pages        = {272--280},
  publisher    = {{AAAI} Press},
  year         = {2020},
  url          = {https://ojs.aaai.org/index.php/ICAPS/article/view/6671},
  bibsource    = {dblp computer science bibliography, https://dblp.org}
}

@inproceedings{DaneshKFK21,
  author       = {Mohamad H. Danesh and
                  Anurag Koul and
                  Alan Fern and
                  Saeed Khorram},
  comeditor       = {Marina Meila and
                  Tong Zhang},
  title        = {Re-understanding Finite-State Representations of Recurrent Policy
                  Networks},
  booktitle    = {
                  {ICML} 2021},
  series       = {Proceedings of Machine Learning Research},
  volume       = {139},
  pages        = {2388--2397},
  publisher    = {{PMLR}},
  year         = {2021},
  url          = {http://proceedings.mlr.press/v139/danesh21a.html},
  bibsource    = {dblp computer science bibliography, https://dblp.org}
}

@inproceedings{KoulFG19,
  author       = {Anurag Koul and
                  Alan Fern and
                  Sam Greydanus},
  title        = {Learning Finite State Representations of Recurrent Policy Networks},
  booktitle    = {{ICLR}},
  publisher    = {OpenReview.net},
  year         = {2019},
  url          = {https://openreview.net/forum?id=S1gOpsCctm},
  bibsource    = {dblp computer science bibliography, https://dblp.org}
}

@inproceedings{10.5555/3709347.3743863,
author = {Yang, Yue and Yang, Fan and Bai, Yu and Wang, Hao},
title = {Self-Interpretable Reinforcement Learning via Rule Ensembles},
year = {2025},
isbn = {9798400714269},
publisher = {International Foundation for Autonomous Agents and Multiagent Systems},
comaddress = {Richland, SC},
booktitle = {AAMAS 2025},
pages = {2235–2243},
numpages = {9},
comlocation = {Detroit, MI, USA},
comseries = {AAMAS '25}
}

@inproceedings{Coppens2019,
title = "Distilling Deep Reinforcement Learning Policies in Soft Decision Trees",
author = "Youri Coppens and Kyriakos Efthymiadis and Tom Lenaerts and Ann Nowe",
year = "2019",
month = aug,
day = "11",
language = "English",
pages = "1--6",
comeditor = "Tim Miller and Rosina Weber and Daniele Magazzeni",
booktitle = "IJCAI 2019 Workshop on Explainable Artificial Intelligence",
comnote = "IJCAI 2019 Workshop on Explainable Artificial Intelligence, XAI19 ; Conference date: 11-08-2019",
url = "https://sites.google.com/view/xai2019/home",
}

@inproceedings{DBLP:conf/icml/SilverLHDWR14,
  author       = {David Silver and
                  Guy Lever and
                  Nicolas Heess and
                  Thomas Degris and
                  Daan Wierstra and
                  Martin A. Riedmiller},
  title        = {Deterministic Policy Gradient Algorithms},
  booktitle    = {
                  {ICML} 2014},
  series       = {{JMLR} Workshop and Conference Proceedings},
  volume       = {32},
  pages        = {387--395},
  publisher    = {JMLR.org},
  year         = {2014},
  url          = {http://proceedings.mlr.press/v32/silver14.html},
  bibsource    = {dblp computer science bibliography, https://dblp.org}
}

@article{friedman2008ruleensembles,
  author    = {Jerome H. Friedman and Bogdan E. Popescu},
  title     = {Predictive learning via rule ensembles},
  journal   = {The Annals of Applied Statistics},
  volume    = {2},
  number    = {3},
  pages     = {916--954},
  year      = {2008},
  month     = sep,
  doi       = {DOI: 10.1214/07-AOAS148}
}

@article{Dunnett,
author = {Charles W. Dunnett},
title = {A Multiple Comparison Procedure for Comparing Several Treatments with a Control},
journal = {Journal of the American Statistical Association},
volume = {50},
number = {272},
pages = {1096--1121},
year = {1955},
publisher = {Taylor \& Francis},
doi = {10.1080/01621459.1955.10501294},
URL = {   https://amstat.tandfonline.com/doi/abs/10.1080/01621459.1955.10501294
},
eprint = { 
  https://amstat.tandfonline.com/doi/pdf/10.1080/01621459.1955.10501294
}

}

@article{pb_dunnett,
  author  = {Sarah Alver, Guoyi Zhang},
  title   = {Multiple comparisons of treatment against control under unequal variances using parametric bootstrap},
  journal = {J. Appl. Stat.},
  year    = {2023},
  volume  = {51},
  number  = {10},
  pages   = {1861--1877},
  doi     = {10.1080/02664763.2023.2245179},
}

@inproceedings{FurnkranzK15,
  author       = {Johannes F{\"{u}}rnkranz and
                  Tom{\'{a}}s Kliegr},
  comeditor       = {Nick Bassiliades and
                  Georg Gottlob and
                  Fariba Sadri and
                  Adrian Paschke and
                  Dumitru Roman},
  title        = {A Brief Overview of Rule Learning},
  booktitle    = {RuleML 2015},
  series       = {LNCS},
  volume       = {9202},
  pages        = {54--69},
  publisher    = {Springer},
  year         = {2015},
  url          = {https://doi.org/10.1007/978-3-319-21542-6\_4},
  doi          = {10.1007/978-3-319-21542-6\_4},
  bibsource    = {dblp computer science bibliography, https://dblp.org}
}

@book{Sutton:1998,
  author = {Sutton, Richard S. and Barto, Andrew G.},
  edition = {Second},
  publisher = {The MIT Press},
  title = {Reinforcement Learning: An Introduction},
  year = {2018 }
}

@misc{pycubist,  
  title = {Python package for fitting Quinlan's Cubist v2.07 regression model},  
  author = {{Aselin et al.}},
  year = {2021},  
  url = {https://github.com/pjaselin/Cubist}  
}

@inproceedings{Quinlan1992LearningWC,
  title={Learning With Continuous Classes},
  author={J. Ross Quinlan},
  year={1992},
booktitle={Proceedings of Australian Joint Conference on Artificial Intelligence}
}

@article{scikit-learn,
  title={Scikit-learn: Machine Learning in {P}ython},
  author={Pedregosa, F. and Varoquaux, G. and Gramfort, A. and Michel, V.
          and Thirion, B. and Grisel, O. and Blondel, M. and Prettenhofer, P.
          and Weiss, R. and Dubourg, V. and Vanderplas, J. and Passos, A. and
          Cournapeau, D. and Brucher, M. and Perrot, M. and Duchesnay, E.},
  journal={Journal of Machine Learning Research},
  volume={12},
  pages={2825--2830},
  year={2011}
}

@article{stable-baselines3,
  author  = {Antonin Raffin and Ashley Hill and Adam Gleave and Anssi Kanervisto and Maximilian Ernestus and Noah Dormann},
  title   = {Stable-Baselines3: Reliable Reinforcement Learning Implementations},
  journal = {Journal of Machine Learning Research},
  year    = {2021},
  volume  = {22},
  number  = {268},
  pages   = {1-8},
  url     = {http://jmlr.org/papers/v22/20-1364.html}
}

@inproceedings{DBLP:conf/nips/0001LDL23,
  author       = {Christian Schilling and
                  Anna Lukina and
                  Emir Demirovic and
                  Kim Guldstrand Larsen},
  comeditor       = {Alice Oh and
                  Tristan Naumann and
                  Amir Globerson and
                  Kate Saenko and
                  Moritz Hardt and
                  Sergey Levine},
  title        = {Safety Verification of Decision-Tree Policies in Continuous Time},
  booktitle    = {NeurIPS 2023},
  year         = {2023},
  url          = {http://papers.nips.cc/paper\_files/paper/2023/hash/2f89a23a19d1617e7fb16d4f7a049ce2-Abstract-Conference.html},
  bibsource    = {dblp computer science bibliography, https://dblp.org}
}

@article{gym,
  author       = {Mark {Towers {\em et al.}}},
  title        = {Gymnasium: {A} Standard Interface for Reinforcement Learning Environments},
  journal      = {CoRR},
  volume       = {abs/2407.17032},
  year         = {2024},
  url          = {https://doi.org/10.48550/arXiv.2407.17032},
  doi          = {10.48550/ARXIV.2407.17032},
  eprinttype    = {arXiv},
  eprint       = {2407.17032},
  bibsource    = {dblp computer science bibliography, https://dblp.org}
}

@misc{AI_act,
  author       = "{European Parliament}",
  title        = "Regulation (EU) 2024/1689 of the {European} Parliament and of the Council – The {EU} Artificial Intelligence Act",
  howpublished = "online \url{https://eur-lex.europa.eu/legal-content/EN/TXT/?uri=CELEX%3A32024R1689&qid=1753356665433}",
  month        = "6",
  year         = "2024"
}

@misc{exec_order_14110,
  author       = "{U.S. Executive Office of the President}",
  title        = "Safe, Secure, and Trustworthy Development and Use of Artificial Intelligence",
  comhowpublished = "online \url{https://www.federalregister.gov/documents/2023/11/01/2023-24283/safe-secure-and-trustworthy-development-and-use-of-artificial-intelligence}",
  month        = "11",
  year         = "2023",
  note         = "U.S. Executive
Order 14110"
}

@article{murthy:1994,
author = {Murthy, Sreerama K. and Kasif, Simon and Salzberg, Steven},
title = {A system for induction of oblique decision trees},
year = {1994},
issue_date = {August 1994},
publisher = {AI Access Foundation},
address = {El Segundo, CA, USA},
volume = {2},
number = {1},
issn = {1076-9757},
journal = {J. Artif. Int. Res.},
month = aug,
pages = {1–32},
numpages = {32}
}

@inproceedings{MilaniZTSKPF22,
  author       = {Stephanie Milani and
                  Zhicheng Zhang and
                  Nicholay Topin and
                  Zheyuan Ryan Shi and
                  Charles A. Kamhoua and
                  Evangelos E. Papalexakis and
                  Fei Fang},
  comeditor       = {Massih{-}Reza Amini and
                  St{\'{e}}phane Canu and
                  Asja Fischer and
                  Tias Guns and
                  Petra Kralj Novak and
                  Grigorios Tsoumakas},
  title        = {{MAVIPER:} Learning Decision Tree Policies for Interpretable Multi-agent
                  Reinforcement Learning},
  booktitle    = {
                  {ECML} {PKDD} 2022},
  series       = {LNCS},
  volume       = {13716},
  pages        = {251--266},
  publisher    = {Springer},
  year         = {2022},
  url          = {https://doi.org/10.1007/978-3-031-26412-2\_16},
  doi          = {10.1007/978-3-031-26412-2\_16},
  bibsource    = {dblp computer science bibliography, https://dblp.org}
}

@article{SETIONO19971,
title = {NeuroLinear: From neural networks to oblique decision rules},
journal = {Neurocomputing},
volume = {17},
number = {1},
pages = {1-24},
year = {1997},
issn = {0925-2312},
doi = {https://doi.org/10.1016/S0925-2312(97)00038-6},
url = {https://www.sciencedirect.com/science/article/pii/S0925231297000386},
author = {Rudy Setiono and Huan Liu},
}

@article{Quine1952,
author = {W. V. Quine},
title = {The Problem of Simplifying Truth Functions},
journal = {The American Mathematical Monthly},
volume = {59},
number = {8},
pages = {521--531},
year = {1952},
publisher = {Taylor \& Francis},
doi = {10.1080/00029890.1952.11988183},
URL = { 
        https://doi.org/10.1080/00029890.1952.11988183
},
eprint = { 
        https://doi.org/10.1080/00029890.1952.11988183
}
}

@ARTICLE{McCluskey1956,
  author={McCluskey, E. J.},
  journal={The Bell System Technical Journal}, 
  title={Minimization of {Boolean} functions}, 
  year={1956},
  volume={35},
  number={6},
  pages={1417-1444},
  doi={10.1002/j.1538-7305.1956.tb03835.x}}

@techreport{Moore1990,
author = {Andrew Moore},
title = {Efficient Memory-based Learning for Robot Control},
year = {1990},
month = {November},
institute = {Carnegie Mellon University},
address = {Pittsburgh, PA},
}

@ARTICLE{Govinda2025,
  author={Govinda, Shruti and Brik, Bouziane and Harous, Saad},
  journal={IEEE Transactions on Intelligent Transportation Systems}, 
  title={A Survey on Deep Reinforcement Learning Applications in Autonomous Systems: Applications, Open Challenges, and Future Directions}, 
  year={2025},
  volume={26},
  number={7},
  pages={11088-11113},
  doi={10.1109/TITS.2025.3560379}}

@InProceedings{ross2011,
  title = 	 {A Reduction of Imitation Learning and Structured Prediction to No-Regret Online Learning},
  author = 	 {Ross, Stephane and Gordon, Geoffrey and Bagnell, Drew},
  booktitle = 	 {AISTATS 2011},
  pages = 	 {627--635},
  year = 	 {2011},
  comeditor = 	 {Gordon, Geoffrey and Dunson, David and Dudík, Miroslav},
  volume = 	 {15},
  series = 	 {Proceedings of Machine Learning Research},
  comaddress = 	 {Fort Lauderdale, FL, USA},
  month = 	 {11--13 Apr},
  publisher =    {PMLR},
  url = 	 {https://proceedings.mlr.press/v15/ross11a.html}
}

@article{Krishnamurthy2021,
title = {Explainable classification by learning human-readable sentences in feature subsets},
journal = {Information Sciences},
volume = {564},
pages = {202-219},
year = {2021},
issn = {0020-0255},
doi = {https://doi.org/10.1016/j.ins.2021.02.031},
url = {https://www.sciencedirect.com/science/article/pii/S0020025521001730},
author = {Prashanth Krishnamurthy and Alireza Sarmadi and Farshad Khorrami}
}

@INPROCEEDINGS{Nageshrao2019,
  author={Nageshrao, Subramanya and Costa, Bruno and Filev, Dimitar},
  booktitle={ICMLA 2019}, 
  title={Interpretable Approximation of a Deep Reinforcement Learning Agent as a Set of If-Then Rules}, 
  year={2019},
  volume={},
  number={},
  pages={216-221},
  doi={10.1109/ICMLA.2019.00041}}

@InProceedings{contrerasolivas2025,
author="Contreras Olivas, Daniel Adri{\'a}n
and Martinez-Villase{\~{n}}or, Lourdes",
comeditor="Mart{\'i}nez-Villase{\~{n}}or, Lourdes
and Mart{\'i}nez-Seis, Bella
and Pichardo, Obdulia",
title="Human-Friendly Explanations Checklist for Reinforcement Learning: {XRL H-F-E} Checklist",
booktitle="Artificial Intelligence -- COMIA 2025",
year="2025",
compublisher="Springer Nature Switzerland",
comaddress="Cham",
pages="268--279",
isbn="978-3-031-97910-1"
}

@book{molnar2025,
  title={Interpretable Machine Learning},
  subtitle={A Guide for Making Black Box Models Explainable},
  author={Christoph Molnar},
  year={2025},
  edition={3},
  isbn={978-3-911578-03-5},
  url={https://christophm.github.io/interpretable-ml-book}
}

@article{DBLP:journals/corr/abs-2411-15594,
  author       = {Jiawei Gu and
                  Xuhui Jiang and
                  Zhichao Shi and
                  Hexiang Tan and
                  Xuehao Zhai and
                  Chengjin Xu and
                  Wei Li and
                  Yinghan Shen and
                  Shengjie Ma and
                  Honghao Liu and
                  Yuanzhuo Wang and
                  Jian Guo},
  title        = {A Survey on {LLM}-as-a-Judge},
  journal      = {CoRR},
  volume       = {abs/2411.15594},
  year         = {2024},
  url          = {https://doi.org/10.48550/arXiv.2411.15594},
  doi          = {10.48550/ARXIV.2411.15594},
  eprinttype    = {arXiv},
  eprint       = {2411.15594},
  bibsource    = {dblp computer science bibliography, https://dblp.org}
}

@article{Thorndike_1953, 
title={Who Belongs in the Family?}, 
volume={18}, 
DOI={10.1007/BF02289263}, 
number={4}, 
journal={Psychometrika}, 
author={Thorndike, Robert L.}, 
year={1953}, 
pages={267–276}}
